%% file: root.tex
\documentclass[journal]{IEEEtran}

\IEEEoverridecommandlockouts      

\usepackage{graphics} 
\usepackage{graphicx} %
\usepackage{gensymb}

\usepackage{tikz}
\usepackage{verbatim}
\usetikzlibrary{calc}

\usepackage{amstext} 
\usepackage{amsmath} 
\usepackage{amssymb}  
\usepackage{mathtools}
\usepackage{algorithmic}

\usepackage{dirtytalk}

\usepackage[ruled,lined]{algorithm2e}

\usepackage[utf8]{inputenc}
\usepackage{cite}
\usepackage{textcomp}
\usepackage{tabularx}
\usepackage{xcolor}
\usepackage{subcaption}
\usepackage{soul}
\usepackage{tikz,xcolor}
\usepackage[font=footnotesize]{caption}

\usepackage[colorlinks,citecolor=red,urlcolor=black,bookmarks=false,hypertexnames=true]{hyperref}

\usepackage[dvipsnames]{xcolor} 

\usepackage{multirow}  
\usepackage{array} 
\newcolumntype{L}{>{$}l<{$}} 

\usepackage{comment}

\usepackage{url}

\def\BibTeX{{\rm B\kern-.05em{\sc i\kern-.025em b}\kern-.08em
    T\kern-.1667em\lower.7ex\hbox{E}\kern-.125emX}}

\DeclareMathOperator*{\torsoPose}{torsoPose}
\DeclareMathOperator*{\getReachableRegions}{getReachableRegions}
\DeclareMathOperator*{\isStanceStable}{isStanceStable}
\DeclareMathOperator*{\isSwingStable}{isSwingStable}
\DeclareMathOperator*{\buildNode}{buildNode}
\DeclareMathOperator*{\forwardProgress}{forwardProgress}

\title{\LARGE \bf QuadPiPS: A Perception-informed Footstep Planner for Quadrupeds With Semantic Affordance Prediction}

\author{{Max Asselmeier, Ye Zhao, and Patricio A. Vela}
\thanks{M. Asselmeier, Y.Zhao, and P.A. Vela are with the Institute for Robotics and Intelligent Machines, Georgia Institute of Technology, Atlanta GA 30308, USA,{\tt\small \{mass, yzhao301, pvela\}@gatech.edu}}%
\thanks{The work of Max Asselmeier is supported by the National Science Foundation Graduate Research Fellowship under Grant No. DGE-2039655. Any opinion, findings, and conclusions or recommendations expressed in this material are those of the authors(s) and do not necessarily reflect the views of the National Science Foundation.}
}%

\begin{document}

\maketitle
\thispagestyle{empty}
\pagestyle{empty}


\begin{abstract}
This work proposes QuadPiPS, a perception-informed framework for quadrupedal foothold planning in the perception space. QuadPiPS employs a novel ego-centric local environment representation, known as the legged egocan, that is extended here to capture unique legged affordances through a joint geometric and semantic encoding that supports local motion planning and control for quadrupeds. QuadPiPS takes inspiration from the Augmented Leafs with Experience on Foliations (ALEF) planning framework to partition the foothold planning space into its discrete and continuous subspaces. To facilitate real-world deployment, QuadPiPS broadens the ALEF approach by synthesizing perception-informed, real-time, and kinodynamically-feasible reference trajectories through search and trajectory optimization techniques. To support deliberate and exhaustive searching, QuadPiPS over-segments the egocan floor via superpixels to provide a set of planar regions suitable for candidate footholds. Nonlinear trajectory optimization methods then compute swing trajectories to transition between selected footholds and provide long-horizon whole-body reference motions that are tracked under model predictive control and whole body control. Benchmarking with the ANYmal C quadruped across ten simulation environments and five baselines reveals that QuadPiPS excels in safety-critical settings with limited available footholds. Real-world validation on the Unitree Go2 quadruped equipped with a custom computational suite demonstrates that QuadPiPS enables terrain-aware locomotion on hardware.

\end{abstract}

\input{text_files/introduction}

\input{text_files/related_works}

\input{text_files/framework}

\input{text_files/preliminaries}

\input{text_files/perception}

\input{text_files/planning}

\input{text_files/control}

\input{text_files/experimental_results}

\input{text_files/discussion}

\input{text_files/conclusion}

\bibliographystyle{IEEEtran}
\bibliography{references}

\end{document}

%% file: text_files/introduction.tex
\section{Introduction}
\label{sec:introduction}
The recent proliferation of legged robot platforms within both academia \cite{bledt_mit_2018, chignoli_mit_2021, zhu_artemis_nodate, hutter_anymal_2017} and industry \cite{boston_dynamics_atlas_nodate, boston_dynamics_spot_nodate, unitree_robotics_humanoid_nodate, robotics_fauna_2026, noauthor_hello_nodate} suggests that complex autonomous tasks such as real-world navigation and locomotion may be attainable in the near future. Formal competitions such as the DARPA Subterranean (SubT) challenge \cite{darpa_subterranean_nodate, tranzatto_cerberus_2022, chung_into_2023}, the BARN \cite{perille_benchmarking_2020} and DynaBARN challenges \cite{nair_dynabarn_2022}, and the ICRA Quadruped Challenge \cite{jacoff_taking_nodate} have all demonstrated that for tasks such as these, well-defined hierarchical frameworks enable autonomous and reliable real-world operation. 

Furthermore, navigation and planning performance are heavily coupled with the design of the perception system at hand. Poorly designed environment representations bound the capabilities of downstream planning and control, and these representations often act as the computational bottleneck for online frameworks. Fully autonomous platforms must minimize the latency in environment processing while also maximizing planner expressivity. This work augments the environment representation known as the egocan \cite{smith_aerialpips_2023} arising from the Planning in the Perception Space (PiPS) \cite{smith_pips_2017} philosophy which bypasses sensor preprocessing and casts the act of planning as a sequence of egocentric decision-making problems. Egocan extensions include the geometric attribute of surface normals and the semantic environmental affordances \cite{gibson_ecological_2014} of steppability labels which are both vital to communicating movement opportunities for legged systems to planning and control layers. Moreover, only a single front-facing depth camera is required to construct this full 360$\degree$ model of the local environment. This work marks the first exploration into how the egocan can be utilized for legged locomotion. It also serves as an investigation into how alternative representations beyond traditional elevation maps \cite{miki_elevation_2022, hoeller_neural_2022} and voxel grids \cite{miki_learning_2024} may be better suited for an efficient perception-to-action pipeline for local motion planning on quadrupeds.  

\begin{figure}[t!]
    \centering
    \includegraphics[width=0.99\linewidth]{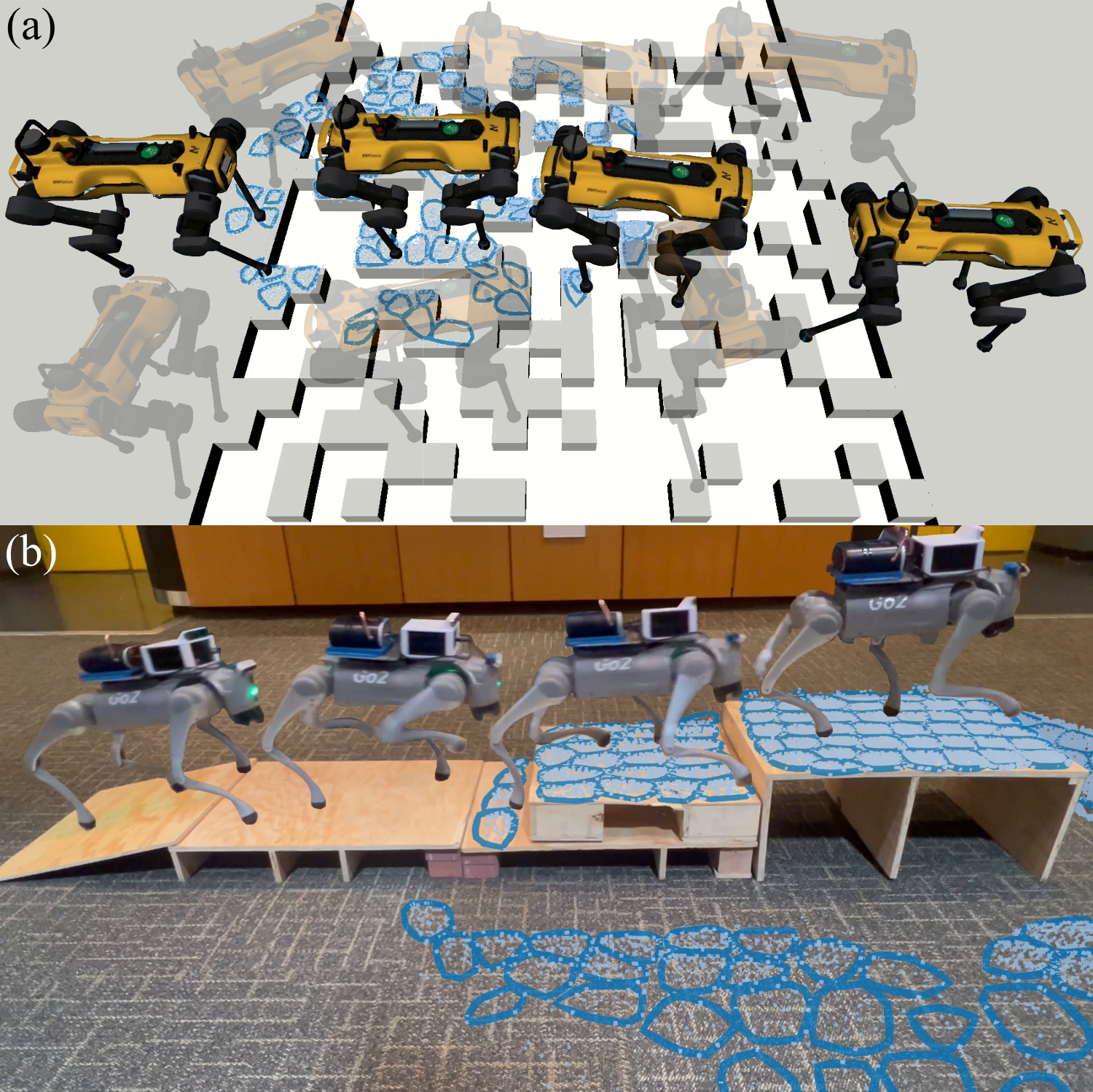}
    \caption{QuadPiPS framework deployed in simulation on the ANYmal C and on hardware on the Unitree Go2. (a) ANYmal C planning over a set of sparsely distributed stepping stones. Transparent robots represent stance configurations visited during the graph search while solid robots represents the actual path taken by the robot. (b) Unitree Go2 equipped with a custom computational suite for onboard deployment planning over complex terrain including ramps and stairs. Superpixels are represented as blue points and boundaries. }
    \label{fig:vision_figure}
    \vspace{-2em}
\end{figure}

To perform foothold planning over the egocan, this work draws inspiration from the Augmented Leafs with Experience on Foliations (ALEF) framework \cite{kingston_scaling_2023}. However, existing ALEF implementations \cite{asselmeier_hierarchical_2024} perform offline kinematic planning over entirely known environments, and the time required to generate feasible paths ranges from seconds to minutes. In this work, an extension of the Simple Linear Iterative Clustering (SLIC) \cite{achanta_slic_2012} superpixels algorithm oversegments the egocan floor to facilitate exhaustive searching over candidate footholds through a contact mode family transition graph. Then, a nonlinear trajectory optimization formulation synthesizes kinodynamically-feasible and real-time whole body reference trajectories that transition the system between stances. Lastly, a Model Predictive Controller (MPC) and Whole Body Controller (WBC) track this reference trajectory. This entire framework for \textbf{Quad}rupedal Foothold \textbf{P}lanning \textbf{i}n the \textbf{P}erception \textbf{S}pace is referred to hereafter as \textbf{QuadPiPS}.

Analysis of the implementation consists of simulation-based benchmarking over ten diverse environments – featuring gaps, beams, and barriers – and five disparate foothold planning baselines – featuring model-based, model-free, and hybrid approaches – using the ANYmal C quadruped \cite{hutter_anymal_2017}. The experimental outcomes include that (a) the augmented egocan-based perception pipeline improves upon traditional elevation mapping-based representations in environments with numerous planar regions, and (b) the QuadPiPS framework as a whole improves upon heuristic foothold planners in environments with a limited number of available footholds. Validation of the QuadPiPS framework consists of fully onboard and sensor-informed real-world deployment over five challenging terrain configurations – featuring ramps, stairs, and stepping stones – using the Unitree Go2 quadruped \cite{unitree_robotics_robot_nodate} equipped with a custom fabricated computational suite, as seen Figure \ref{fig:vision_figure}.

In summary, the contributions of this work are as follows.
\begin{enumerate}
    \item Augmentation of the egocan representation to support legged environmental affordances including geometric surface normals and semantic steppability labels
    \item Synthetic data generation through primitive shapes-based techniques to construct simulation scenes for predicting semantic environmental affordances
    \item Oversegmentation of the egocan floor image using a novel implementation of the SLIC algorithm for sparse multi-channel images to facilitate foothold searching
    \item Adaptation of the ALEF framework to support perception-informed, real-time, and dynamically-feasible motion planning for quadrupeds
    \item Simulation benchmarking of QuadPiPS against state-of-the-art foothold planners across ten diverse environments using the ANYmal C platform
    \item Real-world validation of QuadPiPS on five difficult terrain setups using the Unitree Go2 quadruped fitted with a custom onboard computational suite
\end{enumerate}

%% file: text_files/related_works.tex
\section{Related Works} \label{sec:related_works}

\subsection{Legged Environment Representations}
Within an autonomy stack, one of the earliest design decisions is how to represent the robot's local environment. In the literature for legged robots, the most popular model is that of the 2.5D height map or elevation map \cite{thrun_learning_2003, fankhauser_robot-centric_2014, fankhauser_probabilistic_2018, karkowski_prediction_2019, wellhausen_rough_2021, behnke_team_2015, miller_mine_2020, miki_elevation_2022}. An elevation map is a grid structure represented in a Cartesian frame, either an inertial frame or robot-aligned frame. This representation has roots in the early stages of perception-informed legged locomotion, namely the DARPA Learning Locomotion project \cite{neuhaus_comprehensive_2011, pippine_overview_2011} where high-resolution height maps were provided to each team. Height maps are popular due to their simple structure, Cartesian frame representation, and open-sourced implementation \cite{noauthor_anyboticselevation_mapping_2024}. However, constructing such a model requires point cloud processing which often necessitates GPU parallelization \cite{miki_elevation_2022} or multi-threading to operate in real time. 


Elevation maps can be used to build further representations such as discrete occupancy maps \cite{stepanas_ohm_2022} through height thresholding or fast collision checking, continuous cost maps \cite{jenelten_perceptive_2020} which build cost functions based on neighboring height information, or signed distance fields \cite{zucker_optimization_2011, zucker_chomp_2013, grandia_perceptive_2022} which calculate gradient information regarding distance between the environment and the robot. More recently, neural scenes \cite{hoeller_neural_2022} use learned networks to fill in gaps in environment representations due to occlusions or sensor artifacts. Attention-based map encodings \cite{he_attention-based_2025, zhang_ame-2_2026} learn attention weights to influence where footholds should be planned. Elevation maps are also often used to construct planar polygonal regions \cite{bertrand_detecting_2020, miki_elevation_2022, karkowski_prediction_2019, grandia_perceptive_2022, griffin_footstep_2019, grandia_multi-layered_2021, acosta_bipedal_2023} where further processing steps can be performed to extract planar regions of the environment that can support a footstep.

For environments where it is important to capture multiple height layers --- including floors, obstacles, and ceilings --- then 3D environment representations can be deployed such as the voxel grid \cite{oleynikova_voxblox_2017, agha-mohammadi_confidence-rich_2019, frey_locomotion_2022}. In this case, all three dimensions are discretized, giving the user the freedom to model layers at the cost of memory and computation. For the SubT competition \cite{darpa_subterranean_nodate}, which occurred in underground settings such as tunnels and caves, many teams opted for these models \cite{hines_virtual_2021, tranzatto_cerberus_2022}. Another flavor of voxel grids is the Octomap \cite{hornung_octomap_2013}, which exploits environmental sparsity to efficiently build a volumetric grid.

All representations discussed here require point cloud processing to be constructed. This includes costly steps such as filtering, ray casting, and transformations. By propagating environment information through the egocan, QuadPiPS can construct a height map without any of this preprocessing. 



%

%

\subsection{Legged Affordances}
The distinct ways in which legged robots can interact with the environment, referred to as affordances \cite{gibson_ecological_2014}, are far more diverse than those of their wheeled or tracked counterparts. It then becomes important to capture these movement opportunities in the local environment representation to imbue an autonomy stack with the full capabilities of the particular robot platform. While affordances have existed as a concept outside of robotics for many years, they have only been adopted into autonomy frameworks as a way to ground reasoning in embodiments since the DARPA Robotics Challenge \cite{fallon_architecture_2015, althoefer_contact_2019, fang_saga_2025}.

One of these affordances is what a majority of the literature refers to as \textit{traversability} \cite{lin_humanoid_2018, jenelten_perceptive_2020, agha_nebula_2022, kottege_heterogeneous_2025, yoon_state-nav_2026}. Terrain traversability can be viewed as a continuous score assigned to a subsection of the local environment that reflects the ability to traverse the terrain if a foothold is placed there. Traversability is often calculated as a function of terrain properties such as height discontinuities \cite{stepanas_ohm_2022}, gradients \cite{miller_mine_2020, dixit_step_2025}, and curvature \cite{biggie_flexible_2023}. Traversability has also been learned through neural networks \cite{wellhausen_rough_2021, miki_elevation_2022}. This metric is typically stored in a 2D cost map that is either searched \cite{agha_nebula_2022, kottege_heterogeneous_2025} or optimized \cite{jenelten_perceptive_2020} over for planning.  Some approaches do leverage the image space for terrain processing, but purely for the task of terrain class identification \cite{wellhausen_where_2019}. This motivates a deeper investigation into image space-based representations for legged platforms.

While continuous scoring of terrain is an important metric for footstep planning, the discrete decomposition of the local environment into \textit{steppable} and \textit{non-steppable} regions is paramount. The level of granularity can differ across approaches, with some making a binary classification between these two labels \cite{karkowski_prediction_2019, wellhausen_rough_2021} while others further decompose the environment into gait- or behavior-specific regions such as jumping \cite{kim_vision_2020}.  Another common approach is to decompose the environment into a set of polygons \cite{griffin_footstep_2019, bertrand_detecting_2020, buchanan_perceptive_2021, grandia_perceptive_2022} that represent support regions for footholds. Such a representation fits nicely into optimization frameworks for footstep planning. While existing label schemes \cite{kim_vision_2020} may be similar to the one presented in this proposed work, all methods discussed here maintain a robot-centric 2D grid-based representation. The image space-based labeling in QuadPiPS fits well with computer vision-based segmentation tasks and aligns with the PiPS ideology.

\subsection{Perception-Planning-Control Frameworks for Quadrupeds}
For long-horizon planning and navigation, hierarchical frameworks are essential in decomposing these complex missions into simpler subtasks that can each be addressed through disparate planning and control modules. For the SubT Challenge \cite{darpa_subterranean_nodate}, the CERBERUS team \cite{tranzatto_cerberus_2022}, equipped with the ANYmal platform, employed a combination of voxel grids \cite{kulkarni_autonomous_2022} for global planning and height maps for local navigation planning \cite{wellhausen_artplanner_2023}. The team searched over global and local graphs for torso paths. Footholds were resolved explicitly through heuristic placement and local optimization via MPC \cite{jenelten_perceptive_2020} or implicitly through learned low-level control policies \cite{lee_learning_2020}. The CoSTAR team \cite{agha_nebula_2022} constructed elevation maps that they computed traversability \cite{dixit_step_2025} over for risk-aware geometric global planning \cite{bouman_autonomous_2020} with the Boston Dynamics Spot. This global planner \cite{kim_plgrim_2021} performed a trajectory library-based search for a global torso path, and a kinodynamic MPC planner then tracked waypoints along the global path. Both teams relied heavily upon graph searches for autonomous planning, but they restricted their search space to the torso pose. QuadPiPS shows that under the proper low-dimensional contact mode family representations, real-time searching for footholds is not only viable, but also preferable in safety-critical environments.

In the Vision-Based Terrain-Aware Locomotion (ViTAL) framework \cite{fahmi_vital_2023}, the vision-based foothold adaptation \cite{villarreal_fast_2019} and vision-based pose adaptation algorithms plan according to a shared set of foothold evaluation criteria. ViTAL uses a height map to predict optimal footholds through a convolutional neural network, and a pose optimizer finds a torso pose which maximizes the set of safe footholds available. Then, a whole body controller \cite{fahmi_passive_2019} generates torques that are tracked by a joint impedance controller \cite{boaventura_model-based_2015} on the HyQ platform \cite{semini_design_2011}. ViTAL successfully plans over stairs and rough terrain, but it does not deploy in environments where precise foothold planning is necessary. The MIT Mini-Cheetah navigated to a goal and avoided torso-level obstacles using a height map and artificial potential field methods \cite{kim_vision_2020} as well as search-based methods \cite{dudzik_robust_2020}, both applied in the torso pose space. Grid map filtering provided labels similar to those presented in QuadPiPS which are used to accept or reject heuristically-calculated foothold positions. However, these labels are calculated through pure geometric reasoning such as height gradient thresholding. Heuristic footholds are finally tracked by a whole body controller \cite{kim_highly_2019}

More recently, systems opt to hierarchically decompose navigation into simpler subtasks which are then handled via learned modules. Both the ANYmal C \cite{hoeller_anymal_2024} and Unitree Go2 \cite{chen_learning_2025} navigated through cluttered real-world terrains using a reconstructed elevation map, a learned local navigation policy which provides a velocity command for the system, and a learned locomotion policy capable of maneuvers including walking, crawling, and jumping. These learned frameworks demonstrate impressive robustness and generalization, but they are still tested in environments with numerous available footholds. 

All frameworks discussed in this section rely on 2.5D elevation maps. Though, these maps possess computational and representational drawbacks including reliance on point clouds and sensitivity to confined spaces.

\subsection{Planning in the Perception Space (PiPS)}
This proposed work was designed according to the PiPS philosophy \cite{smith_pips_2017}. A perception space-based approach to planning eschews preprocessing steps for perception data and instead acts directly on the raw data representations that an external sensor provides. This means keeping the sensor data in an egocentric reference frame and recasting the task of local perception-informed planning as a robot-centric decision-making process. 

To date, PiPS has provided strong benefits for fast depth space collision-checking \cite{smith_pips_2017}, collision-free local robot navigation \cite{xu_potential_2021, smith_real-time_2020, smith_egoteb_2020}, and task and motion planning \cite{driess_deep_2020}. The PiPS approach has also proven effective for both model-based \cite{miller_mine_2020, feng_gpf-bg_2023} and model-free \cite{sorokin_learning_2022, agarwal_legged_2023} approaches to legged motion planning. However, torso-level planning and end-to-end efforts have dominated the PiPS literature. A thorough investigation into how the PiPS paradigm should be applied to foothold planning has yet to be performed. 

While PiPS can facilitate certain local robot processes such as motion planning and collision checking, the perception space has its drawbacks as well. The rich nearby depth information is offset by sparse environment information at further distances, which limits the spatio-temporal horizons in which planning in the perception space is effective. 

%% file: text_files/framework.tex
\section{Overall QuadPiPS Framework} \label{sec:framework}

\begin{figure*}[t!]
    \centering
    \begin{tikzpicture}
    \hspace*{-0.5in}
    \draw node[scale=1.0](figure) at (0, 0){\input{text_files/diagram2}};
    \end{tikzpicture}
    \caption{Workflow for the QuadPiPS framework. The red modules represent the perception pipeline, blue modules represent the planning components, and green modules represent control. Dashed blocks represent novel components that are introduced in this proposed framework. Arrows represent downstream dependencies, meaning that the robot state $\mathbf{x}$ is incorporated into several modules downstream from the semantic egocan.}
    \label{fig:info_flow}
\end{figure*}

The QuadPiPS framework (see Figure \ref{fig:info_flow}) addresses perception-informed motion planning for quadrupedal locomotion over complex terrain. In this work, the only information known a priori is a goal pose in the world to attain.

The original egocan environment representation (Section \ref{sec:preliminaries}) possesses depth information obtained from an incoming stereo image stream. QuadPiPS augments the egocan to also include geometric surface normals (Section \ref{sec:normal_estimation}) at imaging frame rate ($30$~Hz on hardware) and semantic steppability labels (Section \ref{sec:steppability}) at $5$~Hz.  Surface normals are estimated from incoming depth image gradients. Steppability labels are predicted through a semantic segmentation model. New affordance information is inserted into the egocan (Section \ref{sec:legged_egocan}), and old information is propagated forward in time at $150$~Hz to enable a $360\degree$ view of the local environment. Near-ground and under-foot data is maintained through an egocan floor image which is oversegmented into superpixel planar regions (Section \ref{sec:superpixels}) at $20$~Hz which are convexified (Section \ref{sec:convex_planar_regions}) and passed on for foothold planning.  


To solve this motion planning problem (Section \ref{sec:problem_statement}), the incoming set of planar regions is used to synthesize a planning graph online for footholds. A coarse torso search (Section \ref{sec:torso_search}) at $30$~Hz informs the foothold graph (Section \ref{sec:graph_search}) which is built online according to geometric and kinematic constraints, allowing the structure and sparsity of the local environment to inform planning. This planning graph is searched over at $15$~Hz with $A^*$ \cite{hart_formal_1968} using solely a Euclidean distance heuristic to guide the search. Once this search returns a set of kinematically feasible stance configurations that takes the quadruped from start to goal, a sequence of trajectory optimization subproblems (Section \ref{sec:traj_opt}) are run at $7$~Hz to synthesize dynamic whole body trajectories that carry the robot between stance configurations.

Once this whole body reference trajectory (Section \ref{sec:ref_traj_pub}) has been generated, it is sent to an MPC+WBC framework for tracking. The MPC formulation (Section \ref{sec:mpc}) runs at $50$~Hz and mirrors the formulation used to generate the reference trajectory, while the WBC formulation (Section \ref{sec:wbc}) can run at over $1$~kHz and takes on a hierarchical QP structure, prioritizing dynamic feasibility constraints over less critical terms including reference trajectory tracking.

%% file: text_files/diagram2.tex
\tikzstyle{block} = [draw, rectangle, text centered, thick,rounded corners=2pt,
                     minimum height=1.5em, minimum width=5em, inner sep=4pt]
\tikzstyle{contribBlock} = [draw, rectangle, text centered, ultra thick,
					 minimum height=1.5em, 
					 minimum width=5em, inner sep=4pt,
					 dash pattern=on 1pt off 2pt on 4pt off 2pt]
\tikzstyle{typical} = [fill=white!95!black]
\tikzstyle{reddish} = [draw=red,fill=white!95!red]
\tikzstyle{blueish} = [draw=blue,fill=white!95!blue]
\tikzstyle{orangeish} = [draw=orange!40!gray,fill=white!95!orange]

\tikzstyle{greenish} = [draw=green!40!gray,fill=white!95!green]

\tikzstyle{dashedBlock} = [draw, dashed, rectangle, ultra thick, minimum height=1em, minimum width=1em, inner sep=4pt]
\tikzstyle{newtip} = [->, very thick]
\tikzstyle{notip} = [-, very thick]
\tikzstyle{bidir} = [<->, very thick]
\tikzstyle{newtip_dashed} = [->, very thick, dashed]

\begin{tikzpicture}[auto, inner sep=0pt, outer sep=0pt, >=latex]


\node[anchor=center, text width=6em] (quadpips) at ($(0.0, 0.0)$) {\centering \textbf{ \large QuadPiPS}};

\node[anchor=center, text width=6em] (d435_image) at ($(quadpips.center) + (0.25, -1.5)$) {\includegraphics[width=0.75\linewidth]{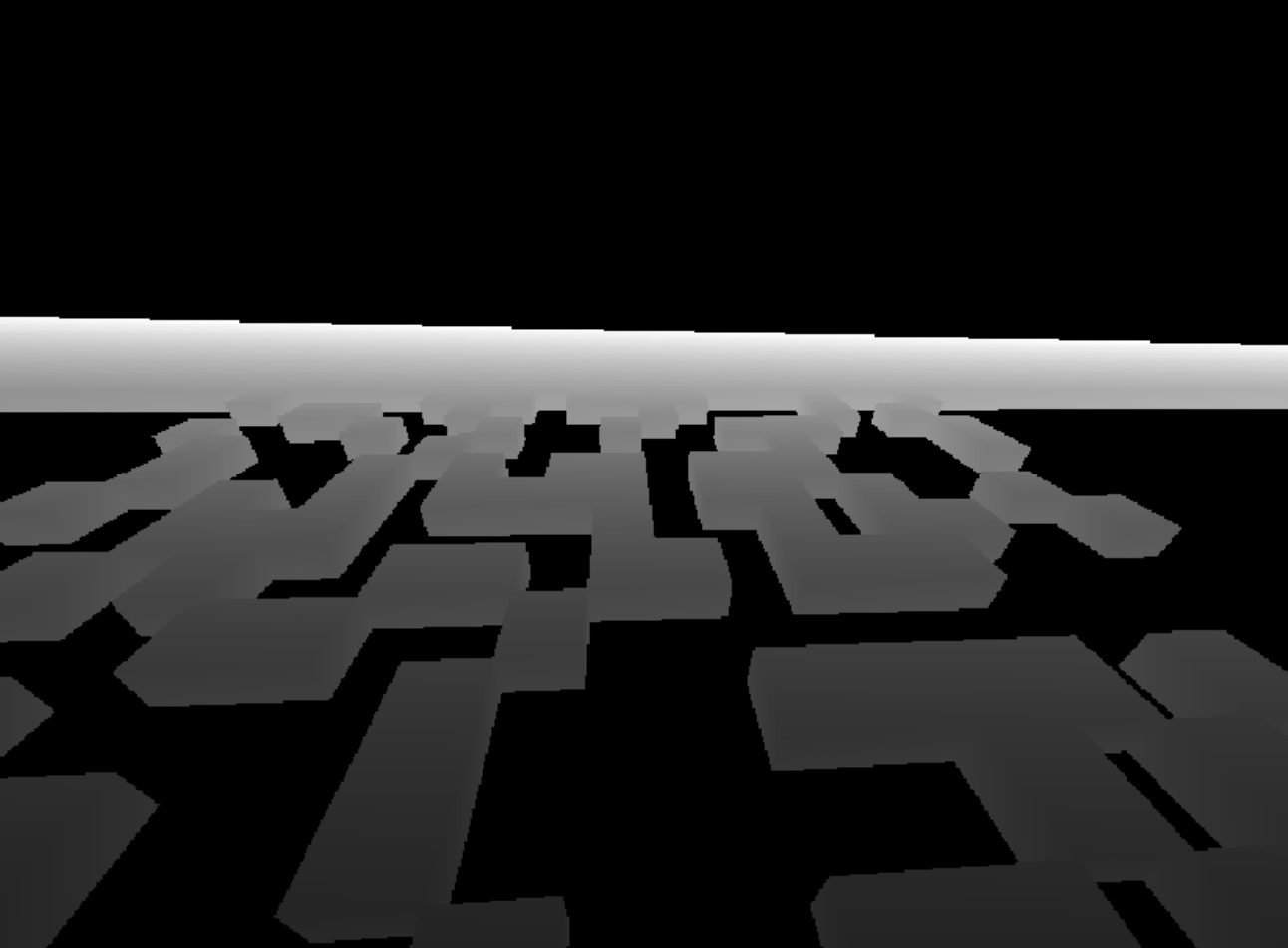}};

\node[block, reddish, anchor=center, text width=4em] (depth_image) at ($(d435_image.center) + (-0.25, -1.25)$) {\centering Depth \\ Image};

\node[contribBlock,gray,anchor=center,minimum width=4em,text width=4em, minimum height=3em] (legend) at ($(depth_image.center) + (0, -2.5)$) {};
\node[anchor=center] (legend_name) at ($(legend.center) + (0, -0.75)$) {Contributions};


\node[anchor=center] (depth_image_symbol) at ($(depth_image.center) + (3.0, 0.0)$ ) {\centering $I_{\rm depth}$};

\node[block, reddish, anchor=center, text width=6em] (normal_estimation) at ($(depth_image_symbol.center) + (0.0, 1.50)$) {\centering Normal \\ Estimation (Section \ref{sec:normal_estimation})};

\node[anchor=center, text width=6em] (normal_image) at ($(normal_estimation.center) + (0.25, 1.50)$) {\includegraphics[width=0.75\linewidth]{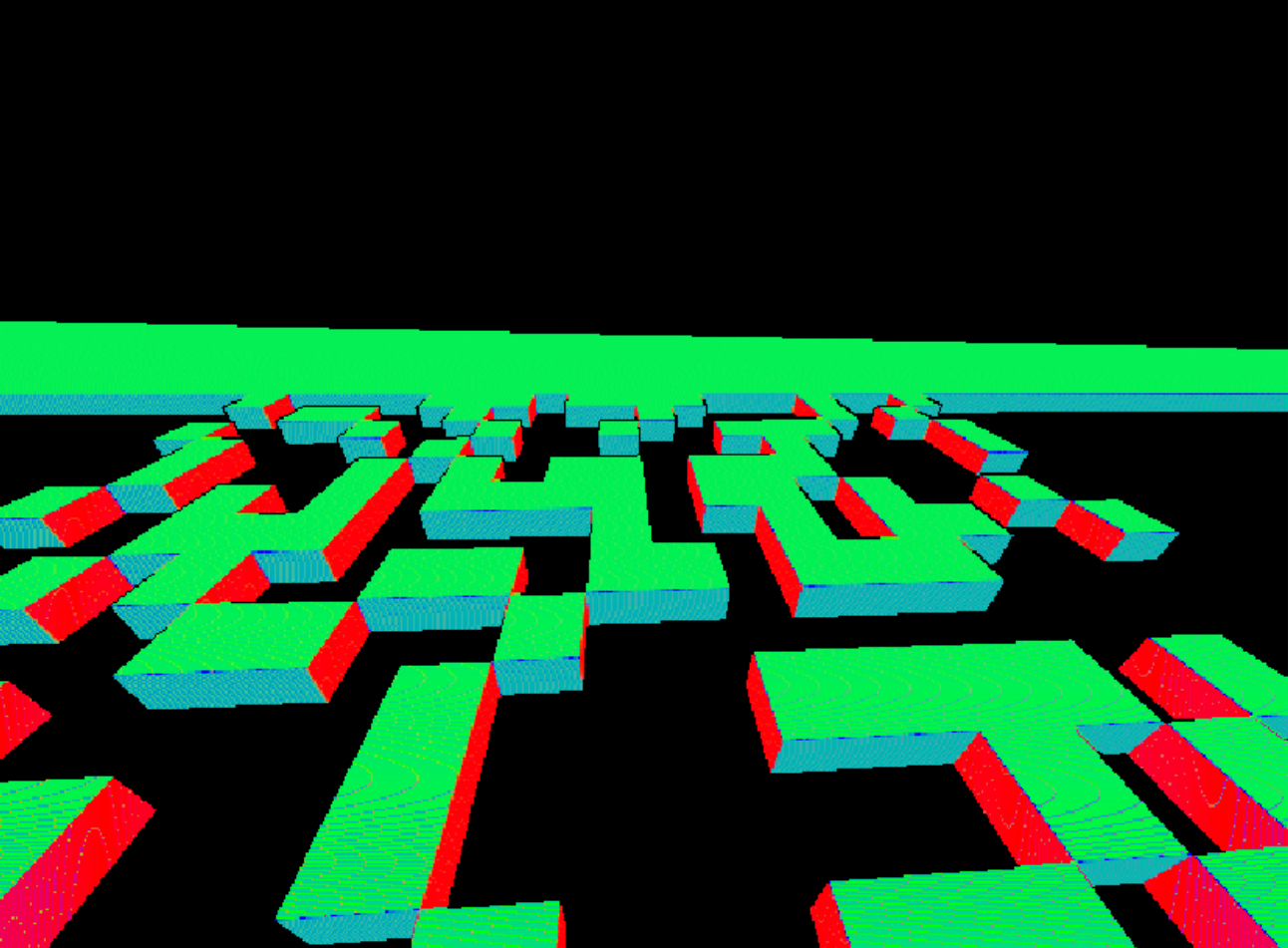}};

\node[anchor=center] (normal_image_symbol) at ($(normal_image.center) + (1.625, -1.25)$ ) {\centering $I_{\rm normal}$};

\node[anchor=center, text width=6em] (steppability_image) at ($(depth_image_symbol.center) + (0.25, -1.5)$) {\includegraphics[width=0.75\linewidth]{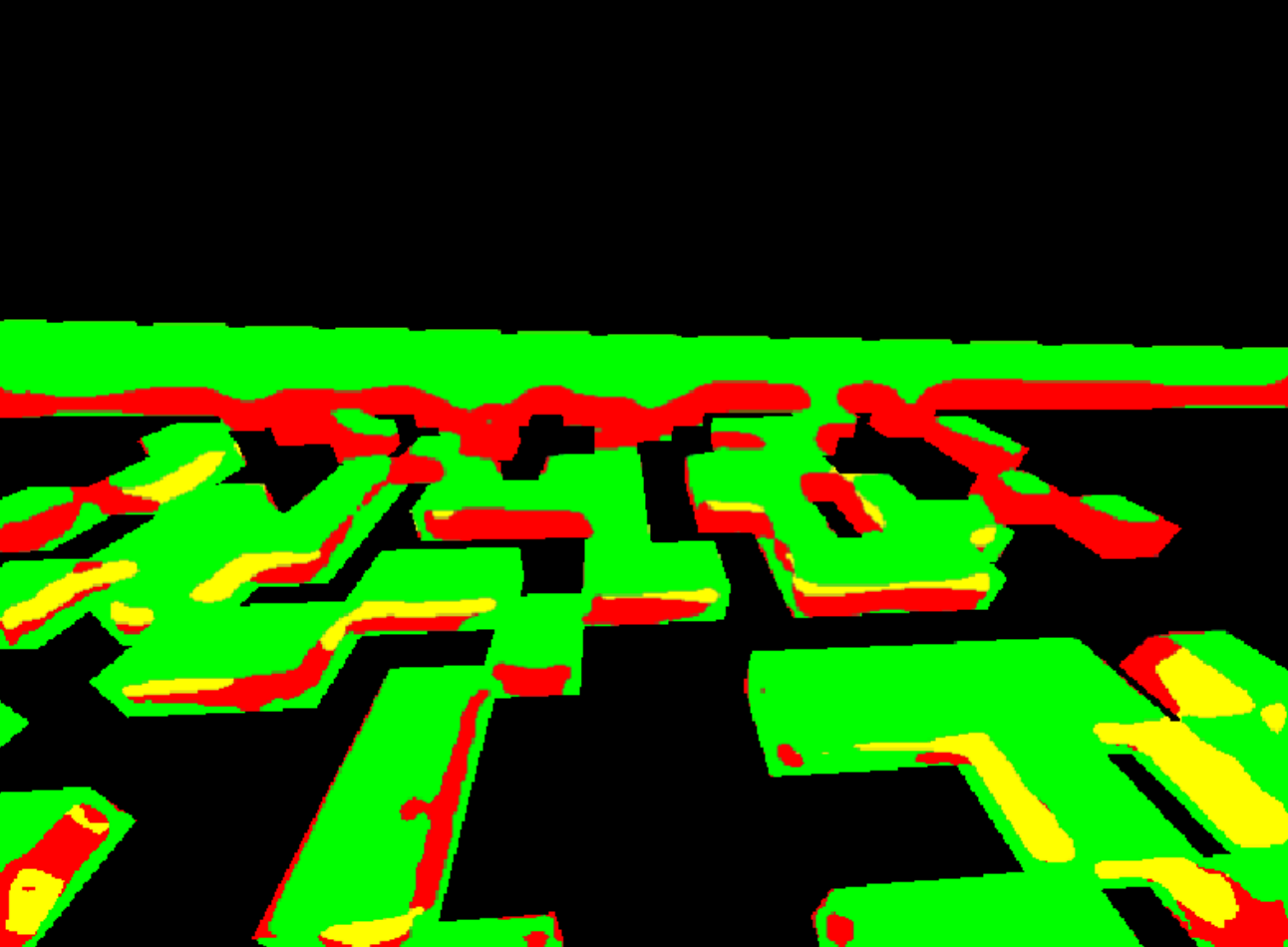}};

\node[contribBlock,reddish,anchor=center, text width=6em] (steppability_model) at ($(steppability_image.center) + (-0.25, -1.50)$) {\centering Steppability Segmentation (Section \ref{sec:steppability}) };

\node[anchor=center] (step_image_symbol) at ($(steppability_model.center) + (2.0, 0.15)$ ) {\centering $I_{\rm step}$};

\draw[notip, black] ($(depth_image.east)$) -- ($(depth_image_symbol.west) + (-0.10, 0.0)$);

\draw[notip, black] ($(depth_image.east) + (0.375, 0.0)$) -- ($(depth_image.east) + (0.375, 1.5)$);

\draw[newtip, black] ($(depth_image.east) + (0.375, 1.5)$) -- ($(normal_estimation.west)$);

\draw[notip, black] ($(depth_image.east) + (0.375, 0.0)$) -- ($(depth_image.east) + (0.375, -3.0)$);

\draw[newtip, black]  ($(depth_image.east) + (0.375, -3.0)$) -- ($(steppability_model.west)$);


\node[contribBlock,reddish,anchor=center, text width=7.25em] (legged_egocan) at ($(depth_image_symbol.center) + (4.0, -2.25)$) {\centering Legged Egocan (Section \ref{sec:legged_egocan})};

\node[anchor=center, text width=6em] (egocan_image) at ($(legged_egocan.center) + (0.25, 1.25)$) {\includegraphics[width=0.75\linewidth]{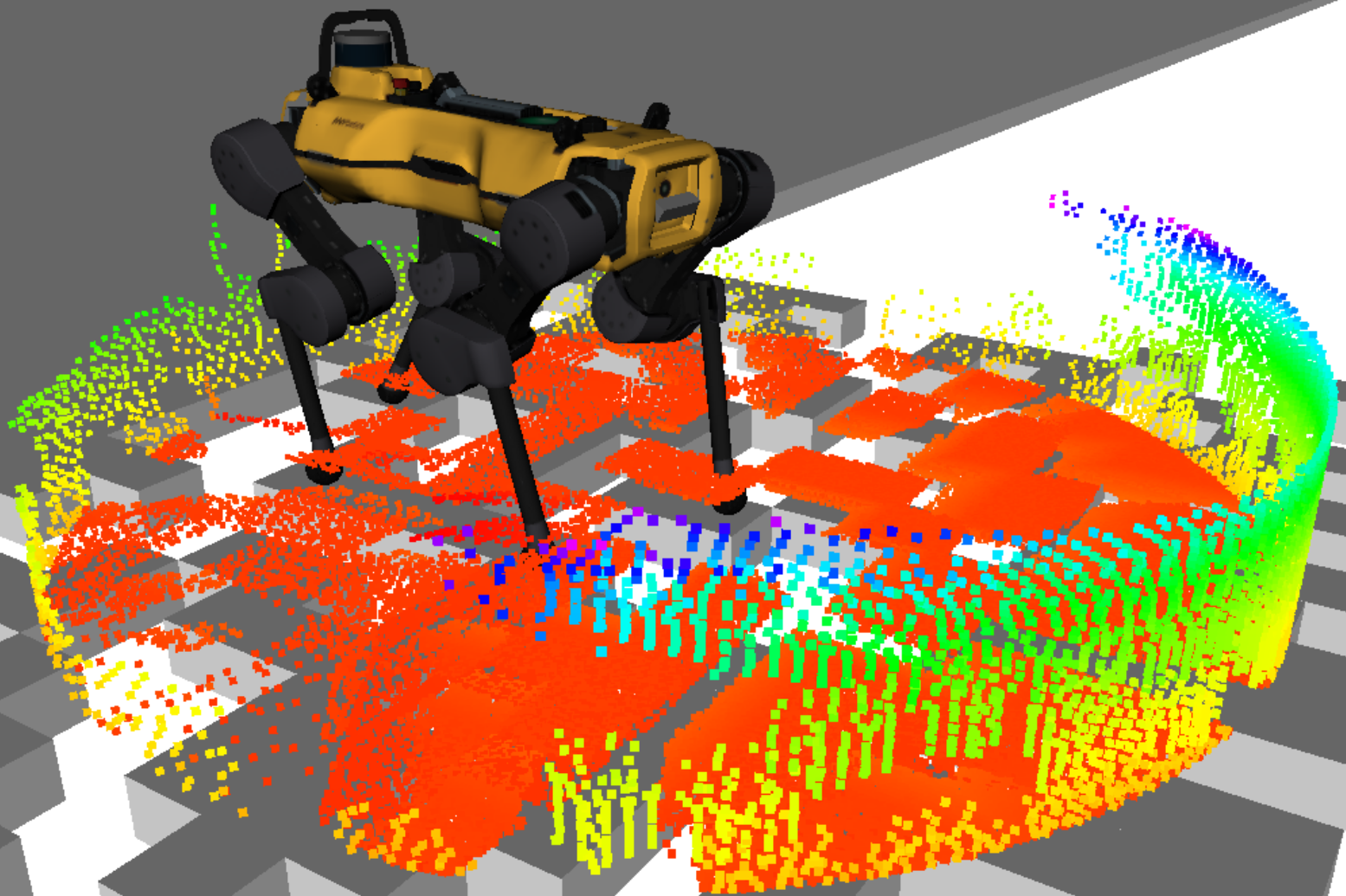}};

\node[anchor=center] (floor_image_symbol) at ($(legged_egocan.center) + (0.50, 2.375)$ ) {\centering $I_{\rm low}$};

\node[contribBlock,reddish,anchor=center, text width=7.50em] (superpixels) at ($(depth_image_symbol.center) + (4.0, 1.25)$) {\centering Superpixels Oversegmentation (Section \ref{sec:superpixels})};

\node[anchor=center] (planar_regions_symbol) at ($(superpixels.center) + (1.75, 1.35)$ ) {\centering $\mathcal{P}$};

\node[anchor=center, text width=6em] (superpixels_image) at ($(superpixels.center) + (0.25, 1.625)$) {\includegraphics[width=0.75\linewidth]{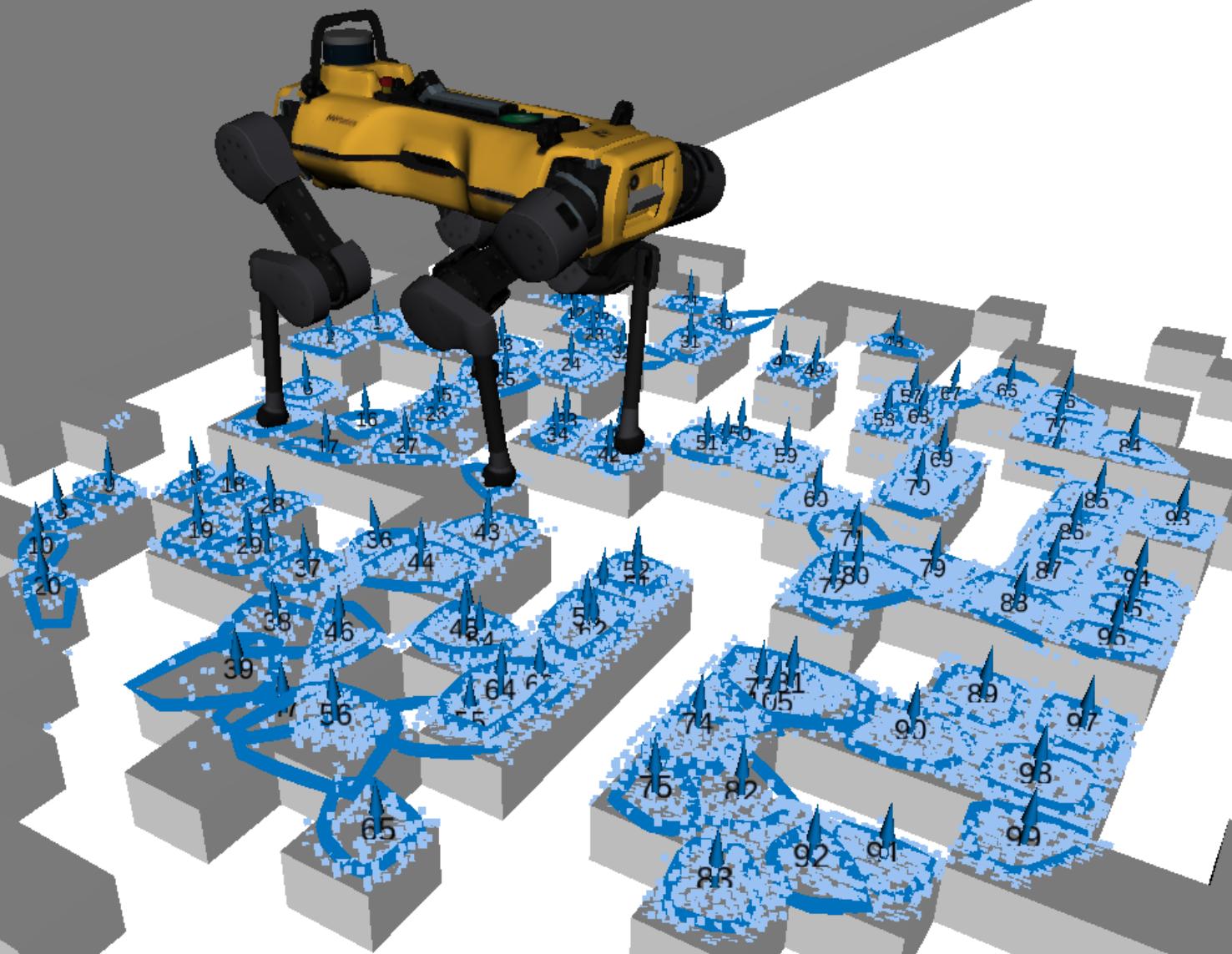}};

\draw[notip, black] ($(steppability_model.east)$) -- ($(steppability_model.east) + (0.25, 0.0)$);

\draw[notip, black] ($(steppability_model.east) + (0.25, 0.0)$) -- ($(steppability_model.east) + (0.25, 0.50)$);

\draw[newtip, black] ($(steppability_model.east) + (0.25, 0.50)$) -- ($(legged_egocan.west) + (0.0, -0.25)$);

\draw[notip, black] ($(depth_image_symbol.east) + (0.10, 0.0)$) -- ($(depth_image_symbol.east) + (1.25, 0.0)$);

\draw[notip, black] ($(depth_image_symbol.east) + (1.25, 0.0)$) -- ($(depth_image_symbol.east) + (1.25, -2.25)$);

\draw[newtip, black] ($(depth_image_symbol.east) + (1.25, -2.25)$) -- ($(legged_egocan.west)$);

\draw[notip, black] ($(normal_estimation.east)$) -- ($(normal_estimation.east) + (0.75, 0.0)$);

\draw[notip, black] ($(normal_estimation.east) + (0.75, 0.0)$) -- ($(normal_estimation.east) + (0.75, -3.5)$);

\draw[newtip, black] ($(normal_estimation.east) + (0.75, -3.5)$) -- ($(legged_egocan.west) + (0.0, 0.25)$);

\draw[newtip, black] ($(legged_egocan.center) + (0.0, 2.0)$) -- ($(superpixels.south)$);


\node[block, blueish, anchor=center, text width=7em] (torso_path_planner) at ($(depth_image_symbol.center) + (7.5, 2.35)$) {\centering Guiding Torso Path Planner (Section \ref{sec:torso_search})};

\node[anchor=center, text width=7em] (torso_path_planner_image) at ($(torso_path_planner.center) + (0.125, 1.40)$) {\includegraphics[width=0.875\linewidth]{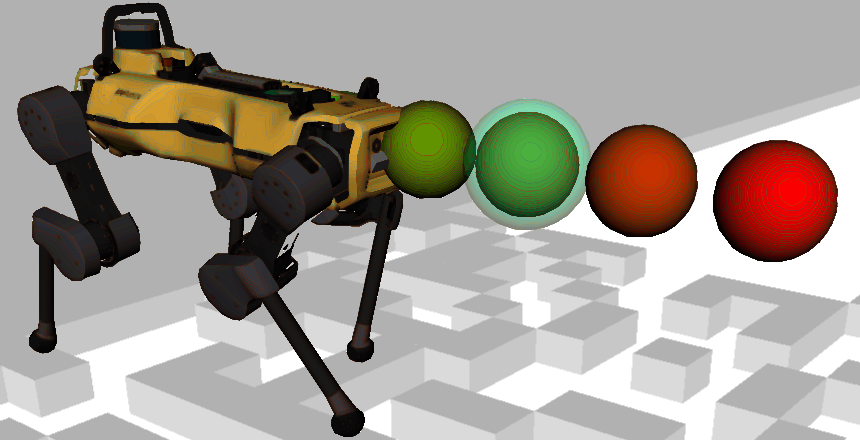}};

\node[contribBlock, blueish, anchor=center, text width=7em] (graph_search) at ($(depth_image_symbol.center) + (7.5, -1.0)$) {\centering Contact Mode Graph Search (Section \ref{sec:graph_search}) };

\node[anchor=center, text width=6em] (graph_search_image) at ($(graph_search.center) + (0.0, 1.50)$) {\includegraphics[width=\linewidth]{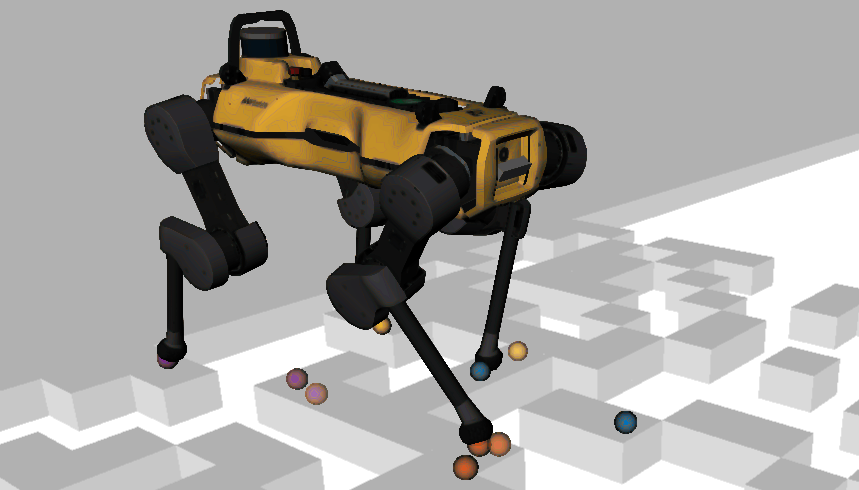}};

\node[anchor=center] (graph_search_symbol) at ($(graph_search.center) + (0.50, -1.0)$ ) {\centering $\mathcal{T}_{\rm feet}$};

\node[block, blueish, anchor=center, text width=7em] (traj_opt) at ($(depth_image_symbol.center) + (7.5, -4.0)$) {\centering Whole Body TO (Section \ref{sec:traj_opt}) };

\node[anchor=center, text width=6em] (traj_opt_image) at ($(traj_opt.center) + (0.125, 1.125)$) {\includegraphics[width=0.80\linewidth]{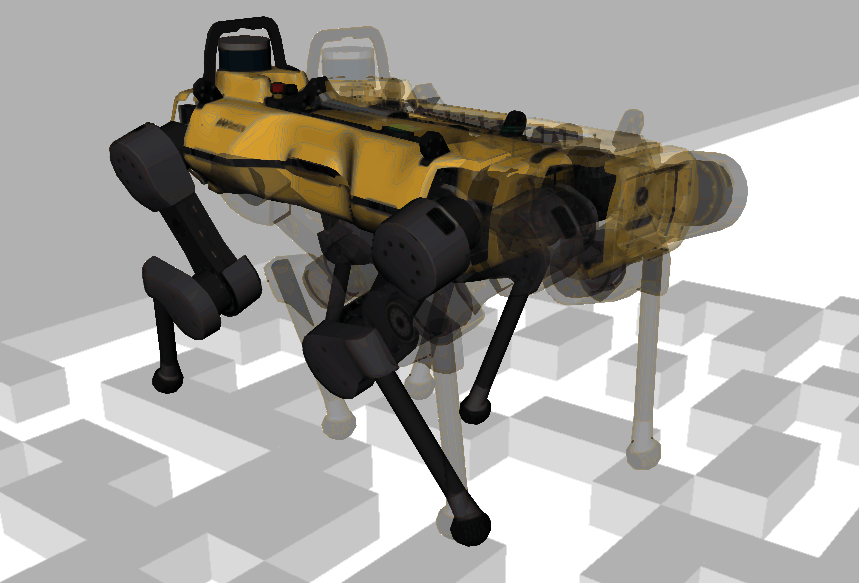}};

\node[anchor=center] (torso_path_symbol) at ($(torso_path_planner) + (0.50, -0.95)$ ) {\centering $\mathcal{T}_{\rm torso}$};

\draw[notip, black] ($(superpixels.east)$) -- ($(superpixels.east) + (0.25, 0.0)$);

\draw[notip, black] ($(superpixels.east) + (0.25, 0.0)$) -- ($(superpixels.east) + (0.25, 1.1)$);

\draw[newtip, black] ($(superpixels.east) + (0.25, 1.1)$) -- ($(torso_path_planner.west)$);

\draw[newtip, black] ($(torso_path_planner.south)$) -- ($(torso_path_planner.south) + (0.0, -0.50)$);

\draw[newtip, black] ($(graph_search.south)$) -- ($(graph_search.south) + (0.0, -0.50)$);


\node[block, greenish, anchor=center, text width=7em] (mpc) at ($(depth_image_symbol.center) + (12.0, 2.0)$) {\centering MPC \\(Section \ref{sec:mpc})};

\node[anchor=center] (ref_traj_symbol) at ($(mpc.center) + (-2.25, 0.375)$ ) {\centering $\mathbf{x}^{\rm ref}, \mathbf{u}^{\rm ref}$};

\node[anchor=center, text width=6em] (mpc_image) at ($(mpc.center) + (0.125, 1.175)$) {\includegraphics[width=0.875\linewidth]{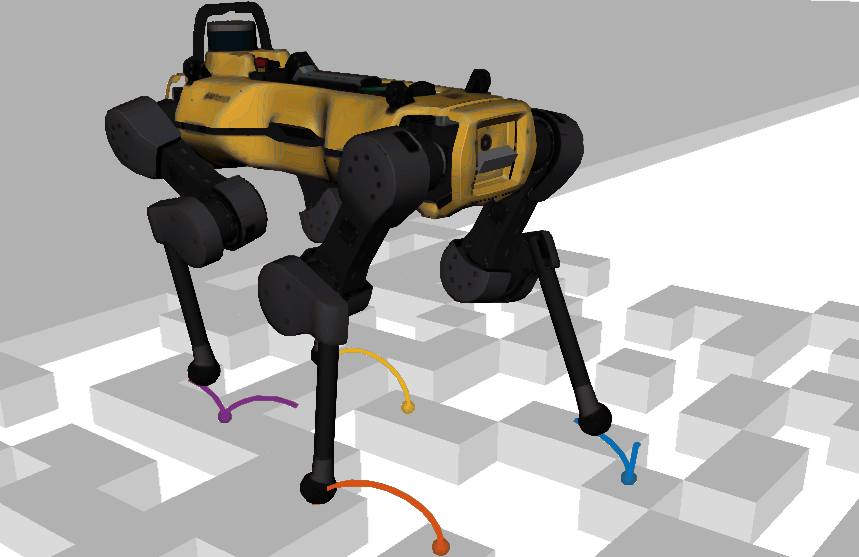}};

\node[anchor=center] (opt_state_input_symbol) at ($(mpc.center) + (0.625, -0.75)$ ) {\centering $\mathbf{x}^{*}, \mathbf{u}^{*}$};

\node[block, greenish, anchor=center, text width=7em] (wbc) at ($(depth_image_symbol.center) + (12.0, -1.0)$) {\centering WBC \\(Section \ref{sec:wbc})};

\node[anchor=center, text width=6em] (wbc_image) at ($(wbc.center) + (0.125, 1.175)$) {\includegraphics[width=0.875\linewidth]{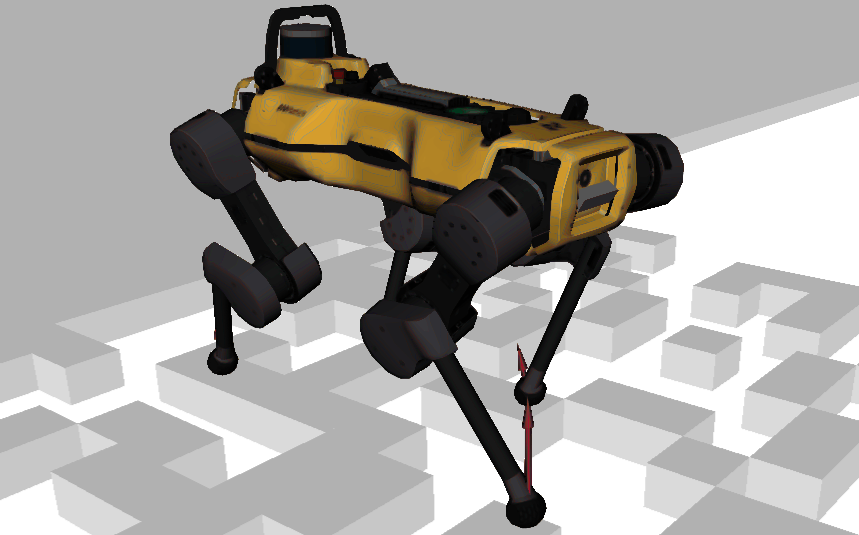} };

\node[anchor=center] (torque_symbol) at ($(wbc.center) + (0.50, -0.75)$ ) {\centering $\boldsymbol{\tau}$};

\node[anchor=center, text width=6em] (anymal) at ($(depth_image_symbol.center) + (12.25, -3.25)$) {\includegraphics[width=0.75\linewidth]{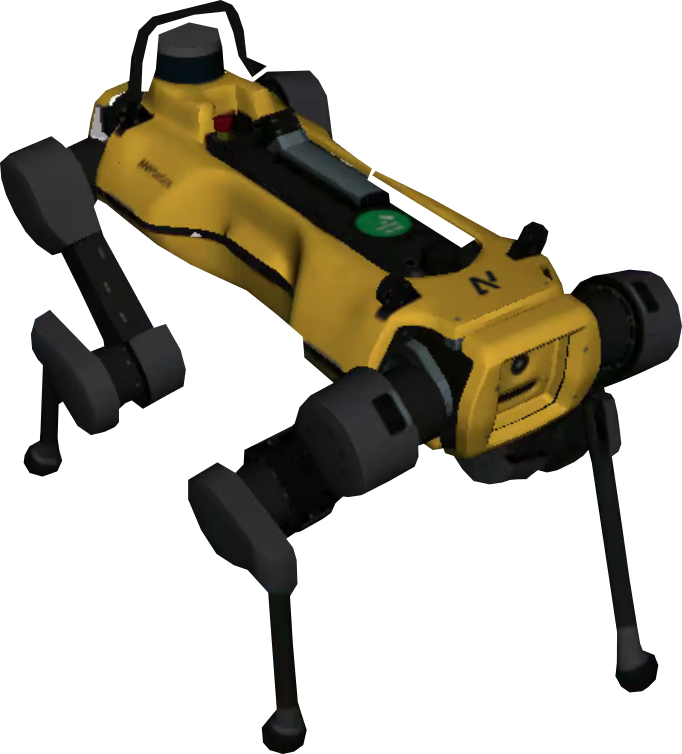}};

\draw[notip, black] ($(traj_opt.east)$) -- ($(traj_opt.east) + (0.5, 0.0)$);

\draw[notip, black] ($(traj_opt.east) + (0.5, 0.0)$) -- ($(traj_opt.east) + (0.5, 6.0)$);

\draw[newtip, black] ($(traj_opt.east) + (0.5, 6.0)$) -- ($(mpc.west)$);

\draw[newtip, black] ($(mpc.south)$) -- ($(mpc.south) + (0.0, -0.75)$);

\draw[newtip, black] ($(wbc.south)$) -- ($(wbc.south) + (0.0, -0.75)$);

\draw[notip, black] ($(anymal.center) + (-0.25, -1.0)$)  -- ($(anymal.center) + (-0.25, -1.50)$);

\draw[notip, black] ($(anymal.center) + (-0.25, -1.50)$)  -- ($(anymal.center) + (-8.25, -1.50)$);

\draw[newtip, black] ($(anymal.center) + (-8.25, -1.50)$)  -- ($(legged_egocan.south)$);

\node[anchor=center] (state_symbol) at ($(anymal.center) + (-2.0, -1.25)$ ) {\centering $\mathbf{x}$};

\end{tikzpicture}

%% file: text_files/preliminaries.tex
\section{Preliminaries} \label{sec:preliminaries}
\subsection{Egocan Representation}

\subsubsection{Egocylinder}

The concept of the egocylinder will be briefly covered in this section. Readers interested in further details are referred to the existing literature \cite{matthies_stereo_2014, smith_real-time_2020}.

\begin{figure}[h!]
    \centering
    \includegraphics[width=0.99\linewidth]{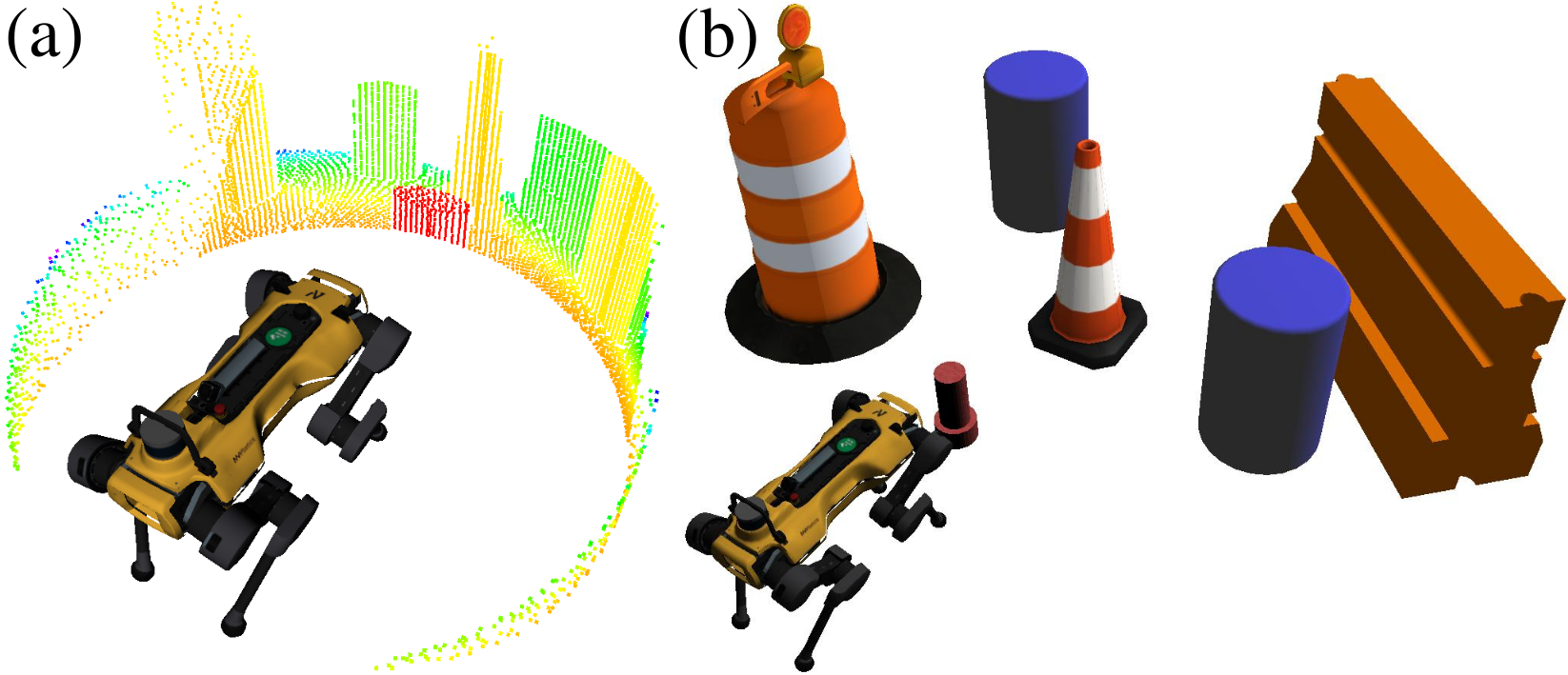}
    \caption{(a) Example egocylinder and (b) its respective scene. Proximity is represented from near to far as red to blue. The missing parts of the egocylinder are due to depth sensing range limits. }
    \label{fig:egocylinder}
\end{figure}

The concept of the egocylinder is rooted in the PiPS paradigm. LIDARs can provide full $360\degree$ views of the environment, but this comes at the cost having to process an unordered, dense point cloud. Traditional depth cameras can only provide a limited field of view of roughly $60\degree - 90\degree$ horizontally. This view greatly limits the horizon of planning that one can perform with a single depth image. Additionally, for a task such as perceptive locomotion where the environment is being revealed to the robot in real time, it is important to store prior views as the robot translates and rotates throughout the environment. To address this need, the egocylinder takes world points from a depth image sensor and projects them onto a virtual cylinder that surrounds the robot. Points on the cylinder are discretized into a panoramic image to leverage the favorable aspects of the image space representation such as parallelization. An example egocylinder image can be seen in Figure \ref{fig:egocylinder}.

The egocylinder takes as input a depth image $I_{\rm depth} \in \mathbb{R}^{w_{\rm cam} \times h_{\rm cam} \times 1}$ of width $w_{\rm cam}$ and height $h_{\rm cam}$. Each pixel $\mathbf{r}_{\rm cam} = \begin{bmatrix} u_{\rm cam} & v_{\rm cam} \end{bmatrix}^T \in \mathbb{R}^2$ stored in $I_{\rm depth}$ can be parameterized in the camera frame as $\mathbf{p}_{\rm cam} = \begin{bmatrix} x_{\rm cam} &  y_{\rm cam} & z_{\rm cam} \end{bmatrix}^T \in \mathbb{R}^3$ where $z_{\rm cam}$ is the depth value read off of $I_{\rm depth}$. The point $\mathbf{p}_{\rm cam}$ can be computed with the focal length $f_{\rm cam}$ and central pixel $\mathbf{r}_{{\rm cam}, 0} = \begin{bmatrix} u_{{\rm cam}, 0} & v_{{\rm cam}, 0} \end{bmatrix}^T\in \mathbb{R}^2$.

New points from incoming sensor streams are projected onto the virtual cylinder that surrounds the ego-robot. This cylinder is propagated forward in time as the robot moves. This cylindrical surface is discretized into an image $I_{\rm cyl} \in \mathbb{R}^{w_{\rm cyl} \times h_{\rm cyl} \times 1}$ of width $w_{\rm cyl}$ and height $h_{\rm cyl}$ which stores the egocylindrical range of each point. This range allows the egocylinder to retain information behind the robot. Therefore, incoming Cartesian camera frame information is transformed into egocylindrical coordinates via
\begin{equation}
    \mathbf{p}_{\rm cyl} = \begin{bmatrix}
        \rho_{\rm cyl} \\ \theta_{\rm cyl} \\ z_{\rm cyl}
    \end{bmatrix}
    =
    T_{\rm c2e}(\mathbf{p}_{\rm cam}) = \begin{bmatrix}
        \sqrt{x_{\rm cam}^2 + y_{\rm cam}^2} \\ 
        \text{Arg}(x_{\rm cam} + \mathbf{j} z_{\rm cam}) \\ 
        y_{\rm cam}
    \end{bmatrix}
\end{equation}
where $\mathbf{p}_{\rm cyl} \in \mathbb{R}^3$ is the egocylindrical coordinate, $\rho_{\rm cyl}$ represents the range, $\theta_{\rm cyl}$ is the angle, and $z_{\rm cyl}$ is the height. The matrix $T_{\rm c2e} \in SE(3)$ transforms from Cartesian to egocylindrical coordinates. Resulting egocylindrical coordinates are then mapped to image coordinates $\mathbf{r}_{\rm cyl} = \begin{bmatrix} u_{\rm cyl} & v_{\rm cyl} \end{bmatrix}^T \in \mathbb{R}^2$ using the egocylinder projection matrix $K_{\rm cyl}\in \mathbb{R}^{2\times3}$:
\begin{equation}
    \mathbf{r}_{\rm cyl} = K_{\rm cyl} \begin{bmatrix}
        \theta_{\rm cyl} \\ z_{\rm cyl} \\ 1
    \end{bmatrix} \textrm{ with } K_{\rm cyl} = \begin{bmatrix}
        f_{\rm cyl} & 0 & u_{{\rm cyl}, 0} \\
        0 & f_{\rm cyl} & v_{{\rm cyl}, 0}
    \end{bmatrix}
\end{equation}
where $f_{\rm cyl} = \cot{(\frac{2 \pi}{w_{\rm cyl}} )}$, $u_{{\rm cyl}, 0} = \frac{w_{\rm cyl}}{ 2}$, and $v_{{\rm cyl}, 0} = \frac{h_{\rm cyl}}{2}$.

Existing egocylindrical coordinates are mapped back to Cartesian coordinates, propagated forward in time under a Euclidean transform for the induced pose $ T_{\rm move} \in SE(3)$ between the prior and current timesteps, and mapped back to egocylindrical coordinates as
\begin{equation}
\mathbf{p}_{\rm cyl}' = T_{\rm c2e} \circ T_{\rm move} \circ T_{\rm e2c}(\mathbf{p}_{\rm cyl}) \textrm{ with } T_{\rm e2c} = \begin{bmatrix}
        \rho_{\rm cyl} \cos(\theta_{\rm cyl}) \\ z_{\rm cyl} \\ \rho_{\rm cyl} \sin(\theta_{\rm cyl}) \end{bmatrix}
\end{equation}
where $T_{\rm e2c} \in SE(3)$ transforms from egocylindrical coordinates to Cartesian coordinates.

\subsubsection{Egocan}
\begin{figure}[h!]
    \centering
    \includegraphics[width=0.99\linewidth]{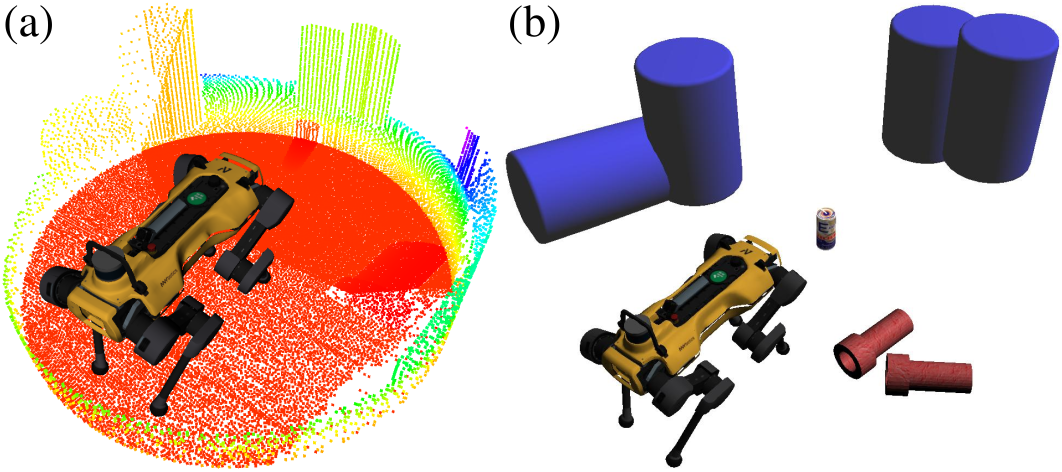}
    \caption{(a) Example egocan and (b) its respective scene. This representation captures nearby ground elevation information including the can in front of the robot which is otherwise missed by the egocylinder.}
    \label{fig:egocan}
\end{figure}

During egocylindrical propagation, points that do not map onto the panoramic image are lost. This means that points that pass over or under the cylinder are not retained. Note that in Figure \ref{fig:egocylinder}, a portion of the short red rod in front of the robot is \textit{not} captured on the egocylinder as it passes underneath the egocylinder field of view. To address this need, the egocan extends the egocylinder by adding upper and lower \textit{egocaps} on the top and bottom of the egocylindrical image to create a closed surface. The egocan was introduced for aerial vehicles \cite{smith_aerialpips_2023}, but has yet to see use in locomotion. For this work, the lower cap, visualized in Figure \ref{fig:egocan}, is used to capture prior scenes as they travel underfoot, which are later used for footstep planning. Functionally, the egocan floor surface acts much like a height map. However, no expensive point cloud processing is required to build it. Furthermore, the egocan floor information is maintained as an image which allows QuadPiPS to leverage the benefits of the data structure. 

Egocap surfaces map to the images $I_{\rm low}, I_{\rm up} \in \mathbb{R}^{ w_{\rm cap} \times h_{\rm cap} \times 1}$ with width $w_{\rm cap}$ and height $h_{\rm cap}$ where $w_{\rm cap} = w_{\rm low} = w_{\rm up} = h_{\rm cap} = h_{\rm low} = h_{\rm up}$. A new or existing point $\mathbf{p}_{\rm cyl}$ that does not map onto $I_{\rm cyl}$ is mapped onto $I_{\rm low}$ if $z_{\rm cyl} > 0$ and $I_{\rm up}$ otherwise. Throughout this work, $I_{\rm low}$ may be referred to as the egocan floor image. The upper image $I_{\rm up}$ is not used in this work and will no longer be addressed for clarity. The height information $z_{\rm cyl}$ will be placed at the lower image coordinate $\mathbf{r}_{\rm low}= \begin{bmatrix} u_{\rm low} & v_{\rm low} \end{bmatrix}^T \in \mathbb{R}^2$ where
\begin{equation}
    \mathbf{r}_{\rm low} = K_{\rm low} \begin{bmatrix}
       \frac{x_{\rm cam}}{|y_{\rm cam}|} \\ \frac{z_{\rm cam}}{|y_{\rm cam}|} \\ 1
    \end{bmatrix} \textrm{ with } K_{\rm low} = \begin{bmatrix}
        f_{\rm low} & 0 & u_{{\rm low}, 0} \\
        0 & f_{\rm low} & v_{{\rm low}, 0}
    \end{bmatrix}
\end{equation}
and the lower cap focal length $f_{\rm low} = \frac{v_{\rm fov} w_{\rm low}}{4}$ where $v_{\rm fov}$ is the vertical field of view of the egocylinder which is a user-specified parameter between $45\degree$ and $90\degree$. The lower cap center coordinate is expressed as $\mathbf{r}_{{\rm low}, 0}= \begin{bmatrix} u_{{\rm low}, 0} & v_{{\rm low}, 0} \end{bmatrix}^T \in \mathbb{R}^2$ where $u_{{\rm low}, 0} = \frac{w_{\rm low}}{2}$ and $v_{{\rm low}, 0} = \frac{h_{\rm low}}{2}$.

%% file: text_files/perception.tex
\section{Legged Egocan}
\label{sec:perception}

\subsection{Image Space Normal Estimation} \label{sec:normal_estimation}
For the task of perceptive locomotion over complex terrain, additional environmental affordances are required beyond just depth. The QuadPiPS framework adopts the depth image gradients-based approach \cite{nakagawa_estimating_2015} for surface normal estimation in the image space. The approach is briefly covered here, and more information can be found in the original work. The subscript $(\cdot)_{\rm cam}$ for the camera frame will be temporarily dropped in this section for clarity. 

\begin{figure}[h!]
    \centering
    \includegraphics[width=0.99\linewidth]{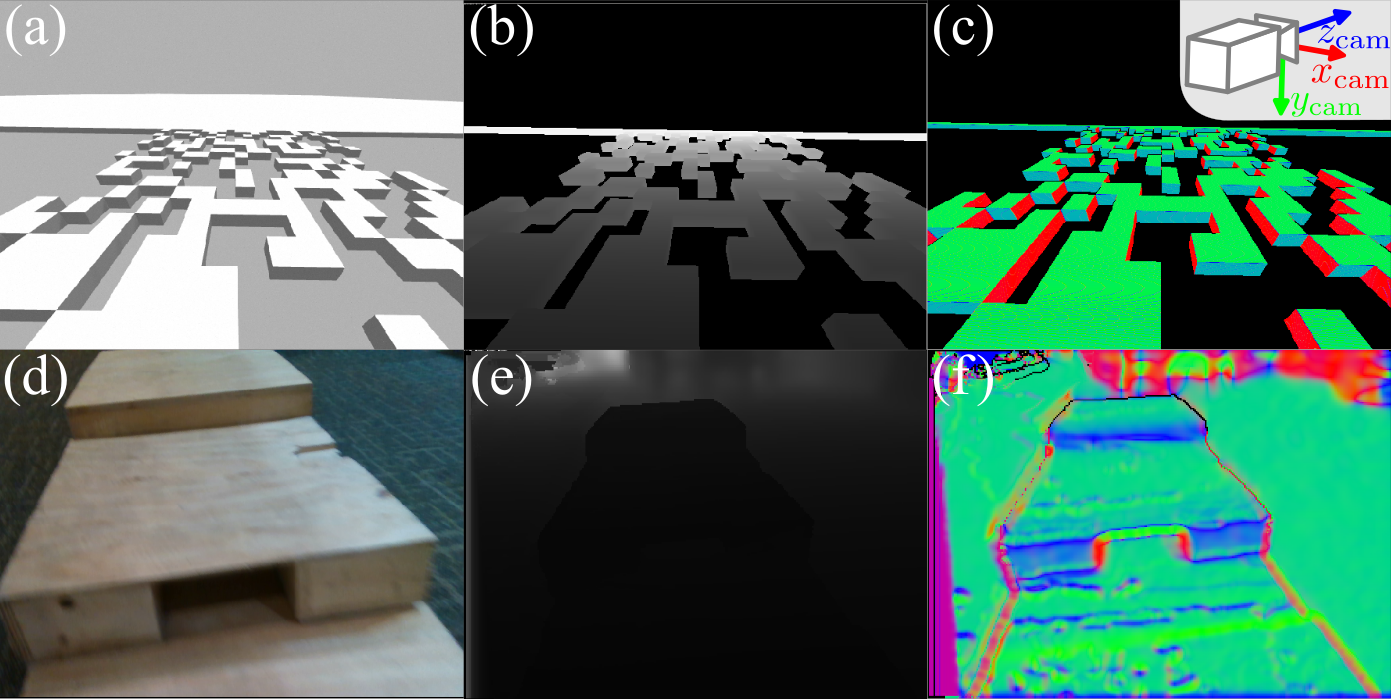}
    \caption{Example scenes with their estimated image space normals. The top row is an example scene in simulation and the bottom row is an example scene on hardware. The left images (a) and (d) are color images of the two scenes, the middle images (b) and (e) are depth images, and the right images (c) and (f) are the estimated normals. The RGB color scheme corresponds to the $xyz$ axes of the camera frame where the absolute value of the estimated surface normal vector components are used to calculate colors. The region of purple pixels in (f) are due to artifacts from a hole-filling filter. 
    }
    \label{fig:normals_figure}
\end{figure}

This normal estimation technique relies on building a local planar approximation at the point $\mathbf{p}(u,v)$ using its directional derivatives $\mathbf{p}_u(u,v)$ and $\mathbf{p}_v(u,v)$. These directional derivatives can be computed as 
\begin{equation}
    \begin{split}
        \mathbf{p}_u(u, v) & = \Big(\frac{\partial x(u, v)}{\partial u}, \frac{\partial y(u, v)}{\partial u}, \frac{\partial z(u, v)}{\partial u}\Big) \\
        \mathbf{p}_v(u, v) & = \Big(\frac{\partial x(u, v)}{\partial v}, \frac{\partial y(u, v)}{\partial v}, \frac{\partial z(u, v)}{\partial v}\Big).
    \end{split}
\end{equation}
The partial derivative terms can then be calculated as
\begin{equation}
    \begin{split}
        \frac{\partial x(u, v)}{\partial u} & = \frac{z(u, v)}{f} + \frac{(u - u_0)}{f}\frac{\partial z(u, v)}{\delta u} \\
        \frac{\partial y(u, v)}{\partial u} & = \frac{(y - y_0)}{f}\frac{\partial z(u, v)}{\delta u} \\
        \frac{\partial x(u, v)}{\partial v} & = \frac{(u - u_0)}{f}\frac{\partial z(u, v)}{\delta v} \\
        \frac{\partial y(u, v)}{\partial v} & = \frac{z(u, v)}{f} + \frac{(y - y_0)}{f}\frac{\partial z(u, v)}{\delta v},
    \end{split}
\end{equation}
with
\begin{equation}
    \begin{split}
        \frac{\partial z(u, v)}{\delta u} & \approx z(u+1, v) - z(u, v) \\
        \frac{\partial z(u, v)}{\delta v} & \approx z(v+1) - z(u, v).
    \end{split}
\end{equation}
Lastly, the directional cross product can be calculated as 
\begin{equation}
    \mathbf{n}(u, v) = \mathbf{p}_v(u, v) \times \mathbf{p}_u(u, v)
\end{equation}
and subsequently normalized:
\begin{equation}
    \mathbf{\hat{n}}(u, v) = \frac{\mathbf{n}(u, v)}{\| \mathbf{n}(u, v) \|_2}.
\end{equation}
These pixel-wise normals comprise $I_{\rm normal} \in \mathbb{R}^{w_{\rm cam} \times h_{\rm cam} \times 3}$. Example images and normals are shown in Figure \ref{fig:normals_figure}.

\subsection{Steppability Segmentation} \label{sec:steppability}
Depth and normal information make up the geometric affordances present in the legged egocan representation. However, it is also important to be able to semantically reason about local terrain to encode preferences or behaviors that cannot be ascertained through geometry alone. For foothold planning, semantic labels for preferring to step on, over, or around objects can help guide the planner away from regions that are geometrically permissible, but not preferable.  

In this section, the process in which steppability labels are predicted for the local environment is detailed. First, the simulation scene creation process used to generate synthetic steppability data is discussed. Here, QuadPiPS adapts a primitive shapes-based technique that was previously used for manipulation tasks \cite{nieuwenhuisen_shape-primitive_2012, lin_using_2020}. Primitive shapes-based approaches hypothesize that the geometry of graspable objects can be decomposed into a set of primitive shapes where each primitive shape class has a particular family of effective grasps. For data collection, the ground truth labels of these objects can then be ascertained through the color of the primitives within the simulation scene. QuadPiPS applies this ideology to generate class labels for steppability. The synthetic scenes used for training data are assembled according to a key set of design parameters that are detailed below. 

\subsubsection{Primitive Shape Classes}
\begin{figure}[h!]
  \centering
  \includegraphics[width=0.90\linewidth]{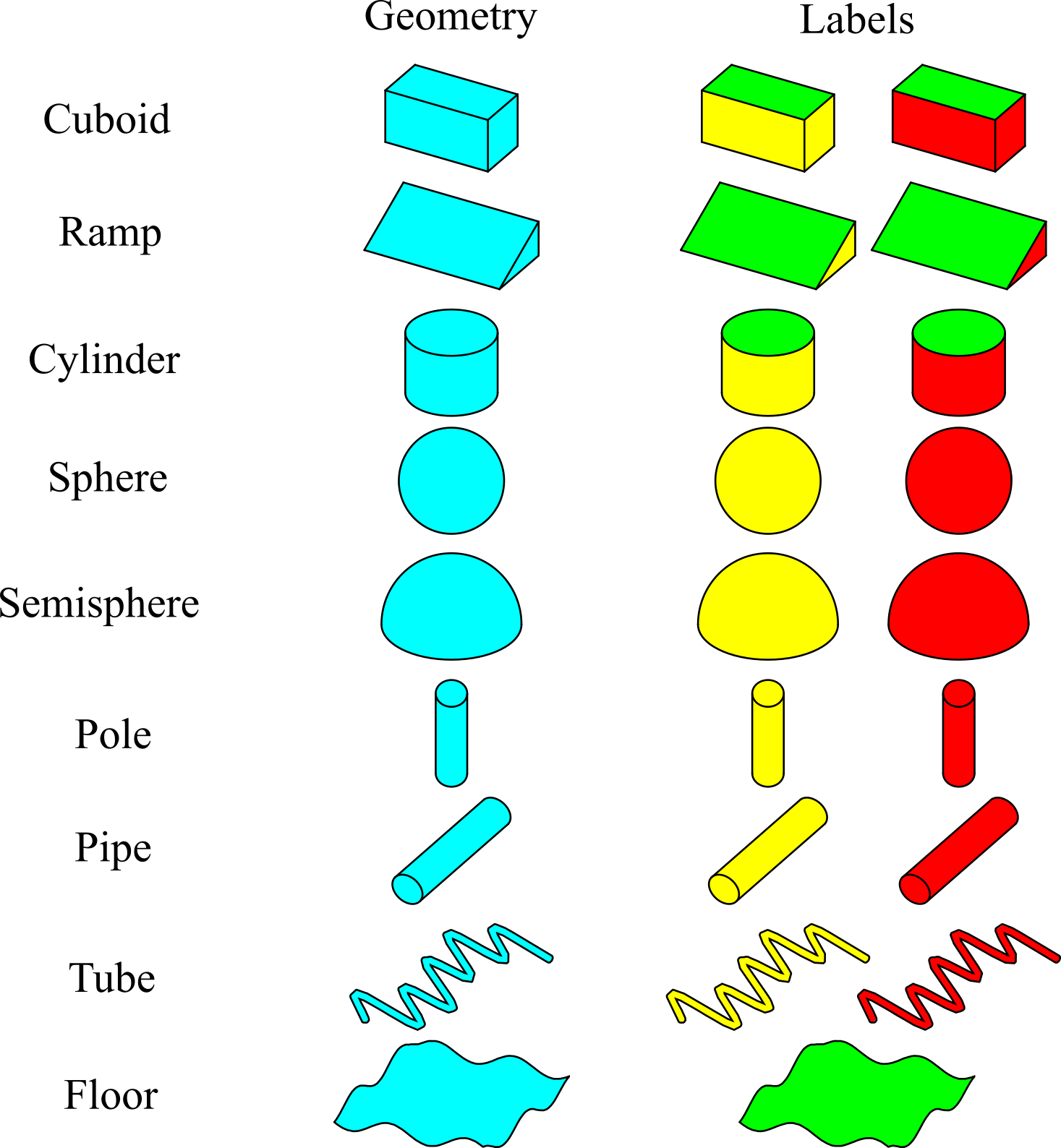}
  \caption{Example visualizations of the primitive shape classes used to construct the simulation scenes that comprise the synthetic data for training the steppability policy. Green corresponds to steppable, yellow corresponds to passable, and red corresponds to non-passable.}
  \label{fig:primitive_shape}
\end{figure}

Nine primitive shape classes are used to build scenes: \textit{Cuboid}, \textit{Ramp}, \textit{Cylinder}, \textit{Sphere}, \textit{Semisphere}, \textit{Pole}, \textit{Pipe}, \textit{Tube}, and \textit{Floor}. The Cuboid and Ramp classes are parameterized by length $l_{\rm prim}$, width $w_{\rm prim}$, and height $h_{\rm prim}$, the Cylinder class is parameterized by radius $r_{\rm prim}$ and height $h_{\rm prim}$, the Sphere and Semisphere classes are parameterized by radius $r_{\rm prim}$, and the Pipe, Pole, and Tube classes are parameterized by length $l_{\rm prim}$ and radius $r_{\rm prim}$. The floor class is non-parametric in that all of its instances are 4~m $\times$ 4~m with a set of 3D surface equations to capture non-flat terrain. Each shape class is shown in Figure \ref{fig:primitive_shape}.

\begin{figure*}[t]
  \centering
  \includegraphics[width=0.99\linewidth]{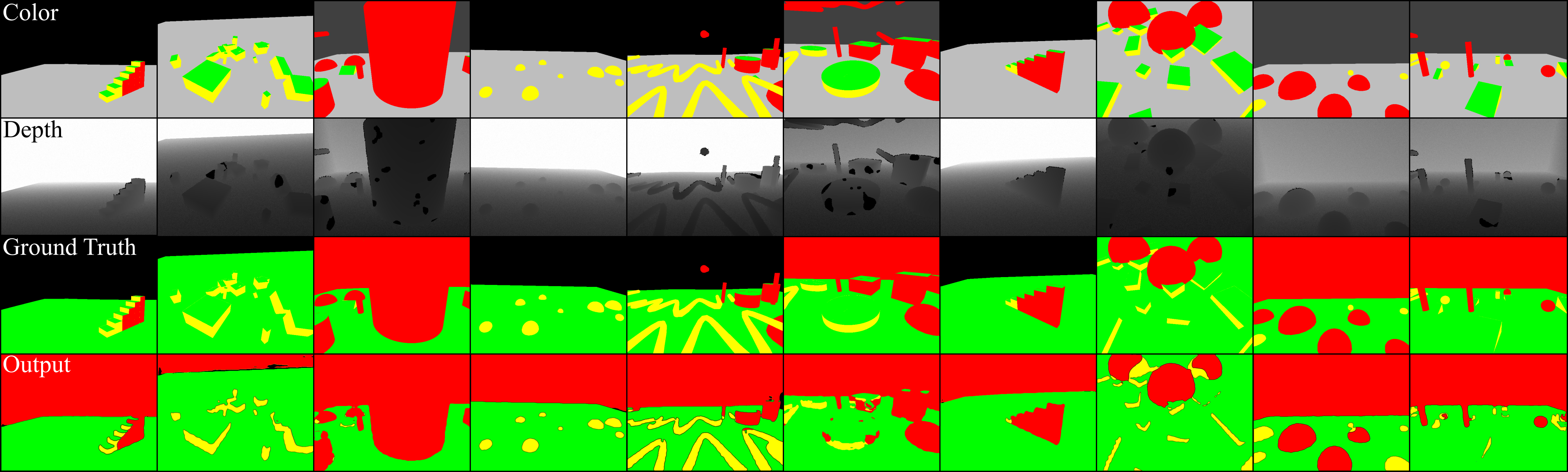}
  \caption{Example simulation scenes along with steppability model predictions. Top row: color images. Second row: depth images. These images also display the noisy artifacts injected into the training data including surface and boundary depth loss along with Gaussian blur. Third row: ground truth steppability labels of the corresponding column's input depth image. Bottom row: model outputs for the corresponding column's input depth image.}
  \label{fig:example_outputs}
\end{figure*}

\subsubsection{Primitive Shape Steppability Policies}
Each primitive shape class is assigned a mesh-based steppability policy that defines which faces of the shapes should be assigned which labels. The three labels used are:

\begin{itemize}
    \item \textbf{Steppable (Green):} can support a stable foothold
    \item \textbf{Passable (Yellow):} can \textit{not} support a stable foothold, but can be stepped over by a foot swing trajectory
    \item \textbf{Non-passable (Red):} can \textit{not} support a stable foothold and can \textit{not} be stepped over by a foot swing trajectory
\end{itemize}

For Cuboids, Ramps, and Cylinders, the top face is labeled as steppable due to its planar geometry. If the height discontinuity surrounding the primitive exceeds a maximum swing height for the robot leg, defined as $h^{\rm max}_{\rm swing} = 0.15$~m, then all vertical faces are labeled as non-passable. Otherwise, the vertical faces are labeled as passable. For Spheres, Semispheres, Poles, Pipes, and Tubes, the entire primitive is labeled as non-passable if the $z$-coordinate of the top of the primitive exceeds $h^{\rm max}_{\rm swing}$. Otherwise, the entire primitive is labeled as passable. Lastly, floors are labeled as completely steppable.

\subsubsection{Primitive Shape Pose}
The six-dimensional pose of each primitive shape in the scene frame, $\mathbf{q}_{\rm scene} \in \mathbb{R}^6$, can be set according to desired scene attributes. For instance, scenes can be set to feature clusters of primitive shapes localized within a particular region of the scene, the primitives can be scattered throughout the scene, or the poses can be overwritten to accept manually defined entries if the user wants to create more contrived scenes with staircases or stepping stones. The $z$-dimension of all shapes, $z_{\rm scene}$, is restricted so that they are placed on support surfaces outside of Pipes and Tubes, which are allowed to be suspended in air. The orientations of Cuboids, Ramps, and Cylinders are restricted to ensure that the face labeled as steppable is the top face in the scene.

\subsubsection{Camera pose}
The pose of the camera for each simulation scene can also be parameterized to emulate expected real-world circumstances for onboard sensing. Given the desired application of quadrupedal locomotion, the camera is set at a nominal height and pitch to approximate the pose of the depth camera attached to the quadrupedal hardware platform. To capture multiple frames of a single scene, the camera is set to follow a prescribed trajectory that approximates how the quadruped's torso would move through a real-world environment. Gaussian noise is also applied to all six pose dimensions to capture the jitter that the camera would experience during locomotion in the real world.

\subsubsection{Scene Environment}
Lastly, the overall synthetic environment that the primitive shapes are placed within can also be controlled. A random binary variable is sampled which chooses between indoor environments with walls and a ceiling and outdoor environments with an infinite horizon.

\subsubsection{Domain Alignment}
The process of bi-directional domain alignment is performed in which (a) simulation data is corrupted to approximate defects observed in real-world sensing, and (b) real-world sensor data is post-processed to mitigate these flaws and artifacts. 

For simulation data, a sequence of corrupting filters is applied to the depth data. Gaussian noise and an oil painting filter are applied to corrupt the depth data within dilated object boundaries. Then, randomly connected groups of pixels are voided on the boundaries and surfaces of objects to approximate depth loss. On hardware, a sequence of post-processing filters provided by the RealSense API including decimation and hole-filling \cite{grunnet-jepsen_depth_nodate} is applied to the depth data.

From each scene, a depth image $I_{\rm depth}$ is extracted along with a ground truth steppability mask $I^*_{\rm step} \in \mathbb{R}^{w_{\rm cam} \times h_{\rm cam} \times 1}$.

\subsubsection{Model Training} \label{sec:model_training}
\begin{figure}[h!]
  \centering
  \includegraphics[width=0.99\linewidth]{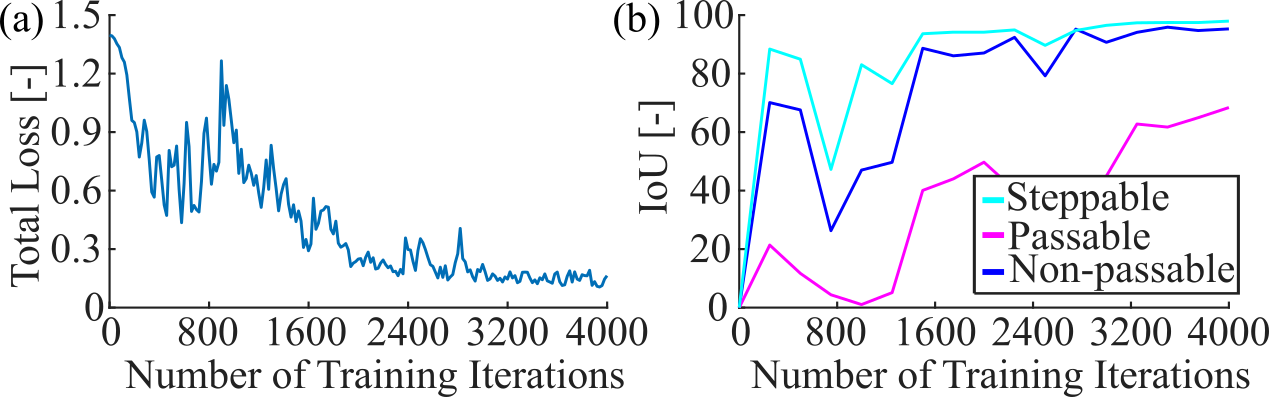}
  \caption{Results of training. (a) Total training loss over the training iterations. ~(b) Intersection-over-union (IoU) of all classes over the training iterations.}
  \label{fig:training_results}
\end{figure}

A DeepLabV3+ \cite{ferrari_encoder-decoder_2018} model from the Detectron2 deep learning library \cite{yuxin_wu_detectron2_2019} is used for steppability prediction. The dataset contains $600$ scenes with $5$ frames each. In total, dataset generation took roughly $12$ hours.
The generated data was put into 80/10/10\% splits for training, validation, and test subsets, respectively. Model training took $65$ minutes on a Dell Precision 5820 Tower with an Intel Xeon W-2223 CPU (single-thread passmark of $2,199$; multi-thread score of $8,586$) and a NVIDIA T1000 GPU. Example scene data is shown in Figure \ref{fig:example_outputs}. Training loss and intersection-over-union (IoU) plots can be seen in Figure \ref{fig:training_results}. 

The model output is represented as the predicted steppability mask $I_{\rm step} \in \mathbb{R}^{w_{\rm cam} \times h_{\rm cam} \times 1}$ in which each pixel contains a label $l(u_{\rm cam}, v_{\rm cam})$.

\subsection{Legged Egocan Representation} \label{sec:legged_egocan}
The legged egocan takes as input a multi-channel affordance image  $I_{\rm afford} = \begin{bmatrix} I_{\rm depth} & I_{\rm normal} & I_{\rm step} \end{bmatrix} \in \mathbb{R}^{w_{\rm cam} \times h_{\rm cam} \times 5}$. The egocan propagation step is kept the same, the only difference being that additional normal and steppability information is stored along with depth information. To reflect this, the floor image also takes on the dimensionality $I_{\rm low} \in \mathbb{R}^{ w_{\rm cap} \times h_{\rm cap} \times 5}$.

\subsection{Superpixels Oversegmentation} \label{sec:superpixels}

Once the egocan has been populated, it is still necessary to identify what parts of the terrain can support footholds. It is common to decompose a height map into a set of planar convex regions that can support footholds \cite{grandia_perceptive_2022}. While such regions work well for linear optimization constraints, the process for searching over these regions is unclear. If the ego-robot is presented with a large, flat floor, one large convex region may be generated. This is not conducive to a search framework. For this reason, QuadPiPS adopts an oversegmentation approach via superpixels.

Superpixel algorithms \cite{achanta_slic_2012, neubert_compact_2014} are intended to cluster pixels into perceptually homogeneous regions which are evenly distributed throughout the image space. These homogeneous regions, or superpixels, are meant to modify or replace the uniform gridding of the pixel matrix within the image. In the case of this work, superpixels act as an insightful primitive that allows the foothold graph search to not only select which terrain features to plan footholds on, but also decide where within the feature to step.

\subsubsection{Health Check}
\label{sec:health_check}
As input, this superpixel algorithm takes the egocan floor image $I_{\rm low}$. Prior to passing $I_{\rm low}$ in for oversegmentation, a health check is performed on $I_{\rm low}$. Due to the nature of the egocan propagation, the floor image is sparse. Normally, superpixel algorithms are deployed on full RGB or RGB-D \cite{weikersdorfer_depth-adaptive_2012, weikersdorfer_depth-adaptive_2013} images. However, in this use case, a majority of the pixels lack data. First, pixels with no data or invalid data are set to zero. Pixels with unstable normals, defined as $\mathbf{\hat{n}_{\rm cam}}(u_{\rm low},v_{\rm low}) \cdot - \mathbf{\hat{e}}_{y_{\rm cam}} \leq 0.5$, are also set to zero to avoid building planar regions with them. The term $\mathbf{\hat{e}}_{y_{\rm cam}}$ represents the basis vector for the $y$-dimension of the camera frame. Pixels with labels $l(u_{\rm low}, v_{\rm low})$ that are not assigned steppable are also removed.

\subsubsection{Image Preprocessing}
\label{sec:image_preprocessing}
When the egocan is initialized, $I_{\rm low}$ is empty. However, this yields no planar regions. QuadPiPS assumes that planning begins on flat, obstacle-free terrain. Therefore, regions of the floor image that have not yet been populated by sensor data are initialized to default heights and normals that represent standard ground floor terrain. 

\subsubsection{Superpixels Approach}
\label{sec:superpixels_approach}
The oversegmentation algorithm can be understood as a modified version of the Simple Linear Iterative Clustering (SLIC) \cite{achanta_slic_2012} algorithm that is designed to handle sparse multi-channel images.
The initial set of superpixel cluster centers $\mathcal{S}$ are defined by uniformly sampling pixels in $I_{\rm low}$ at regular grid steps $s_{\rm low} = \sqrt{\frac{n_{\rm low}}{k_{\rm des}}}$. The value $n_{\rm low}=w_{\rm low} \times h_{\rm low}$ is the number of pixels in the floor image and $k_{\rm des}$ is the user-defined desired number of superpixels. Each cluster $\zeta \in \mathcal{S}$ contains a pixel $\mathbf{r}_\zeta = \begin{bmatrix} u_\zeta & v_\zeta \end{bmatrix}^T \in \mathbb{R}^2$, a point $\mathbf{p}_{\rm cam}(u_\zeta, v_\zeta) \in \mathbb{R}^3$ which is abbreviated as $\mathbf{p}_\zeta$, and a normal $\mathbf{\hat{n}}_{\rm cam}(u_\zeta, v_\zeta) \in \mathbb{R}^3$ which is abbreviated as $\mathbf{\hat{n}}_\zeta$. The initial clusters undergo a refinement step in which they are moved to the centroid of the $\frac{s_{\rm low}}{4} \times \frac{s_{\rm low}}{4}$ neighborhood surrounding the initial cluster center. This helps in generating smoother superpixel regions.

Then, the proposed algorithm iterates through the set of clusters and calculates the distance between the current cluster $\zeta$ and the current pixel $\mathbf{r}_{\rm low}$. Only pixels in a $2s_{\rm low} \times 2s_{\rm low}$ neighborhood are evaluated. The distance metric scores deviation from the local plane formed by the cluster $\zeta$ as well as cluster compactness. For $\zeta$ and $\mathbf{r}_{\rm low}$, the complete distance calculation is
\begin{equation}
\begin{split}
    d_{\rm total}(\zeta, \mathbf{r}_{\rm low}) & =  w_n \frac{d_n(\zeta, \mathbf{r}_{\rm low})}{d_n^{\rm max}} + w_p \frac{d_p(\zeta, \mathbf{r}_{\rm low})}{d_p^{\rm max}} \\
                        & + w_w \frac{d_w(\zeta, \mathbf{r}_{\rm low})}{d_w^{\rm max}} + w_c \frac{d_c(\zeta, \mathbf{r}_{\rm low})}{d_c^{\rm max}}.
\end{split}
\end{equation}
The term
\begin{equation}
    d_n(\zeta, \mathbf{r}_{\rm low}) = 1 - \mathbf{\hat{n}}_{\zeta} \cdot \mathbf{\hat{n}}_{\rm cam}(u_{\rm low}, v_{\rm low})
\end{equation}
measures deviation from the cluster normal $\mathbf{\hat{n}}_\zeta$ and the current pixel's normal $\mathbf{\hat{n}}_{\rm cam}(u_{\rm low}, v_{\rm low})$. The term
\begin{equation}
    d_p(\zeta, \mathbf{r}_{\rm low}) = | (\mathbf{p}_{\zeta} - \mathbf{p}_{\rm cam}(u_{\rm low}, v_{\rm low})) \cdot \mathbf{\hat{n}}_{\zeta} |
\end{equation}
measures distance between the current pixel's camera frame position $\mathbf{p}_{\rm cam}(u_{\rm low}, v_{\rm low})$ and the cluster plane $(\mathbf{p}_\zeta, \mathbf{\hat{n}}_{\zeta})$. The term
\begin{equation}
    d_w(\zeta, \mathbf{r}_{\rm low}) = \| \mathbf{p}_{\zeta} - \mathbf{p}_{\rm cam}(u_{\rm low}, v_{\rm low}) \|_2,
\end{equation}
measures Euclidean distance between the current point and the cluster center, and the final term
\begin{equation}
    d_c(\zeta, \mathbf{r}_{\rm low}) = \sqrt{ (u_\zeta - u_{\rm lower})^2 + (v_\zeta - v_{\rm lower})^2}
\end{equation}
measures Euclidean pixel distance between the current pixel and the cluster center. The values $w_{(\cdot)}$ are positive scalar values for weighting importance. The notation $d^{\rm max}_{(\cdot)}$ represents the maximum distance for the respective terms. These are applied to scale the distance terms.

\begin{figure}[h!]
  \centering
  \includegraphics[width=0.99\linewidth]{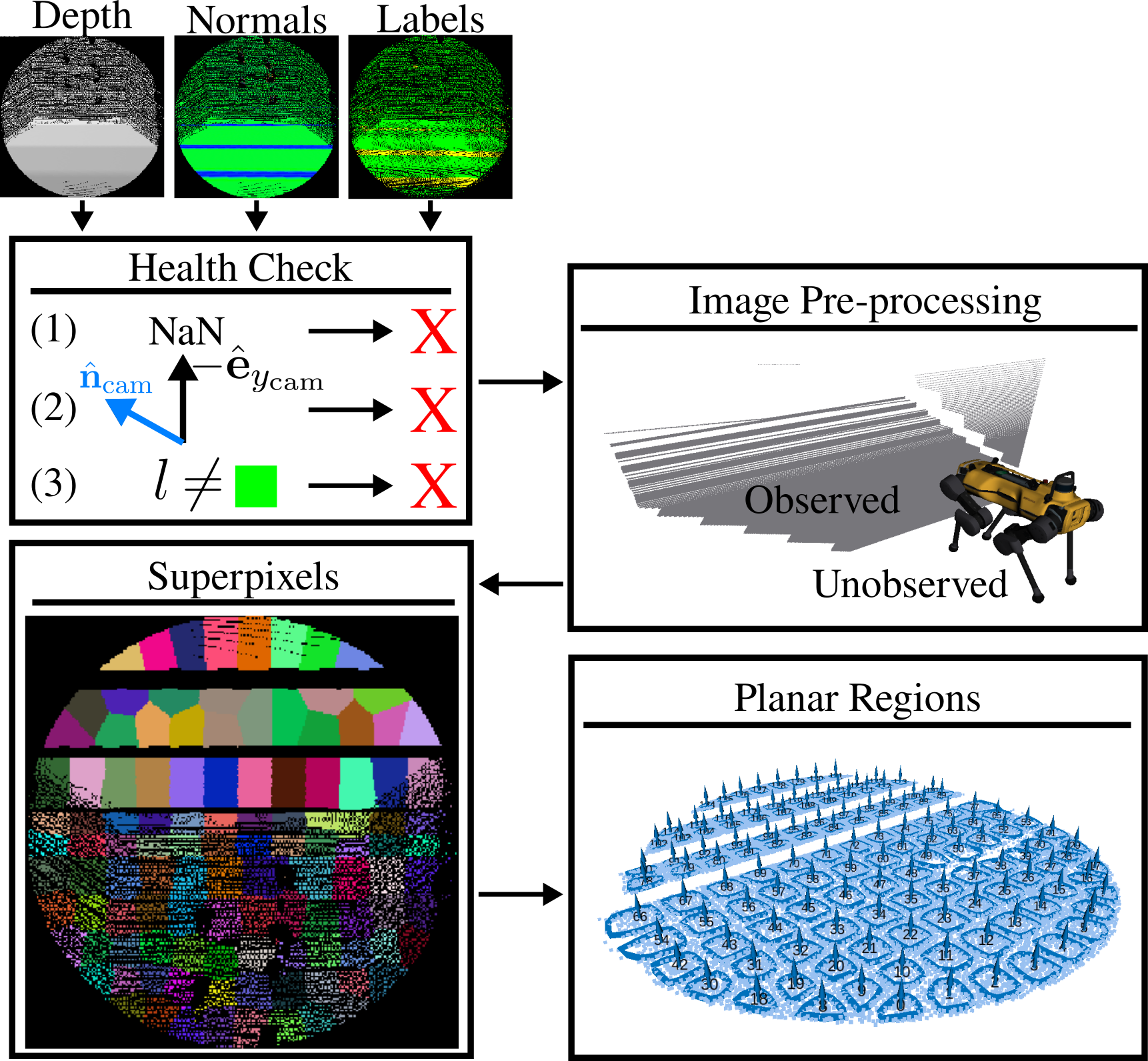}
  \caption{Information flow for the proposed superpixels oversegmentation and planar region extraction method. The input is the multi-channel floor image $I_{\rm low}$. First, the image is cleaned to remove all invalid pixels as well as pixels corresponding to non-upright normals and non-steppable labels. Next, regions of the image that have not yet been populated from sensor data are initialized to standard ground plane data. Then, the superpixels algorithm is run to generate a set of superpixel clusters, and those clusters are then passed to a convex hullification step to obtain closed planar regions.}
  \label{fig:superpixel_diagram}
\end{figure}

Once pixel-wise distances are calculated and clusters are assigned, cluster centroids are updated using the newly assigned pixels. Cluster normals are also refined via RANSAC \cite{fischler_random_1981}. This assignment and update process is repeated for a user-specified number of iterations $n_{\rm iter}$. A diagram for the workflow is also shown in Figure \ref{fig:superpixel_diagram}.

\subsection{Convex Planar Region Generation} \label{sec:convex_planar_regions}

Downstream foothold planning expects a set of convex planar regions as a characterization of the local environment. Therefore, these superpixel regions must be convexified. 

First, all points in a superpixel cluster $\zeta$ are projected onto the plane described by $(\mathbf{p}_\zeta, \mathbf{\hat{n}}_\zeta)$ from the region's centroid. Then, this set of 2D points is passed to the Graham Scan algorithm \cite{graham_efficient_1972} to compute the convex hull. The set of points defining the hull are then passed on to foothold planning as a set of planar regions $\mathcal{P}$. Each region $\varphi  \in \mathcal{P}$ is defined by a pose $\mathbf{q}_\varphi  \in \mathbb{R}^6$ in an inertial frame and a list of 2D boundary points $\{ \mathbf{b}_{\varphi ,0}, \mathbf{b}_{\varphi ,1}, \dots, \mathbf{b}_{\varphi ,n_b} \} \in \mathbb{R}^{2 \times n_b}$ that lie in the plane defined by $\mathbf{q}_\varphi $. The set size $n_b$ can vary between regions.

%% file: text_files/planning.tex
\section{Quadrupedal Motion Planning} \label{sec:planning}


\subsection{Problem Statement} \label{sec:problem_statement}

\begin{figure*}[t]
    \centering
    \includegraphics[width=0.99\textwidth]{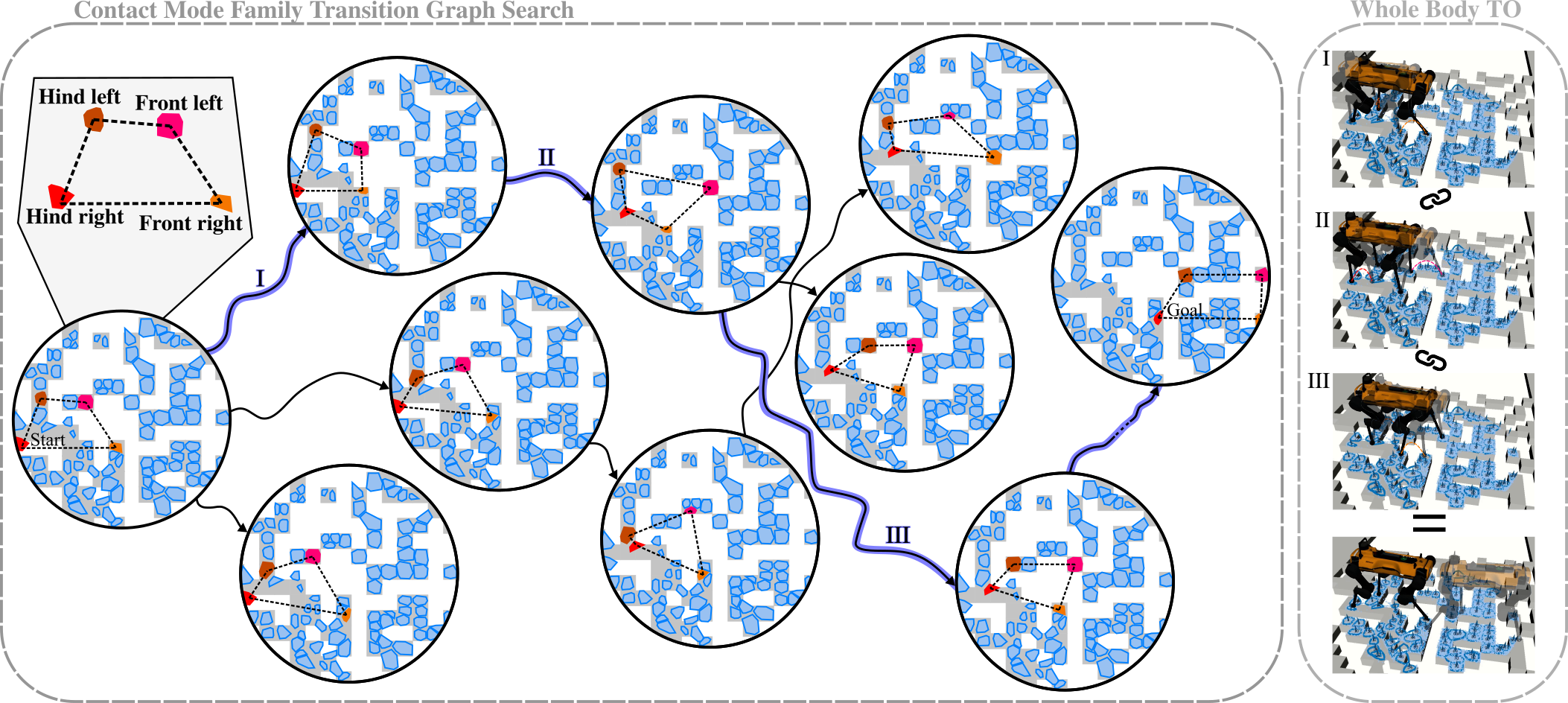}
    \caption{Overall diagram of the proposed planning framework. A graph search (Section \ref{sec:graph_search}) is performed over contact mode family transitions defined through a series of geometric and kinematic constraints including a user-defined gait, kinematic reachability volumes, and stance stability. Then, the suggested contact sequence is passed to a long-horizon trajectory optimization (Section \ref{sec:traj_opt}) problem to synthesize the whole body reference trajectory.}
    \label{fig:diagram}
\end{figure*}

Consider a quadrupedal robot with a configuration space $\mathcal{Q} \subset \mathbb{R}^{6 + n_j}$. The robot configuration $\mathbf{q} \in \mathcal{Q}$ is comprised of a torso pose $\mathbf{q}_{\rm torso} \in \mathbb{R}^6$ represented in an inertial frame and a sequence of $n_j$ joints $\mathbf{q}_j \in \mathbb{R}^{n_j}$. The pose $\mathbf{q}_{\rm torso}$ can be further decomposed into an orientation $\boldsymbol{\theta}_{\rm torso} = \begin{bmatrix}
    \alpha_{\rm torso} & \beta_{\rm torso} & \gamma_{\rm torso}
\end{bmatrix}^T \in \mathbb{R}^3$ and position $\mathbf{p}_{\rm torso} = \begin{bmatrix} x_{\rm torso} & y_{\rm torso} & z_{\rm torso} \end{bmatrix}^T\in \mathbb{R}^3$. The proposed planning framework enables the robot to travel from a starting configuration $\mathbf{q}_{\rm start}$ to a goal region $\mathcal{Q}_{\rm goal} = \{ \mathbf{q} \in \mathcal{Q} \hspace{0.05cm} \Big| \hspace{0.05cm} \| \mathbf{q}_{\rm torso} - \mathbf{q}_{\rm goal} \|_2 < \epsilon_{\rm torso} \} $ where $\epsilon_{\rm torso}$ is a user-defined distance threshold. In this work, the environment is entirely unknown a priori and partially observable. Therefore, the robot must use its onboard sensing to reveal the environment online. A pictorial representation of the planning framework is provided in Figure \ref{fig:diagram}.

\subsection{Torso Search} \label{sec:torso_search}
First, QuadPiPS performs a coarse initial search over the torso pose $\mathbf{q}_{\rm torso}$ to construct a guiding torso path through the local environment. The graph itself will be represented as $\mathcal{G}_{\rm torso} = ( \mathcal{N}_{\rm torso}, \mathcal{E}_{\rm torso})$. For each planar region $\varphi  \in \mathcal{P}$, a set of nodes $\{ \eta_{\varphi , 0}, \dots \eta_{\varphi , n_{\rm offset}} \}$ are created at the region pose $\mathbf{q}_\varphi $ plus an offset. The pose of each node is offset in the direction of the region's normal vector by a nominal torso height $h^{\rm nom}_{\rm torso}$ and in the region yaw by $\Delta \gamma_{\rm torso}$. An edge is added between nodes with a Euclidean distance under $d^{\rm max}_{\rm torso}$ and a yaw difference under $\Delta \gamma_{\rm torso}^{\rm max}$.

The node closest to the global goal is marked as the goal node. The A* search algorithm \cite{hart_formal_1968} is employed with Euclidean distance plus yaw difference as the edge weight and cost-to-go heuristic. The sequence of torso nodes is passed onto the foothold search as a guiding torso path $\mathcal{T}_{\rm torso} = \{ \mathbf{q}_{\rm torso, 0}, \dots, \mathbf{q}_{\rm torso, n_{\rm torso}} \}$ where the furthest pose $\mathbf{q}_{\rm torso, n_{\rm torso}}$ is taken as the local waypoint $\mathbf{q}_{\rm LG}$ for footstep planning.


\subsection{Foothold Search} \label{sec:graph_search}
QuadPiPS adapts its core hierarchical philosophy from the Augmented Leafs with Experience on Foliations (ALEF) framework \cite{kingston_scaling_2023, asselmeier_hierarchical_2024}. To perform perception-informed and kinodynamically-feasible quadrupedal motion planning in real time, QuadPiPS decomposes the task of foothold planning into its discrete and continuous planning spaces. The discrete space consists of the set of potential footholds, and the continuous space consists of the whole body configurations that the robot can assume to perform said footholds. In this section, the discrete space is discussed. 

Foothold planning is treated as a search problem over this lower-dimensional discrete planning space. This search employs the graph $\mathcal{G}_{\rm foot} = (\mathcal{N}_{\rm foot}, \mathcal{E}_{\rm foot})$ in which each node $\eta_{\rm foot} \in \mathcal{N}_{\rm foot}$ represents a contact mode family. A contact mode family is a partial stance in which a proper subset of the quadruped's feet are in contact with a unique combination of planar regions in the environment. For a single contact mode family, each stance foot must be in contact with a different region. Each edge $\xi_{\rm foot} \in  \mathcal{E}_{\rm foot}$ represents a transition between two partial stances, meaning that the edge itself is a full stance in which all feet are in contact. 

\subsubsection{Contact Mode Family Transition Graph Construction}
To initialize the contact mode family transition graph, a starting mode family $\eta_{\rm start}$ is constructed from the current robot state and inserted into a priority queue. This proposed framework employs lazy graph construction: during the search, the highest priority node $\eta_{\rm high}$ is popped from the queue and its neighboring nodes and their respective edges are only added to the graph once $\eta_{\rm high}$ is expanded. Nodes and edges are added according to set of constraints detailed below.


\textbf{Contact Sequence:} First, a contact sequence constraint is applied. Whichever contact phase the current node is in, the candidate nodes must satisfy the following phase of a user-defined gait. While this constraint is not strictly necessary and relaxing it would enable acyclic planning, it does help to reduce the number of potential transitions. 

\begin{figure}[h!]
    \centering
    \includegraphics[width=0.49\textwidth]{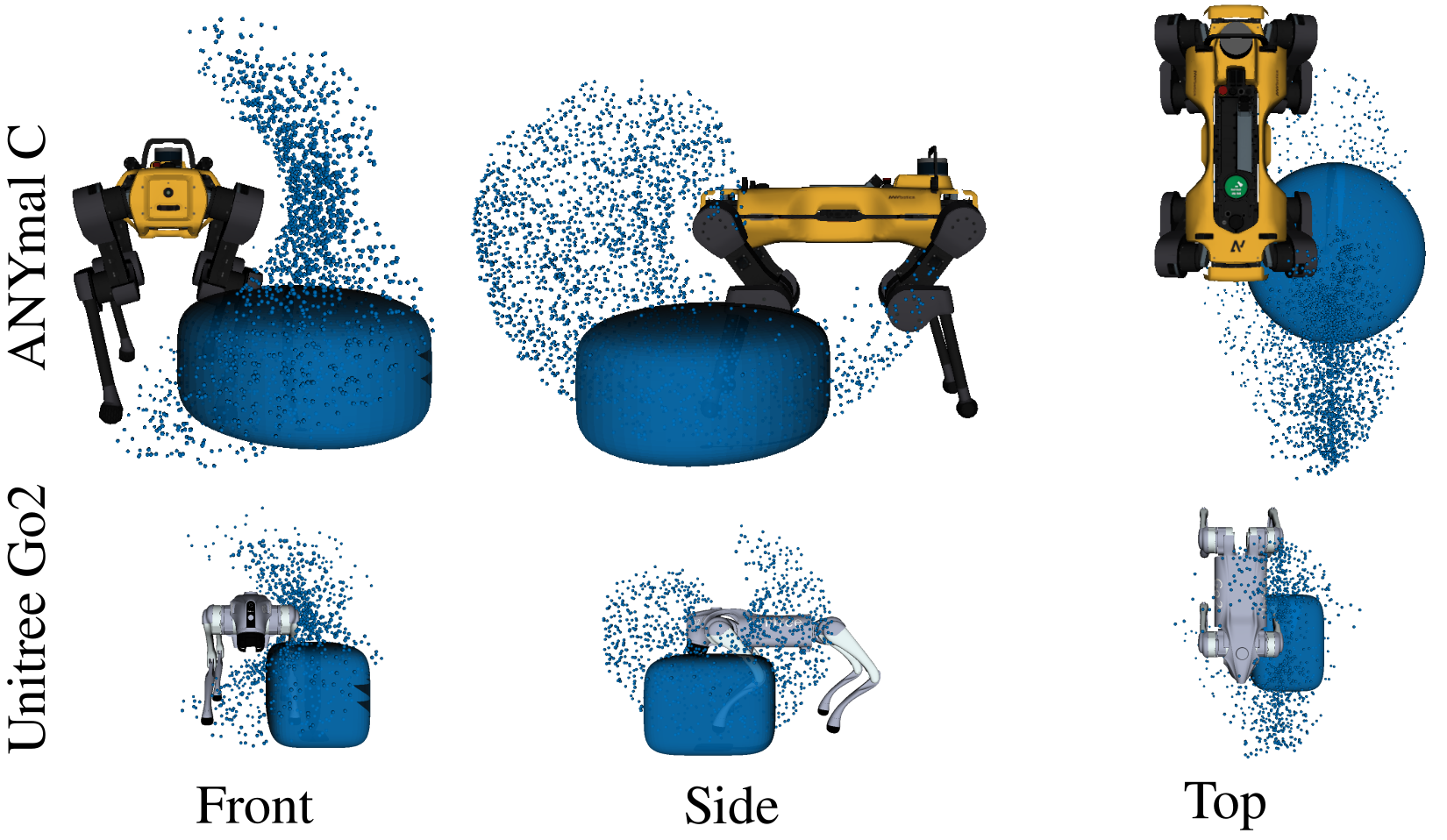}
    \caption{Visualizations of reachable volumes for ANYmal C (top row) and Unitree Go2 (bottom row). Only the reachable volumes for the front left foot are visualized here. Four volumes in total are generated, one for each foot.}
    \label{fig:superpixels}
\end{figure}

\textbf{Reachability:} Once the next set of stance feet has been decided, a set of reachable regions $\{ \mathcal{R}_{\rm FL}, \mathcal{R}_{\rm FR}, \mathcal{R}_{\rm HL}, \mathcal{R}_{\rm HR} \}$ are compared against the current set of perceived planar regions $\mathcal{P}$ to determine which reachable planar regions $\mathcal{P}_{\rm reach} = \{ \mathcal{P}_{\rm FL}, \mathcal{P}_{\rm FR}, \mathcal{P}_{\rm HL}, \mathcal{P}_{\rm HR} \}$ can be transitioned to for the current swing feet. The reachable regions are defined as superquadrics \cite{melon_receding-horizon_2021}, an expressive family of 3D volumes. The set of points within a superquadric centered at $(x_0, y_0, z_0)$ are defined as
\begin{equation} \label{eq:superquadrics}
    \mathcal{R} = \big\{ (x,y,z) \in \mathbb{R}^3 \hspace{0.05cm} \Big| \hspace{0.05cm} \big| \frac{x - x_0}{a} \big| ^d + \big| \frac{y - y_0}{b} \big| ^e + \big| \frac{z - z_0}{c} \big|^f \leq 1 \big\},
\end{equation}
where scalars $a, b, c$ and $d, e, f$ define the dimensions and curvature of the regions, respectively. The reachable volumes for the Go2 and ANYmal are visualized in Figure \ref{fig:superpixels}.

For each planar region $\varphi  \in \mathcal{P}$, the pose $\mathbf{q}_\varphi $ is evaluated in Equation \ref{eq:superquadrics} to determine if it can be reached for a given foot. Then, this graph expansion step iterates through $\mathcal{P}_{\rm reach}$. Each combination of reachable regions defines a candidate node $\eta_{\rm cand}$ that must be evaluated for addition to the graph.



\textbf{Stance stability:} The stance formed by the edge between the current node $\eta_{\rm curr}$ and the candidate node $\eta_{\rm cand}$ is checked for stability. A nominal torso pose $\mathbf{q}_{\rm nom} = \begin{bmatrix} \boldsymbol{\theta}_{\rm nom} & \mathbf{p}_{\rm nom} \end{bmatrix}^T~\in~\mathbb{R}^6$ and footholds $\{\mathbf{p}_{\rm FL}, \mathbf{p}_{\rm FR},\mathbf{p}_{\rm HL}, \mathbf{p}_{\rm HR}\}$ are estimated from this candidate edge, and the widths $w_{\rm front}, w_{\rm hind}$ and lengths $l_{\rm left}, l_{\rm right}$ of the stance feet positions are evaluated against a set of ideal stance dimensions taken at stand up. The stance positions are calculated as
\begin{equation}
    \begin{split}
        w_{\rm front} & = \| R^T(\boldsymbol{\theta}_{\rm nom}) \cdot (\mathbf{p}_{\rm FR} - \mathbf{p}_{\rm FL}) \cdot \mathbf{\hat{e}}_{y_{\rm torso}} \|_2  \\
        w_{\rm hind} & = \| R^T(\boldsymbol{\theta}_{\rm nom}) \cdot (\mathbf{p}_{\rm HR} - \mathbf{p}_{\rm HL}) \cdot \mathbf{\hat{e}}_{y_{\rm torso}} \|_2  \\
        l_{\rm left} & = \| R^T(\boldsymbol{\theta}_{\rm nom}) \cdot (\mathbf{p}_{\rm FL} - \mathbf{p}_{\rm HL}) \cdot \mathbf{\hat{e}}_{x_{\rm torso}} \|_2  \\
        l_{\rm right} & = \| R^T(\boldsymbol{\theta}_{\rm nom}) \cdot (\mathbf{p}_{\rm FR} - \mathbf{p}_{\rm HR}) \cdot \mathbf{\hat{e}}_{x_{\rm torso}} \|_2  \\
    \end{split},
\end{equation}
where the function $R(\boldsymbol{\theta}_{\rm nom})$ provides an SO(3) rotation matrix for the torso frame orientation $\boldsymbol{\theta}_{\rm nom}$ in the inertial frame and $(\mathbf{\hat{e}}_{x_{\rm torso}}, \mathbf{\hat{e}}_{y_{\rm torso}})$ provide the $x$- and $y$-dimension basis vectors in the torso frame. Stability is then represented as 
\begin{equation}
    \begin{split}
    | w_{\rm front} - w_{\rm ideal} | & < \epsilon_{\rm width} \textrm{ and }     | w_{\rm hind} - w_{\rm ideal} | < \epsilon_{\rm width} \\
    | l_{\rm left} - l_{\rm ideal} | & < \epsilon_{\rm length} \textrm{ and }    | l_{\rm right} - l_{\rm ideal} | < \epsilon_{\rm length} 
    \end{split},
\end{equation}
where $\epsilon_{\rm width}, \epsilon_{\rm length}$ are user-defined error thresholds.

\textbf{Swing distance:} Lastly, a constraint is applied to the distance between the liftoff and touchdown positions of the swing feet, $\mathbf{p}_{\rm liftoff}, \mathbf{p}_{\rm touchdown} \in \mathbb{R}^3$, for the candidate node.

\begin{equation}
    \| \mathbf{p}_{\rm liftoff} - \mathbf{p}_{\rm touchdown} \|_2 \leq d^{\rm max}_{\rm swing},
\end{equation}
where $d^{\rm max}_{\rm swing}$ is a user-defined maximum swing distance.

If all constraints are met, the candidate node $\eta_{\rm cand}$ and edge between $\eta_{\rm curr}$ and $\eta_{\rm cand}$ are added if they do not already exist.

\subsubsection{Contact Mode Family Transition Graph Search} To plan a discrete sequence of footholds, QuadPiPS performs a search over the contact mode family transition graph $\mathcal{G}_{\rm foot}$. The graph search itself follows a standard $A^*$ \cite{hart_formal_1968} implementation.

\subsubsection{Edge weight} For a transition from $\eta_{i}$ to $\eta_{i+1}$, the edge $\xi_{i, i+1} = (\eta_{i}, \eta_{i+1})$ is given the weight $d_{\text{torso}}(\eta_{i}, \eta_{i+1})$ in which $d_{\text{torso}}(\eta_{i}, \eta_{i+1})$ computes the Euclidean distance between nominal torso positions for $\eta_{i}$ and $\eta_{i+1}$.

\subsubsection{Cost-to-go} Each node $\eta$ is added to the priority queue according to the search heuristic $g(\eta) = d_{\rm torso}(\eta, \mathbf{q}_{\rm LG})$.

This graph search returns a discrete sequence of footholds $\mathcal{T}_{\rm feet}$ that are passed to a whole-body TO program.

\subsection{Swing Trajectory Optimization} \label{sec:traj_opt}

\subsubsection{Trajectory Optimization Subproblem Formulation}

The proposed formulation employs the state $\mathbf{x} = \begin{bmatrix} \mathbf{q}_{\rm torso} &\mathbf{h} & \mathbf{q}_j \end{bmatrix} ^T \in \mathbb{R}^{24}$ where the term $\mathbf{h} = \begin{bmatrix} \mathbf{k} & \mathbf{l} \end{bmatrix} ^T \in \mathbb{R}^6$ is the centroidal momentum vector comprised of the angular and linear $\mathbf{k} \in \mathbb{R}^3$, $\mathbf{l} \in \mathbb{R}^3$ momentum, respectively. The input consists of $\mathbf{u} = \begin{bmatrix} \mathbf{f} & \mathbf{v}_j \end{bmatrix} ^T \in \mathbb{R}^{24}$ where $\mathbf{f} \in \mathbb{R}^{12}$ are the contact forces and $\mathbf{v}_j \in \mathbb{R}^{12}$ are the joint velocities. The optimization subproblem for synthesizing a swing trajectory between two stance configurations is then:
\begin{subequations} \label{eq:whole_to}
\begin{align} 
& \underset{\mathbf{x},\mathbf{u}}{\min} && \sum_{k=0}^{{n_{\rm knot}} - 1} \Biggl( \| \mathbf{x}_{k} - \mathbf{x}^{\text{des}}_{k} \|^2 _Q + \| \mathbf{u}_{k}  - \mathbf{u}^{\text{des}}_{k}\|^2_{R} \Biggr) &&& \nonumber \\
& && + \| \mathbf{x}_{{n_{\rm knot}}} - \mathbf{x}^{\text{des}}_{{n_{\rm knot}}} \|^2 _{Q_f} &&& \nonumber \\
& \text{s.t.}  && &&& \nonumber \\
& \text{} && \dot{\mathbf{x}}_{k} = 
    f_{\rm cent}(\mathbf{x}_{k}, \mathbf{u}_{k}) \label{eq:dynamics} \\ 
& \text{} && f_{ {\rm planar},l}(\mathbf{q}_{k}, \varphi _l) = \mathbf{0} \hspace{1.0cm} \forall l \in \mathcal{C}_{i} &&& \label{eq:src_mode} \\ 
& \text{} && \mathbf{f}_{l, k} \in \mathcal{F}_l(\mu, \mathbf{q}_{k}) \hspace{1.675cm} \forall l \in \mathcal{C}_{i} &&& \label{eq:friction_cone} \\ 
& && \mathbf{f}_{l,k} = \mathbf{0} \hspace{1.75cm} \quad \quad \quad \forall l \notin \mathcal{C}_{i} &&& \label{eq:zero_force} \\
& \text{} && \mathbf{q}_{j,k} \in \mathcal{Q}_{\rm feas} &&& \label{eq:state_limits} \\ 
& && \mathbf{v}_{j,k} \in \mathcal{V}_{\rm feas} &&& \label{eq:control_limits}
\end{align}
\end{subequations}
where $f_{\rm cent}$ is the centroidal dynamics model \cite{orin_centroidal_2013}. The set $\mathcal{F}_l(\mu, \mathbf{q}_k)$ represents the friction cone which depends on the friction coefficient $\mu$ and robot pose $\mathbf{q}_k$. The planar region constraint function $f_{ {\rm planar},l}(\mathbf{q}_{k}, \varphi _l)$ is represented as a set of half-space constraints defined by the boundary of the planar region $\varphi _l$ for swing foot $l$. The set $\mathcal{C}_{i}$ represents the set of stance feet for the contact mode family, or node,  $\eta_{i}$. The sets $\mathcal{Q}_{\rm feas}$ and $\mathcal{V}_{\rm feas}$ define joint position and velocity limits. The cost matrices $Q$ and $R$ are defined through user-defined hyperparameters, and $Q_f$ is computed from a linear-quadratic regular solution at a nominal state and input.

\subsubsection{Implementation Considerations}
The above TO subproblem is solved with the Optimal Control for Switched Systems (OCS2) library \cite{farbod_farshidian_and_others_ocs2_nodate}. Each swing trajectory occurs over a time horizon of $t_{f} = 0.5$ seconds using $n_{\rm knot}=50$ knot points. The target state $\mathbf{x}^{\rm des}_{n_{\rm knot}}$ is obtained through random sampling and projection using the pseudo-inverse of the Jacobian matrix \cite{kingston_scaling_2023}. Then, cubic splines between the takeoff and touchdown configurations define the flight phase of $\mathbf{x}^{\rm des}$. The desired input $\mathbf{u}^{\rm des}$ is set to zero to encourage a minimum-energy solution.



The solutions to the sequence of swing trajectory subproblems are appended to form a long-horizon reference trajectory. This reference trajectory is then run through a refinement step to smoothen any irregularities between subproblem solutions. After refinement, the reference trajectories $\mathbf{x}^{\rm ref}$ and $\mathbf{u}^{\rm ref}$ are sent to an MPC module for real-time tracking.


%% file: text_files/control.tex
\section{Motion Tracking Control} \label{sec:control}

\subsection{Reference Trajectory Publication Monitoring} \label{sec:ref_traj_pub}
The proposed motion planner publishes the reference trajectory for tracking and subsequently monitors the robot state for events that signal a re-plan. One such event is that the robot completes the previously sent reference trajectory. Another such event is that the quadruped begins to tip or becomes unstable, signaled by an abnormally large roll or pitch or abnormally large joint velocities or torques. If so, the quadruped is commanded to abort the prior reference trajectory and maintain stability before re-planning.

\subsection{MPC Trajectory Tracking} \label{sec:mpc}
The Model Predictive Controller (MPC) follows the same formulation given in Equation $\ref{eq:whole_to}$. The MPC employs a timestep $\Delta t$ of $15$ milliseconds over a time horizon of $t_f=1.0$ seconds. The solver is also warm-started for each new solution using the prior solution and subsequently run for only one SQP iteration. At this level, the reference trajectory $\mathbf{x}^{\rm ref}$ encourages the MPC to step on the selected regions from the graph search, but the MPC has the authority to assign swing feet to different regions in order to maintain stability. The MPC also synthesizes new swing trajectories depending on the current foot positions to afford the solver more flexibility to deviate from and return to the reference trajectory when necessary.

\subsection{WBC} \label{sec:wbc}

The optimal state $\mathbf{x}^*$ and input $\mathbf{u}^*$ from the MPC are passed on to a whole body controller (WBC) to synthesize the joint torque commands $\boldsymbol{\tau}$. This WBC takes the form of a hierarchical quadratic program (QP) adapted from \cite{dario_bellicoso_perception-less_2016}.

%% file: text_files/experimental_results.tex
\section{Experimental Results} \label{sec:experimental_results}

\subsection{Experimental Setup - Software} \label{sec:experimental_setup_software}
QuadPiPS is integrated into the \texttt{quad\_auto} codebase which is an extension of the existing open-sourced \texttt{legged\_control} \cite{qiayuan_liao_and_others_legged_control_nodate} codebase.  All code for testing and deployment is provided in the project repository\footnote{\href{https://quadpips.github.io/}{https://quadpips.github.io/}}.

\subsection{Simulation Benchmarking} \label{sec:simulation_benchmarking}
Simulation benchmarking contextualizes QuadPiPS performance amongst several other methods for footstep planning. All testing is performed on a Dell Precision 3660 with an Intel i9-12900K CPU (single-thread passmark score of $4,336$; multi-thread score of $41,322$) and an NVIDIA T1000 GPU.

\subsubsection{Baselines}
Benchmarking compares five baselines:

\textbf{Elevation Mapping-based Model Predictive Control (EM-MPC) \cite{grandia_perceptive_2022}:} This baseline performs elevation mapping \cite{miki_elevation_2022, noauthor_anyboticselevation_mapping_2024} to obtain a local 2.5D grid representation that is passed on for filtering, classification, segmentation, and precomputation. Inpainting and median filters are applied to mitigate occlusions and noise. Binary steppability classification is performed to ignore regions of the elevation map that can not support a foothold. Then, planar region segmentation is performed via connected component labelling and contour extraction. Lastly, an SDF is computed over the map for collision detection. The MPC formulation follows the same design as Equation \ref{eq:whole_to} so that the perception pipelines can be compared with the same planning and control.

\textbf{Elevation mapping-based Reinforcement Learning (EM-RL) \cite{miki_learning_2022}:} This reinforcement learning (RL) baseline employs the same elevation mapping pipeline, but passes a downsampled map along with a twist command and proprioceptive state information into a multi-layer perceptron which outputs a leg phase offset and residual joint position targets. 

\textbf{Elevation mapping-based Deep Tracking Control (EM-DTC) \cite{jenelten_dtc_2024}} This baseline employs the same elevation mapping pipeline, but employs both a model-based foothold planner as well as a model-free tracking controller. A terrain-aware trajectory generation scheme \cite{jenelten_tamols_2022} is employed to provide desired footholds, base poses, twists, and accelerations, and joint positions. Then, a learned RL policy tracks the desired state and outputs target joint positions. 

\textbf{Superpixels-based Model Predictive Control (S-MPC):} This baseline employs the proposed perception pipeline in this work. This oversegmented set of planar regions is passed into the MPC formulation defined in Section \ref{sec:mpc}. Footholds are generated through a heuristic calculation and local adjustment based on nearby planar regions.
\textbf{Quadrupedal footstep Planning in the Perception Space (QuadPiPS):} The entire proposed framework.
\subsubsection{Environments}
Benchmarking spans 10 different environments designed to evaluate distinct terrain attributes: \textit{Ramp}, \textit{Stairs}, \textit{Rubble}, \textit{Pegboard}, \textit{Balance beam}, \textit{Ramped balance beam}, \textit{Ramped stepping stones}, \textit{Sparse stepping stones}, \textit{Obstructed balance beam}, and \textit{Winding balance beam}.

\subsubsection{Perception Timing}
\begin{figure}[h!]
    \centering
    \includegraphics[width=0.99\linewidth]{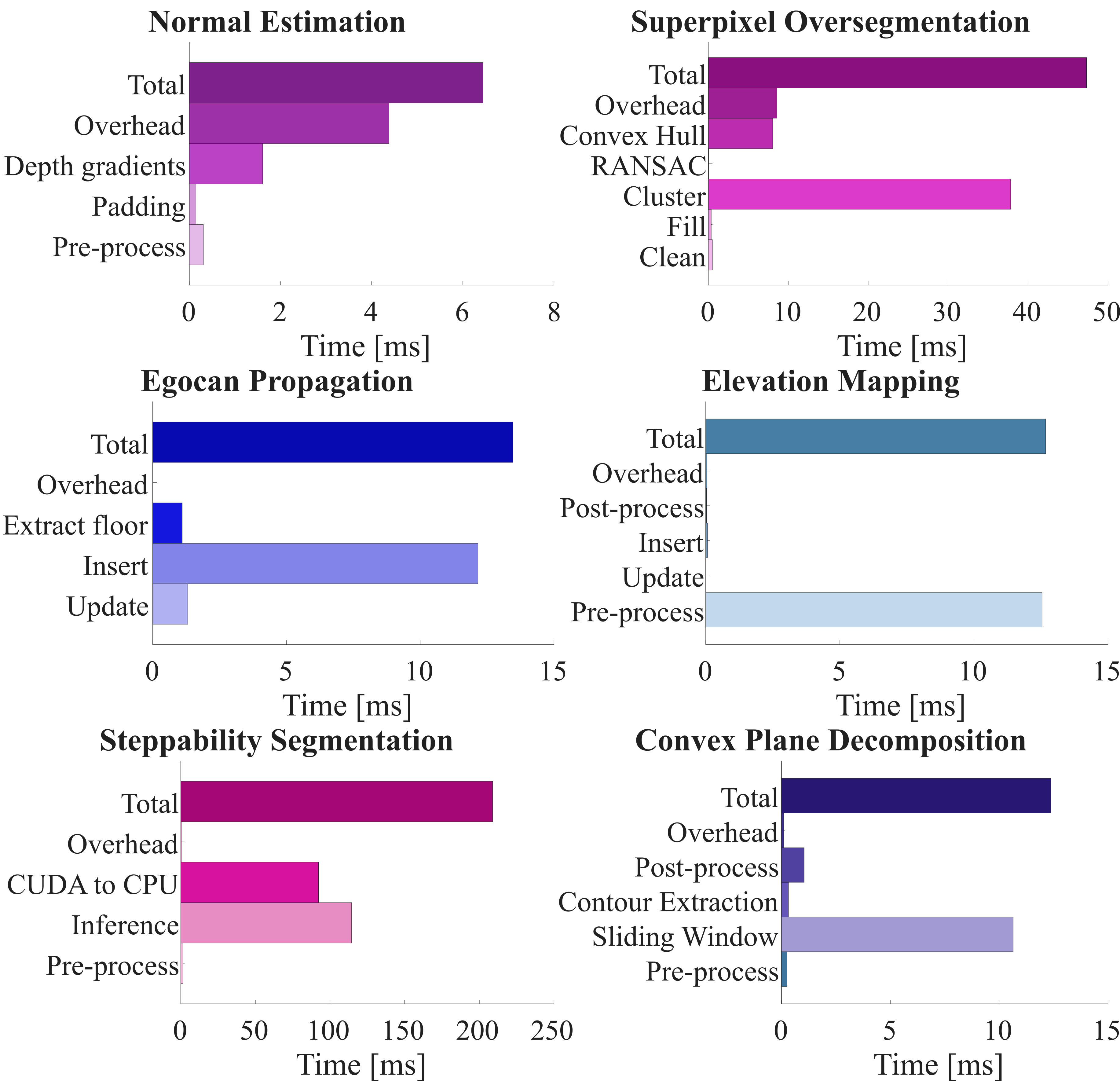}
    \caption{Average timing for each module in the proposed superpixels-based perception pipeline along with the elevation map-based perception pipeline. The steppability model is run in a separate update thread to prevent the large inference time from acting as a bottleneck on the entire pipeline.}
    \label{fig:combined_timing_sim}
\end{figure}

Average timing is recorded across 10 runs in the Ramp simulation environment. A RealSense D435i camera streams depth images at $60$~Hz. A module-by-module breakdown on timing is given in Figure \ref{fig:combined_timing_sim}.

While the steppability segmentation is not able to run in real-time given GPU hardware constraints, the overall update rate of the perception pipeline is not constrained by this bottleneck. When a new segmentation mask is available, that information is included in the egocan update. Otherwise, new points are added with an unknown steppability label and prior points with labels are propagated forward. The perception pipeline can then run at $20$~Hz. 

The elevation mapping update rate is comparable to the egocan update rate. However, the planar region decomposition pipeline runs at roughly four times the speed of the superpixel oversegmentation. The incoming point cloud used for elevation mapping is also downsampled to reduce the time spent preprocessing. This timing difference is the tradeoff for performing oversegmentation. However, this level of granularity pays off in performance, as will be seen in Section \ref{sec:benchmarking_results}.

\subsubsection{Benchmarking Results} \label{sec:benchmarking_results}

\begin{figure*}[h!]
    \centering
    \includegraphics[width=\textwidth]{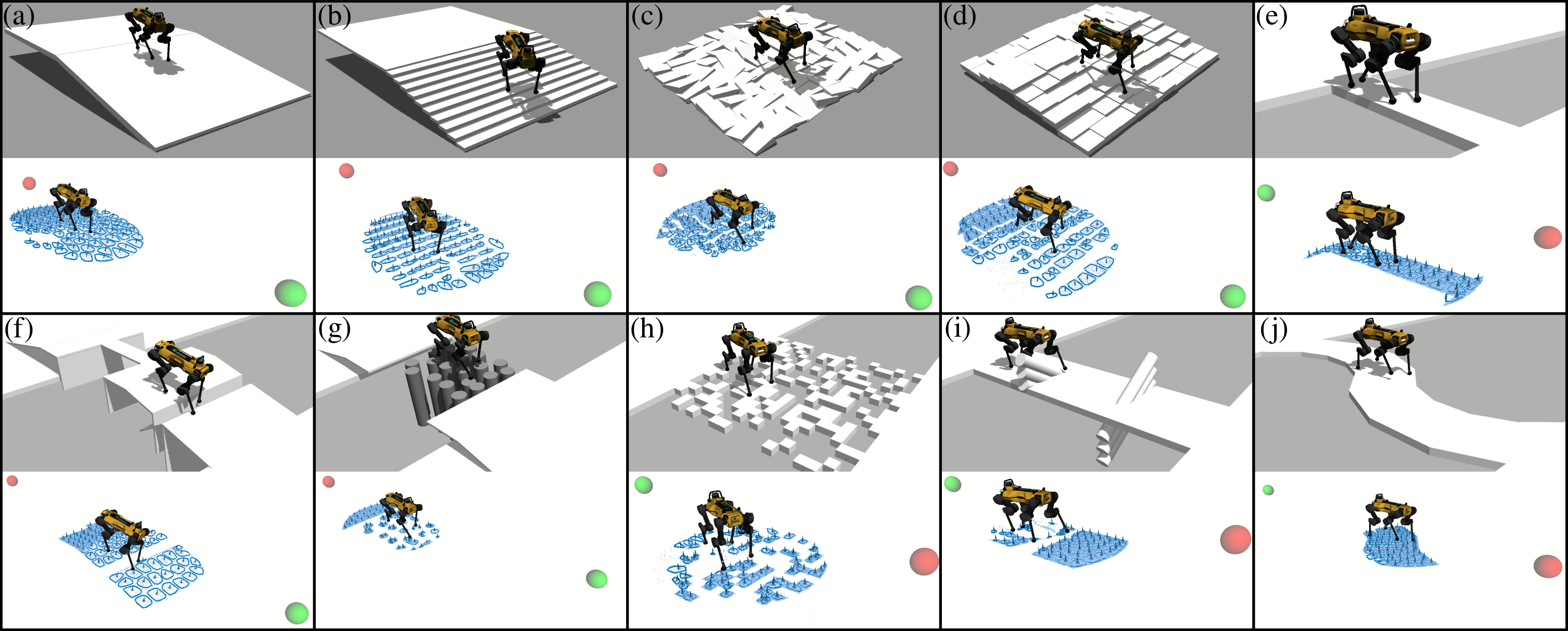}
    \caption{Visualization of environments with corresponding superpixels. (a) Ramp, (b) Stairs, (c) Rubble, (d) Pegboard, (e) Balance beam, (f) Ramped balance beam, (g) Ramped stepping stones, (h) Sparse stepping stones, (i) Obstructed balance beam, and (j) Winding balance beam. Top images are in Gazebo and bottom images are RViz. In RViz, green spheres represent the start torso pose and red spheres represent the goal torso pose. Superpixels are depicted by blue boundaries, normals, and points.}
    \label{fig:environments}
\end{figure*}

\begin{figure*}[h!]
    \centering
    \includegraphics[width=0.99\textwidth]{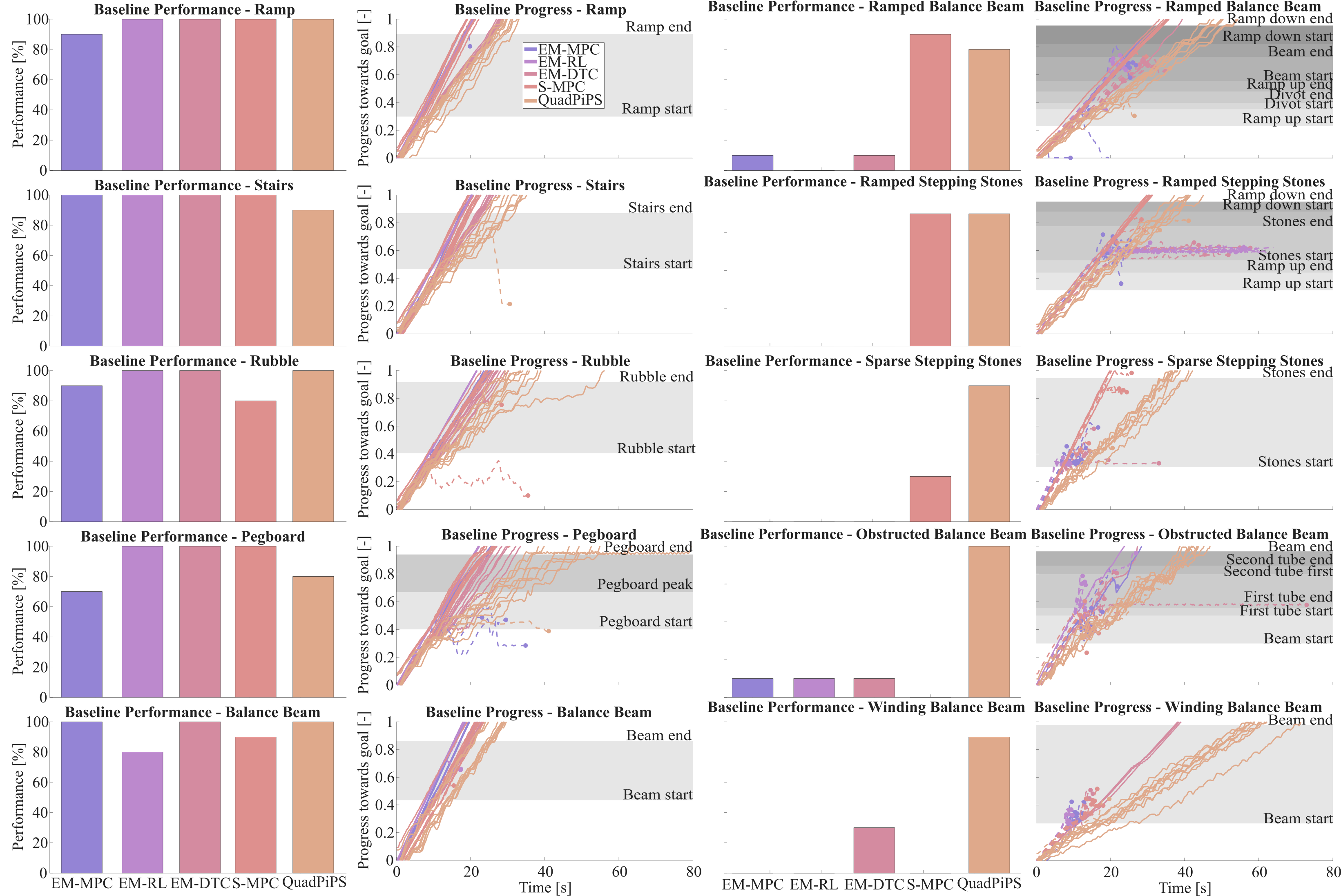}
    \caption{Simulation benchmarking results for the ten environments. For each environment, the left plot shows the success rates for all baselines. The right plot shows task progress percentage over time for each attempted trial. Solid lines are successes and dashed lines are failures. Environmental feature locations are also displayed to give context on where trials failed.}
    \label{fig:baselines}
\end{figure*}

Each baseline is run in each environment over 10 trials. For each trial, the robot is assigned the same start and goal position. All baselines outside of QuadPiPS take a twist command pointing from the current robot position to the goal position as input. For QuadPiPS, the global goal position is provided to the foothold planner. All trials are run on the ANYmal C platform due to the limited availability of learned baselines across quadrupedal platforms; pre-trained models for the EM-RL and EM-DTC were only available for the ANYmal C platform, and the proposed work is comprised of model-based planners that can be adapted to any quadrupedal platform with minimal tuning. For simulation trials, ground truth odometry is provided to maximize performance and restrict independent variables to perception, planning, and control. If the quadruped falls over or steps off of the evaluated terrain elements, the trial is marked as a failure. Otherwise, if the baseline reaches the goal, the trial is marked as a success. Visualizations of the environments along with sample superpixels are provided in Figure \ref{fig:environments} and baseline results are depicted in Figure \ref{fig:baselines}.


\textbf{Ramp:} Only one EM-MPC trial failed on this environment. During this trial, the whole body controller hit the maximum number of recalculations during its QP solution and returned a poorly conditioned set of joint torques. This caused the quadruped to collapse. While somewhat of an outlier, this trial does highlight a key element to the model-based and model-free control tradeoff. Model-based optimization problems have larger variance in solve times due to features such as initial conditions and time-varying constraints. Model-free controllers represented as neural networks have more consistent inference times, which can be seen in the similar progress rates of the trials for EM-RL and EM-DTC. 

\textbf{Stairs:} One QuadPiPS trial failed on this environment. The MPC module received a reference trajectory at a slight time delay due to latency. Tracking was delayed and the robot lurched forward in attempt to catch up to the reference trajectory, causing the robot to fall. This trial highlights the fragility of model-based planners. Even small perturbations from assumed conditions can cause failures. 

The back of the egocan floor image tended to be sparser due to attrition of points during propagation. Planar regions behind the quadruped did span across steps because of this, causing some stumbles that were small enough to recover from. 


\textbf{Rubble:} In this setting the EM-MPC baseline sustained one failure whereas the S-MPC baseline sustained two failures. These failures arose from unexpected contact with the environment. In the case of EM-MPC, this was due to the smoothed elevation map reporting incorrect terrain heights. The elevation map segmented the entire local representation into a single planar region during most runs. In the case of S-MPC, these arose from inaccurate swing trajectory tracking causing contact with the side of a stone as opposed to the top. This is another example of the type of small perturbation from expected conditions, including assumed contact sequences, that model-predictive controllers can struggle to reject.


\textbf{Pegboard:} Three failures are observed for the EM-MPC baseline and two are observed for QuadPiPS. EM-MPC failures stemmed from the large step up required to initially mount the pegboard. As can be seen from Figure \ref{fig:baselines}, the failures are clustered towards the start. For QuadPiPS, observed failures were largely due to instability during tracking. Sequencing multiple TO subproblems together limits the momentum of the system between steps. For environments such as this, maintaining forward momentum is crucial to stability. 

The learned controllers were the smoothest over this setting. An occasional foothold would get stuck in the negative steps on the pegboard, but the learned controllers were able to lift the foot out of these and continue walking.


\textbf{Balance beam:} In this setting, the EM-RL baseline exhibited two failures and the S-MPC baseline exhibited one. The EM-RL stepped off of the balance beam and fell. This can largely be attributed to the lack of explicit foothold planning. While crossing the beam, the EM-RL baseline would also crouch down and place its feet very close to each other. For the S-MPC baseline, the quadruped approached the beam at an offset and the MPC struggled to transition all four feet onto the beam, leading to a slip. In some trials, QuadPiPS executed lateral steps to align the robot with the beam before stepping onto the beam which proved to be useful. The EM-DTC exhibited the impressive ability to recover from one foot stepping off the beam on a handful of trials.


\textbf{Ramped balance beam:} In this setting, the EM-RL baseline reported no successful trials whereas the EM-MPC and EM-DTC trials each report one successful trial. The S-MPC baseline reported nine successes whereas the QuadPiPS baseline reported eight successes. For the elevation mapping baselines, most failures were concentrated on the balance beam. The two failure modes were mounting the balance beam at an offset and failing to transition all four feet onto the beam, or getting onto the beam but losing stability and tipping over. A select few trials failed on the ramp up to the beam. Some baselines stepped in the divot in the ramp and buckled, leading to a fall. 

The superpixels-based perception pipeline presented a significant benefit in this environment. The region oversegmentation provided a natural decomposition over the balance beam that facilitated S-MPC and QuadPiPS with adjusting their desired footholds inwards onto the beam. Figure \ref{fig:environments}\textcolor{red}{(f)} shows that superpixels were not generated for the divot.


\textbf{Ramped stepping stones:} All elevation mapping baselines reported zero successful trials in this environment. Moreover, these baselines made little to no progress over the stepping stones themselves. This came largely from the elevation map representation. The continuous nature of the elevation map led to the height information of smaller stones being merged together, giving the false impression of one large surface that can support footholds. This occurred even when infilling and smoothing filters were removed. With this terrain representation, all baselines planned footholds over unsafe terrain and fell off the stepping stones. The superpixels representation generated clean, disparate regions which enabled stable foothold planning across the terrain. Subsequently, the superpixels baselines were the only planners capable of making progress across the stepping stones, with S-MPC and QuadPiPS both recording nine successful trials. Failures for these baselines arose from inaccurate foothold tracking which caused the robot to slip off the desired stepping stones.


\textbf{Sparse stepping stones:} In this setting, the elevation mapping baselines all reported no successful trials. The S-MPC baseline reported three successful trials, and the QuadPiPS baseline reported eight. The elevation map representation exhibited similar issues to what was discussed for the Ramped Stepping Stones environment in which nearby stepping stones were merged into incorrectly large planar regions. The S-MPC was able to complete some trials, but the nominal foothold positions generated by the MPC constrained the planner to suboptimal locations. All failures observed for S-MPC were from slipping off of stones due to an inability to accurately track the heuristically generated footholds. The search-based foothold planning in the QuadPiPS baseline proved to be enormously beneficial in finding stable stance configurations in this environment. QuadPiPS took longer to find a path to the goal, and these paths were often more circuitous compared to other baselines. However, for an environment such as this where terrain is sparse and precise foothold planning is critical, this tradeoff is heavily preferred.


\textbf{Obstructed balance beam:} In this environment, each elevation map baseline reported one successful trial. These baselines walked directly into the barriers and, by chance, were able to stay upright while circumventing them. For the failed trials, these baselines would often attempt to step onto the barriers which would push the robot towards the edge of the beam, leading to the robot falling off the beam.

For this environment, the legged affordances provided by the steppability model proved essential. Geometric information of depth and surface normals permits planning a foothold on the angled barriers, but this could easily tip the robot over. Therefore, having the steppability model to mark this terrain as passable ensured that the QuadPiPS framework did not step on the barriers. While the S-MPC also does this, the heuristics used to generate nominal footholds still place the footholds in the center of the beam. The MPC then attempts more difficult swing trajectories over taller parts of the barriers. 

\textbf{Winding balance beam:} For this environment, the EM-DTC baseline recorded three successful trials, and the QuadPiPS framework recorded nine. MPC modules do not have much authority or flexibility to deviate significantly from their desired trajectory. If this desired trajectory is infeasible or poorly conditioned, this can generate internal conflict within the solver. For this environment, the reference trajectory encourages the quadruped to keep its torso along the line between the start and the goal. However, this does not align with the available local terrain. This mismatch caused the baselines outside of QuadPiPS to step off of the beam. For QuadPiPS, all levels of planning are informed by the local terrain. Therefore, the guiding torso path encourages the quadruped to follow the curvature of the beam.





\subsection{Simulation Demonstrations} 
\begin{figure}[h!]
    \centering
    \includegraphics[width=0.99\linewidth]{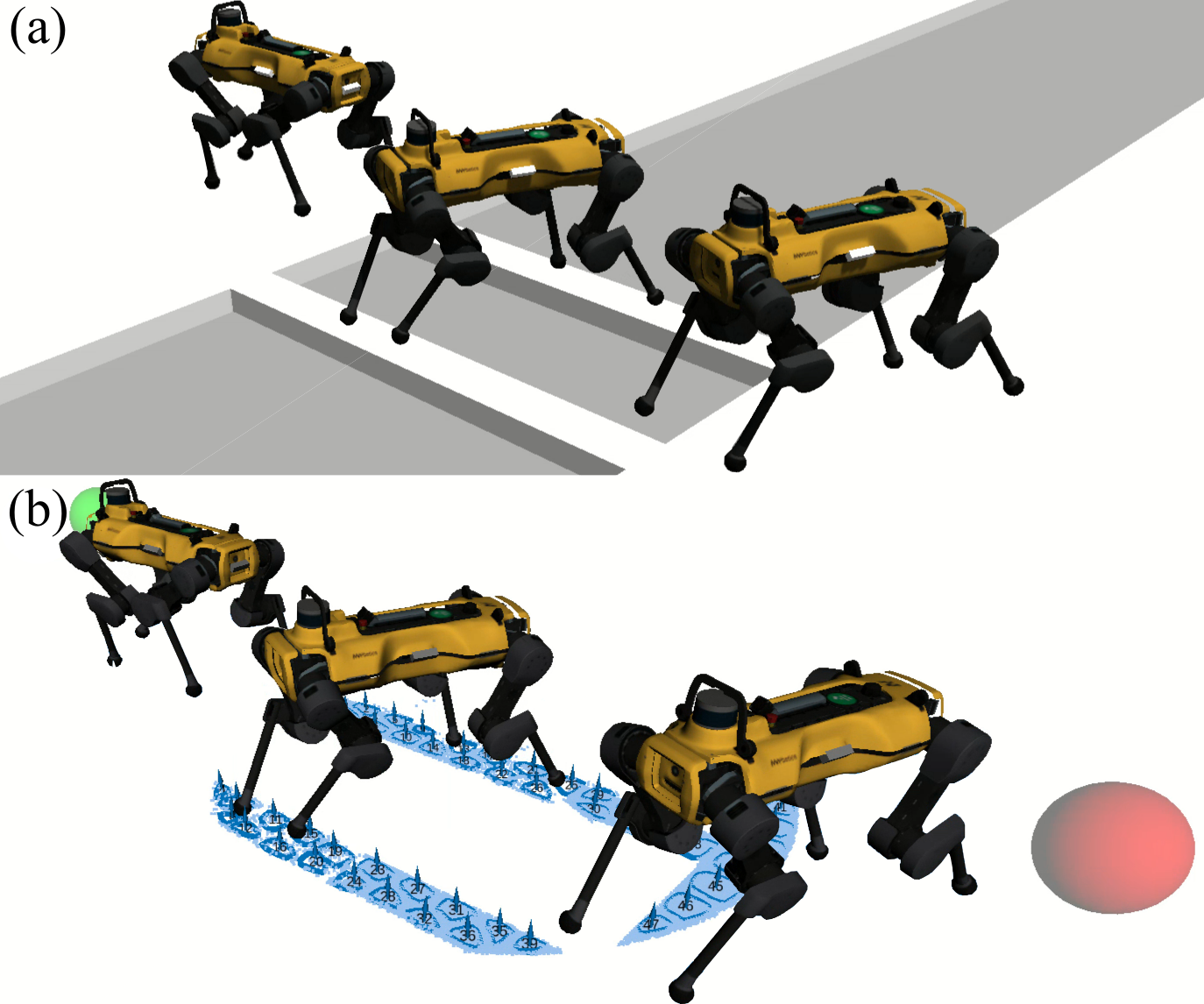}
    \caption{Visualization of the QuadPiPS framework traversing parallel beams. (a) Gazebo depiction, and (b) RViz visualization. In RViz, the green sphere represents the start torso pose and the red sphere represents the goal torso pose. Superpixels are depicted by blue boundaries, normals. and points.}
    \label{fig:side_step_beams}
\end{figure}

Demonstrations are also provided for a parallel beams environment, as seen in Figure \ref{fig:side_step_beams}, to showcase the ability of QuadPiPS to find challenging stance configurations.

\textbf{Parallel beams:} This environment possesses two beams that are each a width of 0.2~m and are separated by a distance of 0.85~m, roughly the length of the ANYmal. Here, the ANYmal must cross by side-stepping across the beams. For this environment, additional depth cameras are attached to the left and right sides of the robot. This is only done because the front-facing camera can not perceive the environment as the quadruped is side-stepping across the beams.

The distance between the parallel beams is equal to the ideal stance length $l_{\rm ideal}$ in the stance stability constraint in the contact mode family transition graph construction. Therefore, when searching over the beams for candidate stances, the graph search is encouraged to face sideways and strafe across the beams instead of facing forward and widening its stance to step along the beams. Heuristic footstep planners simply walk forward across the beams, but the stance required to bridge the two beams is too wide for the robot.  In this setting, large changes to the torso orientation are necessary which heuristic footstep planners or MPC modules struggle to perform unless they are explicitly instructed to do so.

\subsection{Experimental Setup - Hardware} \label{sec:experimental_setup_hardware}
Hardware experiments are run with the Contact-Aided Kalman Filter \cite{hartley_contact-aided_2020} along with a RealSense T265 Localization Camera. The torso orientation $\theta_{\rm torso}$ is observed directly from the onboard IMU. 
Position estimates are fused with $x_{\rm torso}$ and $y_{\rm torso}$ from the T265 and $z_{\rm torso}$ from the Kalman filter.

\subsubsection{Networking}
There are two PCs onboard the quadruped that collectively run the autonomy stack. First, an NVIDIA Jetson Orin NX with an ARM Cortex-A78AE CPU (single-thread passmark score of $938$; multi-thread passmark score of $3,788$) runs the steppability module. Depth images are streamed in from a RealSense D435i camera featuring depth images with $640\times480$ dimensions at $30$~Hz on hardware. Second, a Minisforum UM890 Pro Mini PC with an AMD Ryzen 9 8945HS CPU (single-thread passmark score of $3,822$; multi-thread passmark score of $29,446$) runs the remaining modules.  A third low-level PC on the quadruped handles internal communication, sensor readings, and API requests. This machine sends out joint state readings and receives joint torque commands over ROS2 \cite{macenski_robot_2022} + DDS \cite{pardo-castellote_omg_2003}. Lastly, the user performs remote visualization and monitoring and sends high-level waypoint commands on a Lenovo ThinkPad X1 Carbon laptop with an Intel Core Ultra 7 155U CPU (single-thread passmark score of $3,315$); multi-thread passmark score of $16,160$) over ROS2+DDS as well. Communication between the Jetson, MiniPC, and the internal PC is done over ethernet for minimum latency. Communication between the MiniPC and the laptop is done over WiFi on a private network to ensure real-time performance.

\subsubsection{Computational Suite}
\begin{figure}[h!]
    \centering
    \includegraphics[width=0.99\linewidth]{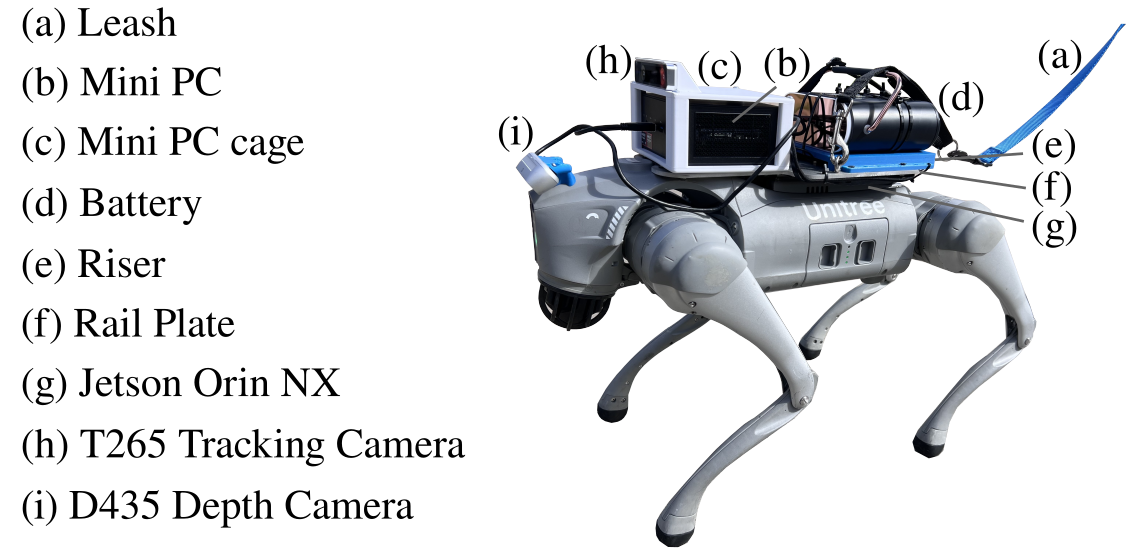}
    \caption{Custom mounting setup onboard the Unitree Go2 quadruped.}
    \label{fig:hardware_setup}
\end{figure}

A custom fabricated onboard mount carries all of the previously mentioned computing devices along with the necessary sensors onboard the Go2. This setup is visualized in Figure \ref{fig:hardware_setup}. 

\begin{figure*}[h!]
    \centering
    \includegraphics[width=0.95\linewidth]{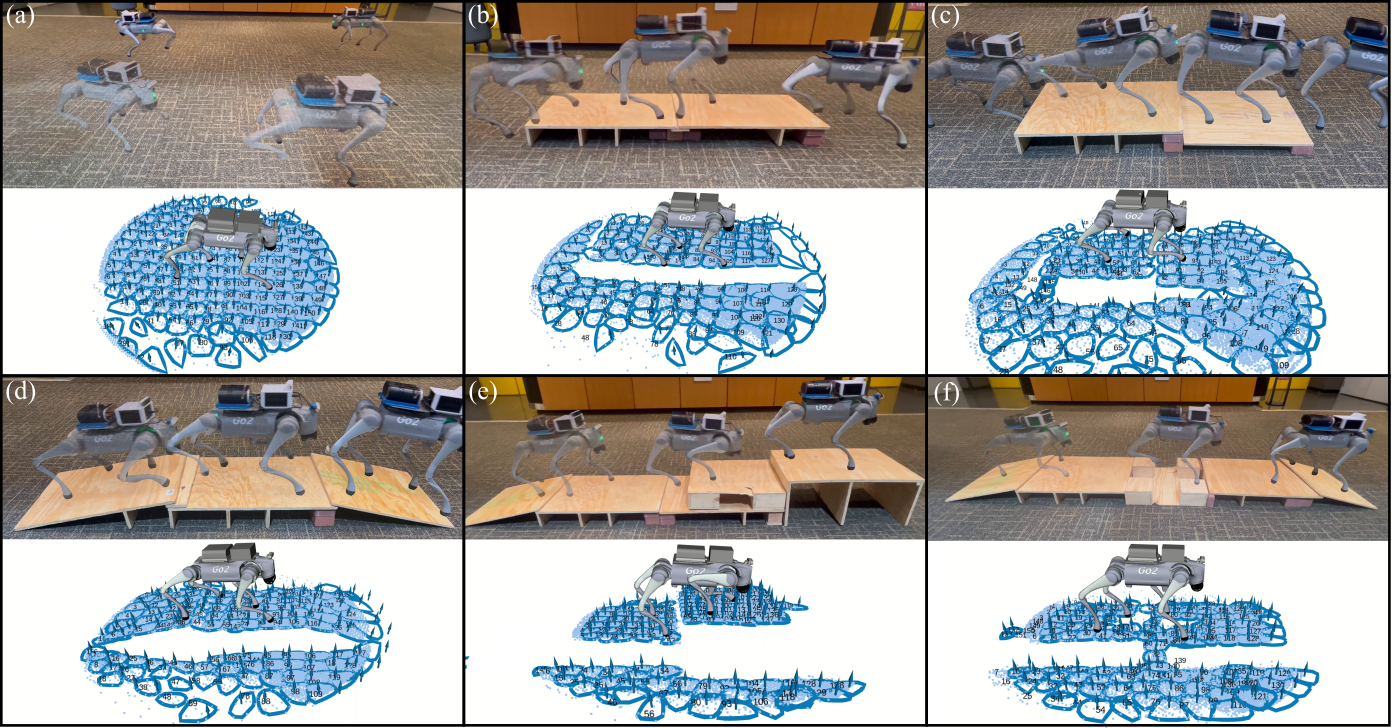}
    \caption{Visualizations of QuadPiPS hardware deployment in both indoor and outdoor settings. (a) Flat, (b) Single Platform, (c) Double platform, (d) Ramped single platform, (e) Triple platform, (f) Ramped stepping stones platform. Top images are the real world and bottom images are RViz. In RViz, superpixel regions are depicted by blue points, normals, and boundaries.}
    \label{fig:hardware_tracking_combined}
\end{figure*}

\subsection{Hardware Demonstrations} \label{sec:hardware_demonstrations}
The proposed work is deployed over real-world terrain setups. Visualizations are provided in Figure \ref{fig:hardware_tracking_combined} and full trials are shown in the accompanying video\footnote{\href{https://quadpips.github.io/}{https://quadpips.github.io/}}. A human operator is present alongside the robot during demonstrations, but they are not providing any support for the robot during deployment. 


\subsubsection{Perception Timing}

\begin{figure}[h!]
    \centering
    \includegraphics[width=0.99\linewidth]{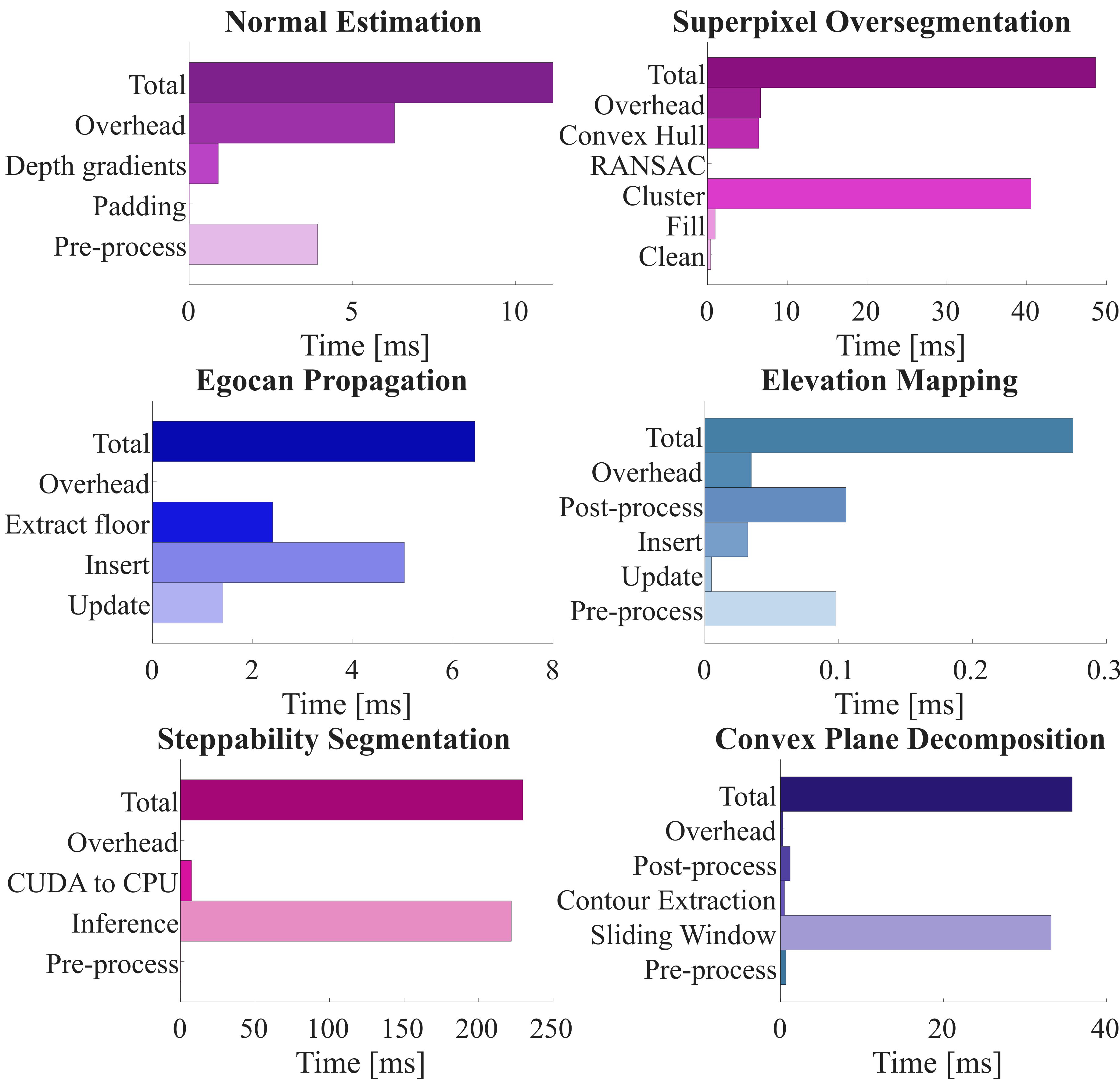}
    \caption{Average timing for each module in the proposed superpixels-based perception pipeline along with the elevation map-based perception pipeline. The steppability model is run in a separate update thread to prevent the large inference time from acting as a bottleneck on the entire pipeline.}
    \label{fig:quadpips_timing_hardware}
\end{figure}

First, timing data is taken on hardware and reported in this section as well as in Figure \ref{fig:quadpips_timing_hardware}. For the superpixels-based perception pipeline, timing on hardware is comparable to timing in simulation. The oversegmentation process runs at $20$~Hz. For the elevation mapping-based perception pipeline, the mapping process itself is remarkably fast, taking under one millisecond. However, the planar region decomposition process scales more aggressively, limiting the overall pipeline to $25$~Hz. Overall, the superpixels pipeline scales nicely onto hardware given its image-based framework.

\subsubsection{Deployment Results}
QuadPiPS plans over five different terrain configurations as well as one flat ground setting.

\textbf{Flat:} Here, QuadPiPS plans a circular trajectory over flat ground to demonstrate baseline performance. During turns, the visibility of the egocan floor is reduced due to the limited field of view of the camera. This manifests in partial loss of data for planar regions, as can be observed in Figure \ref{fig:hardware_tracking_combined}.

\textbf{Single platform:} Here, QuadPiPS plans over a raised platform that is $12$~cm tall. For reference, the quadruped itself has $23$~cm of clearance underneath the torso, and its nominal swing height is $10$~cm tall. To be able to step onto the platform with all four legs, the quadruped pitches its torso upwards to give its legs more clearance. For stepping down off the platform, a camera pointed directly forwards misses a significant portion of critical terrain information immediately in front of the robot. Therefore, the camera was pitched $45\degree$ downwards in order to capture this crucial data. The quadruped stumbles after stepping down off of the platform, but the MPC is able to stabilize the robot and continuing walking. 

\textbf{Double platform:} Here, QuadPiPS plans over one raised platform that is $12$~cm in height and second platform that is $5$~cm in height. As in the prior configuration, the planner performs torso pitch adjustment as it steps onto the terrain. This adjustment is slightly delayed, partially causing tracking mistakes on later steps. For instance, the front left leg plans a foothold on the edge of the first platform. However, during tracking, this foot slips off of the first platform and onto the second. This kind of perturbation is not captured in the QuadPiPS framework. The planned contact sequence is assumed to be followed exactly, so these missteps can lead to tracking failures. In this case, the perturbation was small enough for the MPC to ultimately recover. Although, these common tracking mishaps do motivate the use of more robust tracking controllers such as learned policies.

\textbf{Ramped single platform:} In this trial, QuadPiPS plans over a ramped version of the previously used platforms, here at a $10\degree$ incline and decline. While traversing up the ramp, the terrain boards shift and slide around underneath the weight of the quadruped. However, the superpixels-based perception pipeline reacts to these environmental changes and the low-level control rejects these disturbances. The terrain configuration itself is also not necessarily neat. Given the real-time perception, the terrain does not have to be conveniently organized. The board placed underneath the inclined ramp is partially exposed and the quadruped steps on this part of the board. This is all modeled by the perception and subsequently handled by the planning framework. The MPC is delayed in adjusting the robot's height as well as its pitch. This delay yields a low torso height as the quadruped ascends the terrain and a high torso height as it descends the terrain.

\textbf{Triple platform:} For this trial, QuadPiPS plans up one ramped platform with the previously states dimensions, a second platform that is $13$~cm tall (for a total height of $25$~cm), and a third platform that is $10$~cm tall (for a total height of $35$~cm). The entire terrain configuration is as tall as the robot itself. While ascending the stairs, QuadPiPS plans steps up against the lip of each platform. The terrain-aware swing trajectory augmentation active within OCS2 is used in both the reference trajectory generation and tracking to lift the swing feet over these sudden height increases. Though, performing additional inflation to ensure steps are not planned on the border of planar regions might be useful in environments such as these. As mentioned before, the terrain boards used in this configuration are chipped and warped, leading to significant deflection and displacement when stepped on. Despite this, the low-level control robustly tracks the footholds up the stairs.

\textbf{Ramped stepping stones platform:} Lastly, QuadPiPS plans over a ramped platform with the previously stated dimensions which follows into a set of stepping stones. Each stepping stone is $20\times20$~cm. The stones are separated center-to-center by $40$~cm. All four feet land on both the back and front sets of stones except for the hind right leg which swings over the back stone. This is an environment in which precise foothold planning is crucial. While nothing prevents QuadPiPS from planning footholds on the ground-level terrain around the stepping stones, quickly stepping down onto the ground and back up onto the platform after the stepping stones poses a challenge for stability. QuadPiPS opts to simply plan over the stepping stones to maintain a consistent torso height over the entire terrain configuration.  While descending the ramp after the stones, the front right foot slips at the junction between the platform and the ramp. However, this is a small enough perturbation for the tracking control to reject.

%% file: text_files/discussion.tex
\section{Discussion} \label{sec:discussion}

This work marks the first investigation into how the Planning in the Perception Space (PiPS) philosophy can be applied to quadrupedal foothold planning. The computational benefits from PiPS such as minimal sensor preprocessing and dense local depth information align well with the task of foothold planning. Foothold planning should only be done in close proximity to the ego-robot, so the task itself does not concern the sparser data further away from the robot. For similar reasons, minimizing latency for foothold planning is crucial.

Simulation environments (a)-(e) represent settings where precise foothold planning is not strictly necessary. If a desired foothold was not accurately tracked, there is neighboring terrain where the robot can step. The last five environment (f)-(j) all require more care for foothold placement, and this is exactly where the superpixels baselines outperform the elevation mapping baselines. Beyond this, QuadPiPS outperformed S-MPC in the last three environments with each environment highlighting a core benefit of QuadPiPS. For sparse stepping stones, heuristic foothold placement is insufficient. When available footholds are limited and do not subscribe to a particular pattern in the case of randomly placed terrain, a more exhaustive method such as searching or sampling is much more beneficial. For the obstructed balance beam, semantic steppability labels encode preferences for movement opportunities. This prevents QuadPiPS from planning footholds on the barriers which would disrupt the stability of the system. Lastly, the winding balance beam exemplifies how QuadPiPS excels through its telescopic framework. A guiding torso path informed by the local environment provides a waypoint for the foothold search, but the foothold sequence is not required to track the torso path. This is useful in environments where the torso path might lead the robot to portions of the environment where foothold planning is more difficult. This freedom can be viewed as robustness to incorrect or suboptimal commands.

It should also be said that superpixels are not necessary for all environments. Across the first five environments, EM-DTC was the only baseline to succeed on every trial. For environments such as the rubble or pegboard, precise foothold planning is not necessary. If the robot can scramble across the terrain and stay upright, it will traverse the terrain successfully. Furthermore, the non-uniform gridding from the oversegmentation can yield asymmetric footholds and inefficient locomotion through unnecessary adjustment from step to step.

With regards to limitations for the QuadPiPS framework, the number of superpixels as well as the poses and boundaries of each superpixel are variable between timesteps. One of the advantages of clustering-based algorithms is that when run over time, they can be warm-started with solutions from prior timesteps. Propagating cluster information over time and warm-starting the superpixels algorithm can mitigate the jitter observed between timesteps and expedite the overall runtime. Superpixels can also hang over edges and overlap with one another on occasion. Further image processing to score the density or convexity of regions could generate regions that are better suited for footholds. 

This work explored how to use the egocan floor for foothold planning, but the egocan ceiling can also inform planning about overhead environment information and allow the quadruped to crouch underneath obstacles. QuadPiPS could then plan through environments such as caves or tunnels.


The current contact mode family transition graph provides stance configurations that are kinematically feasible, but not necessarily dynamically feasible. Additional graph constraints such as stance stability are then used to encourage, but not enforce, dynamic feasibility. Re-introducing the experience heuristic from ALEF would provide further coordination between the search and optimization levels of QuadPiPS and ensure the dynamic feasibility of foothold sequences.

Lastly, the authors found that the robustness of the MPC and WBC restricted the breadth of hardware testing available for QuadPiPS. Replacing the low-level controllers with a learned policy such as DTC that can interface with the superpixels representation would allow for more robust tracking and more challenging terrain traversal including outdoor settings.

%% file: text_files/conclusion.tex
\section{Conclusion} \label{sec:conclusion}

This work presents QuadPiPS, a perception-informed framework for foothold planning in the perception space. The local environment representation known as the egocan is extended beyond its use for mobile robots to support legged environmental affordances --- surface normals and steppability labels --- and facilitate foothold searching through a superpixels-based oversegmentation. To construct simulation scenes with accessible ground truth steppability information, QuadPiPS adapts a primitive shapes-based synthetic data generation scheme from the field of grasp prediction. The oversegmented egocan floor is integrated into an extension of the Augmented Leafs with Experience on Foliations (ALEF) framework, which is modified in this work to support (a) perception-informed, (b) real-time, and (c) kinodynamically-feasible motion planning for quadrupeds. Simulation benchmarking across diverse environments and state-of-the-art planners shows that QuadPiPS excels in safety-critical environments where limited footholds are available, and real-world validation on challenging terrain setups indicates that the QuadPiPS framework is readily deployable on legged hardware platforms.

%% file: references.bib
@article{he_attention-based_2025,
	title = {Attention-based map encoding for learning generalized legged locomotion},
	copyright = {Copyright © 2025 The Authors, some rights reserved; exclusive licensee American Association for the Advancement of Science. No claim to original U.S. Government Works},
	language = {EN},
	urldate = {2025-09-10},
	journal = {Science Robotics},
	publisher = {American Association for the Advancement of Science},
	author = {He, Junzhe and Zhang, Chong and Jenelten, Fabian and Grandia, Ruben and Bächer, Moritz and Hutter, Marco},
	month = aug,
	year = {2025},
}

@article{fischler_random_1981,
	title = {Random sample consensus: a paradigm for model fitting with applications to image analysis and automated cartography},
	shorttitle = {Random sample consensus},
	urldate = {2026-04-20},
	journal = {Commun. ACM},
	author = {Fischler, Martin A. and Bolles, Robert C.},
	month = jun,
	year = {1981},
}

@incollection{althoefer_contact_2019,
	title = {Contact {Planning} for the {ANYmal} {Quadruped} {Robot} {Using} an {Acyclic} {Reachability}-{Based} {Planner}},
	language = {en},
	urldate = {2026-04-13},
	booktitle = {Towards {Autonomous} {Robotic} {Systems}},
	publisher = {Springer International Publishing},
	author = {Geisert, Mathieu and Yates, Thomas and Orgen, Asil and Fernbach, Pierre and Havoutis, Ioannis},
	editor = {Althoefer, Kaspar and Konstantinova, Jelizaveta and Zhang, Ketao},
	year = {2019},
}

@article{dixit_step_2025,
	title = {{STEP}: {Stochastic} {Traversability} {Evaluation} and {Planning} for {Risk}-{Aware} {Navigation}; {Results} {From} the {DARPA} {Subterranean} {Challenge}},
	shorttitle = {{STEP}},
	urldate = {2025-09-29},
	journal = {IEEE Transactions on Field Robotics},
	author = {Dixit, Anushri and Fan, David D. and Otsu, Kyohei and Dey, Sharmita and Agha-Mohammadi, Ali-Akbar and Burdick, Joel W.},
	year = {2025},
}

@article{hartley_contact-aided_2020,
	title = {Contact-aided invariant extended {Kalman} filtering for robot state estimation},
	journal = {The International Journal of Robotics Research},
	author = {Hartley, Ross and Ghaffari, Maani and Eustice, Ryan M. and Grizzle, Jessy W.},
	year = {2020},
}

@inproceedings{dario_bellicoso_perception-less_2016,
	title = {Perception-less terrain adaptation through whole body control and hierarchical optimization},
	urldate = {2025-11-21},
	booktitle = {2016 {IEEE}-{RAS} 16th {International} {Conference} on {Humanoid} {Robots} ({Humanoids})},
	author = {Dario Bellicoso, C. and Gehring, Christian and Hwangbo, Jemin and Fankhauser, Péter and Hutter, Marco},
	month = nov,
	year = {2016},
}

@article{orin_centroidal_2013,
	title = {Centroidal dynamics of a humanoid robot},
	journal = {Autonomous Robots},
	author = {Orin, David E. and Goswami, Ambarish and Lee, Sung-Hee},
	month = oct,
	year = {2013},
}

@incollection{ferrari_encoder-decoder_2018,
	title = {Encoder-{Decoder} with {Atrous} {Separable} {Convolution} for {Semantic} {Image} {Segmentation}},
	language = {en},
	urldate = {2026-04-13},
	booktitle = {{ECCV} 2018},
	publisher = {Springer International Publishing},
	author = {Chen, Liang-Chieh and Zhu, Yukun and Papandreou, George and Schroff, Florian and Adam, Hartwig},
	editor = {Ferrari, Vittorio and Hebert, Martial and Sminchisescu, Cristian and Weiss, Yair},
	year = {2018},
}

@inproceedings{nieuwenhuisen_shape-primitive_2012,
	title = {Shape-{Primitive} {Based} {Object} {Recognition} and {Grasping}},
	booktitle = {{ROBOTIK} 2012; 7th {German} {Conference} on {Robotics}},
	author = {Nieuwenhuisen, Matthias and Stueckler, Joerg and Berner, Alexander and Klein, Reinhard and Behnke, Sven},
	month = may,
	year = {2012},
}

@incollection{smith_real-time_2020,
	title = {Real-{Time} {Egocentric} {Navigation} {Using} {3D} {Sensing}},
	language = {en},
	booktitle = {Machine {Vision} and {Navigation}},
	author = {Smith, Justin S. and Feng, Shiyu and Lyu, Fanzhe and Vela, Patricio A.},
	editor = {Sergiyenko, Oleg and Flores-Fuentes, Wendy and Mercorelli, Paolo},
	year = {2020},
}

@article{semini_design_2011,
	title = {Design of {HyQ} – a hydraulically and electrically actuated quadruped robot},
	language = {EN},
	urldate = {2026-04-03},
	journal = {Proceedings of the Institution of Mechanical Engineers, Part I: Journal of Systems and Control Engineering},
	publisher = {IMECHE},
	author = {Semini, C and Tsagarakis, N G and Guglielmino, E and Focchi, M and Cannella, F and Caldwell, D G},
	month = sep,
	year = {2011},
}

@article{villarreal_fast_2019,
	title = {Fast and {Continuous} {Foothold} {Adaptation} for {Dynamic} {Locomotion} through {CNNs}},
	journal = {IEEE Robotics and Automation Letters},
	author = {Villarreal, Octavio and Barasuol, Victor and Camurri, Marco and Franceschi, Luca and Focchi, Michele and Pontil, Massimiliano and Caldwell, Darwin G. and Semini, Claudio},
	month = apr,
	year = {2019},
}

@article{wellhausen_artplanner_2023,
	title = {{ArtPlanner}: {Robust} {Legged} {Robot} {Navigation} in the {Field}},
	journal = {Field Robotics},
	author = {Wellhausen, Lorenz and Hutter, Marco},
	year = {2023},
}

@article{hornung_octomap_2013,
	title = {{OctoMap}: an efficient probabilistic {3D} mapping framework based on octrees},
	shorttitle = {{OctoMap}},
	language = {en},
	urldate = {2024-08-07},
	journal = {Autonomous Robots},
	author = {Hornung, Armin and Wurm, Kai M. and Bennewitz, Maren and Stachniss, Cyrill and Burgard, Wolfram},
	month = apr,
	year = {2013},
}

@article{hines_virtual_2021,
	title = {Virtual {Surfaces} and {Attitude} {Aware} {Planning} and {Behaviours} for {Negative} {Obstacle} {Navigation}},
	urldate = {2024-08-12},
	journal = {IEEE Robotics and Automation Letters},
	author = {Hines, Thomas and Stepanas, Kazys and Talbot, Fletcher and Sa, Inkyu and Lewis, Jake and Hernandez, Emili and Kottege, Navinda and Hudson, Nicolas},
	month = apr,
	year = {2021},
}

@article{zucker_chomp_2013,
	title = {{CHOMP}: {Covariant} {Hamiltonian} optimization for motion planning},
	issn = {0278-3649},
	shorttitle = {{CHOMP}},
	language = {en},
	journal = {The International Journal of Robotics Research},
	publisher = {SAGE Publications Ltd STM},
	author = {Zucker, Matt and Ratliff, Nathan and Dragan, Anca D. and Pivtoraiko, Mihail and Klingensmith, Matthew and Dellin, Christopher M. and Bagnell, J. Andrew and Srinivasa, Siddhartha S.},
	month = aug,
	year = {2013},
}

@article{pippine_overview_2011,
	title = {An overview of the {Defense} {Advanced} {Research} {Projects} {Agency}’s {Learning} {Locomotion} program},
	journal = {The International Journal of Robotics Research},
	author = {Pippine, James and Hackett, Douglas and Watson, Adam},
	month = feb,
	year = {2011},
}

@article{buchanan_perceptive_2021,
	title = {Perceptive whole-body planning for multilegged robots in confined spaces},
	journal = {Journal of Field Robotics},
	author = {Buchanan, Russell and Wellhausen, Lorenz and Bjelonic, Marko and Bandyopadhyay, Tirthankar and Kottege, Navinda and Hutter, Marco},
	year = {2021},
}

@inproceedings{bertrand_detecting_2020,
	address = {Las Vegas, NV, USA},
	title = {Detecting {Usable} {Planar} {Regions} for {Legged} {Robot} {Locomotion}},
	booktitle = {2020 {IEEE}/{RSJ} {International} {Conference} on {Intelligent} {Robots} and {Systems} ({IROS})},
	publisher = {IEEE},
	author = {Bertrand, Sylvain and Lee, Inho and Mishra, Bhavyansh and Calvert, Duncan and Pratt, Jerry and Griffin, Robert},
	month = oct,
	year = {2020},
}

@inproceedings{kim_vision_2020,
	title = {Vision {Aided} {Dynamic} {Exploration} of {Unstructured} {Terrain} with a {Small}-{Scale} {Quadruped} {Robot}},
	booktitle = {2020 {IEEE} {International} {Conference} on {Robotics} and {Automation} ({ICRA})},
	author = {Kim, D. and Carballo, D. and Di Carlo, J. and Katz, B. and Bledt, G. and Lim, B. and Kim, S.},
	month = may,
	year = {2020},
}

@inproceedings{karkowski_prediction_2019,
	title = {Prediction {Maps} for {Real}-{Time} {3D} {Footstep} {Planning} in {Dynamic} {Environments}},
	booktitle = {2019 {International} {Conference} on {Robotics} and {Automation} ({ICRA})},
	author = {Karkowski, Philipp and Bennewitz, Maren},
	month = may,
	year = {2019},
}

@article{miller_mine_2020,
	title = {Mine {Tunnel} {Exploration} {Using} {Multiple} {Quadrupedal} {Robots}},
	volume = {5},
	issn = {2377-3766},
	doi = {10.1109/LRA.2020.2972872},
	number = {2},
	urldate = {2026-04-13},
	journal = {IEEE Robotics and Automation Letters},
	author = {Miller, Ian D. and Cladera, Fernando and Cowley, Anthony and Shivakumar, Shreyas S. and Lee, Elijah S. and Jarin-Lipschitz, Laura and Bhat, Akhilesh and Rodrigues, Neil and Zhou, Alex and Cohen, Avraham and Kulkarni, Adarsh and Laney, James and Taylor, Camillo Jose and Kumar, Vijay},
	month = apr,
	year = {2020},
	pages = {2840--2847},
}

@article{jenelten_perceptive_2020,
	title = {Perceptive {Locomotion} in {Rough} {Terrain} – {Online} {Foothold} {Optimization}},
	journal = {IEEE Robotics and Automation Letters},
	author = {Jenelten, Fabian and Miki, Takahiro and Vijayan, Aravind E and Bjelonic, Marko and Hutter, Marco},
	month = oct,
	year = {2020},
}

@inproceedings{smith_aerialpips_2023,
	title = {{AeriaLPiPS}: {A} {Local} {Planner} for {Aerial} {Vehicles} with {Geometric} {Collision} {Checking}},
	shorttitle = {{AeriaLPiPS}},
	urldate = {2025-01-15},
	booktitle = {2023 {IEEE} {International} {Conference} on {Robotics} and {Automation} ({ICRA})},
	author = {Smith, Justin S. and Vela, Patricio},
	month = may,
	year = {2023},
}

@inproceedings{perille_benchmarking_2020,
	address = {Abu Dhabi, UAE},
	title = {Benchmarking {Metric} {Ground} {Navigation}},
	urldate = {2026-04-12},
	booktitle = {{IEEE} {International} {Symposium} on {Safety}, {Security}, and {Rescue} {Robotics} ({SSRR})},
	author = {Perille, Daniel and Truong, Abigail and Xiao, Xuesu and Stone, Peter},
	month = nov,
	year = {2020},
}

@article{jenelten_tamols_2022,
	title = {{TAMOLS}: {Terrain}-{Aware} {Motion} {Optimization} for {Legged} {Systems}},
	shorttitle = {{TAMOLS}},
	urldate = {2026-04-13},
	journal = {IEEE Transactions on Robotics},
	author = {Jenelten, Fabian and Grandia, Ruben and Farshidian, Farbod and Hutter, Marco},
	month = dec,
	year = {2022},
}

@inproceedings{agarwal_legged_2023,
	title = {Legged {Locomotion} in {Challenging} {Terrains} using {Egocentric} {Vision}},
	language = {en},
	urldate = {2026-04-13},
	booktitle = {Proceedings of {The} 6th {Conference} on {Robot} {Learning}},
	author = {Agarwal, Ananye and Kumar, Ashish and Malik, Jitendra and Pathak, Deepak},
	month = mar,
	year = {2023},
}

@article{sorokin_learning_2022,
	title = {Learning to {Navigate} {Sidewalks} in {Outdoor} {Environments}},
	urldate = {2026-04-13},
	journal = {IEEE Robotics and Automation Letters},
	author = {Sorokin, Maks and Tan, Jie and Liu, C. Karen and Ha, Sehoon},
	month = apr,
	year = {2022},
}

@article{chen_learning_2025,
	title = {Learning {Autonomous} and {Safe} {Quadruped} {Traversal} of {Complex} {Terrains} {Using} {Multi}-{Layer} {Elevation} {Maps}},
	issn = {2377-3766},
	urldate = {2026-04-03},
	journal = {IEEE Robotics and Automation Letters},
	author = {Chen, Yeke and Ma, Ji and Luo, Zeren and Han, Yimin and Dong, Yinzhao and Xu, Bowen and Lu, Peng},
	month = oct,
	year = {2025},
}

@article{boaventura_model-based_2015,
	title = {Model-{Based} {Hydraulic} {Impedance} {Control} for {Dynamic} {Robots}},
	issn = {1941-0468},
	urldate = {2026-04-03},
	journal = {IEEE Transactions on Robotics},
	author = {Boaventura, Thiago and Buchli, Jonas and Semini, Claudio and Caldwell, Darwin G.},
	month = dec,
	year = {2015},
}

@article{fahmi_passive_2019,
	title = {Passive {Whole}-{Body} {Control} for {Quadruped} {Robots}: {Experimental} {Validation} {Over} {Challenging} {Terrain}},
	shorttitle = {Passive {Whole}-{Body} {Control} for {Quadruped} {Robots}},
	urldate = {2026-04-13},
	journal = {IEEE Robotics and Automation Letters},
	author = {Fahmi, Shamel and Mastalli, Carlos and Focchi, Michele and Semini, Claudio},
	month = jul,
	year = {2019},
}

@article{fahmi_vital_2023,
	title = {{ViTAL}: {Vision}-{Based} {Terrain}-{Aware} {Locomotion} for {Legged} {Robots}},
	issn = {1552-3098, 1941-0468},
	shorttitle = {{ViTAL}},
	urldate = {2024-08-08},
	journal = {IEEE Transactions on Robotics},
	author = {Fahmi, Shamel and Barasuol, Victor and Esteban, Domingo and Villarreal, Octavio and Semini, Claudio},
	month = apr,
	year = {2023},
}

@article{kim_plgrim_2021,
	title = {{PLGRIM}: {Hierarchical} {Value} {Learning} for {Large}-scale {Exploration} in {Unknown} {Environments}},
	shorttitle = {{PLGRIM}},
	language = {en},
	urldate = {2026-04-13},
	journal = {Proceedings of the International Conference on Automated Planning and Scheduling},
	author = {Kim, Sung-Kyun and Bouman, Amanda and Salhotra, Gautam and Fan, David D. and Otsu, Kyohei and Burdick, Joel and Agha-mohammadi, Ali-akbar},
	month = may,
	year = {2021},
}

@inproceedings{bouman_autonomous_2020,
	title = {Autonomous {Spot}: {Long}-{Range} {Autonomous} {Exploration} of {Extreme} {Environments} with {Legged} {Locomotion}},
	shorttitle = {Autonomous {Spot}},
	urldate = {2026-04-13},
	booktitle = {2020 {IEEE}/{RSJ} {International} {Conference} on {Intelligent} {Robots} and {Systems} ({IROS})},
	author = {Bouman, Amanda and Ginting, Muhammad Fadhil and Alatur, Nikhilesh and Palieri, Matteo and Fan, David D. and Touma, Thomas and Pailevanian, Torkom and Kim, Sung-Kyun and Otsu, Kyohei and Burdick, Joel and Agha-Mohammadi, Ali-akbar},
	month = oct,
	year = {2020},
	pages = {2518--2525},
}

@article{lee_learning_2020,
	title = {Learning {Quadrupedal} {Locomotion} over {Challenging} {Terrain}},
	urldate = {2022-10-12},
	journal = {Science Robotics},
	author = {Lee, Joonho and Hwangbo, Jemin and Wellhausen, Lorenz and Koltun, Vladlen and Hutter, Marco},
	month = oct,
	year = {2020},
}

@inproceedings{kulkarni_autonomous_2022,
	title = {Autonomous {Teamed} {Exploration} of {Subterranean} {Environments} using {Legged} and {Aerial} {Robots}},
	urldate = {2026-04-13},
	booktitle = {2022 {International} {Conference} on {Robotics} and {Automation} ({ICRA})},
	author = {Kulkarni, Mihir and Dharmadhikari, Mihir and Tranzatto, Marco and Zimmermann, Samuel and Reijgwart, Victor and De Petris, Paolo and Nguyen, Huan and Khedekar, Nikhil and Papachristos, Christos and Ott, Lionel and Siegwart, Roland and Hutter, Marco and Alexis, Kostas},
	month = may,
	year = {2022},
}

@inproceedings{acosta_bipedal_2023,
	title = {Bipedal {Walking} on {Constrained} {Footholds} with {MPC} {Footstep} {Control}},
	urldate = {2026-04-13},
	publisher = {arXiv},
	author = {Acosta, Brian and Posa, Michael},
	month = sep,
	year = {2023},
}

@misc{zhang_ame-2_2026,
	title = {{AME}-2: {Agile} and {Generalized} {Legged} {Locomotion} via {Attention}-{Based} {Neural} {Map} {Encoding}},
	shorttitle = {{AME}-2},
	doi = {10.48550/arXiv.2601.08485},
	urldate = {2026-04-13},
	publisher = {arXiv},
	author = {Zhang, Chong and Klemm, Victor and Yang, Fan and Hutter, Marco},
	month = mar,
	year = {2026},
}

@article{yoon_state-nav_2026,
	title = {{STATE}-{NAV}: {Stability}-{Aware} {Traversability} {Estimation} for {Bipedal} {Navigation} on {Rough} {Terrain}},
	urldate = {2026-04-13},
	journal = {IEEE Robotics and Automation Letters},
	author = {Yoon, Ziwon and Zhu, Lawrence Y. and Lu, Jingxi and Gan, Lu and Zhao, Ye},
	month = feb,
	year = {2026},
}

@inproceedings{fang_saga_2025,
	title = {{SAGA}: {Open}-{World} {Mobile} {Manipulation} via {Structured} {Affordance} {Grounding}},
	shorttitle = {{SAGA}},
	urldate = {2026-04-13},
	publisher = {arXiv},
	author = {Fang, Kuan and Chen, Yuxin and Zhu, Xinghao and Niroui, Farzad and Sun, Lingfeng and Wang, Jiuguang},
	month = dec,
	year = {2025},
}

@article{fallon_architecture_2015,
	title = {An {Architecture} for {Online} {Affordance}-based {Perception} and {Whole}-body {Planning}},
	copyright = {© 2014 Wiley Periodicals, Inc.},
	issn = {1556-4967},
	doi = {10.1002/rob.21546},
	language = {en},
	urldate = {2026-04-13},
	journal = {Journal of Field Robotics},
	author = {Fallon, Maurice and Kuindersma, Scott and Karumanchi, Sisir and Antone, Matthew and Schneider, Toby and Dai, Hongkai and D'Arpino, Claudia Pérez and Deits, Robin and DiCicco, Matt and Fourie, Dehann and Koolen, Twan and Marion, Pat and Posa, Michael and Valenzuela, Andrés and Yu, Kuan-Ting and Shah, Julie and Iagnemma, Karl and Tedrake, Russ and Teller, Seth},
	year = {2015},
}

@inproceedings{miki_elevation_2022,
	title = {Elevation {Mapping} for {Locomotion} and {Navigation} using {GPU}},
	urldate = {2026-04-13},
	booktitle = {2022 {IEEE}/{RSJ} {International} {Conference} on {Intelligent} {Robots} and {Systems} ({IROS})},
	author = {Miki, Takahiro and Wellhausen, Lorenz and Grandia, Ruben and Jenelten, Fabian and Homberger, Timon and Hutter, Marco},
	month = oct,
	year = {2022},
}

@article{agha_nebula_2022,
	title = {{NeBula}: {TEAM} {CoSTAR}'s {Robotic} {Autonomy} {Solution} that {Won} {Phase} {II} of {DARPA} {Subterranean} {Challenge}},
	issn = {2771-3989},
	shorttitle = {{NeBula}},
	urldate = {2026-04-13},
	journal = {Field Robotics},
	author = {Agha, Ali and Otsu, Kyohei and Morrell, Benjamin and Fan, David D. and Thakker, Rohan and Santamaria-Navarro, Angel and Kim, Sung-Kyun and Bouman, Amanda and Lei, Xianmei and Edlund, Jeffrey and Ginting, Muhammad Fadhil and Ebadi, Kamak and Anderson, Matthew and Pailevanian, Torkom and Terry, Edward and Wolf, Michael and Tagliabue, Andrea and Vaquero, Tiago Stegun and Palieri, Matteo and Tepsuporn, Scott and Chang, Yun and Kalantari, Arash and Chavez, Fernando and Lopez, Brett and Funabiki, Nobuhiro and Miles, Gregory and Touma, Thomas and Buscicchio, Alessandro and Tordesillas, Jesus and Alatur, Nikhilesh and Nash, Jeremy and Walsh, William and Jung, Sunggoo and Lee, Hanseob and Kanellakis, Christoforos and Mayo, John and Harper, Scott and Kaufmann, Marcel and Dixit, Anushri and Correa, Gustavo J. and Lee, Carlyn and Gao, Jay and Merewether, Gene and Maldonado-Contreras, Jairo and Salhotra, Gautam and Da Silva, Maira Saboia and Ramtoula, Benjamin and Fakoorian, Seyed and Hatteland, Alexander and Kim, Taeyeon and Bartlett, Tara and Stephens, Alex and Kim, Leon and Bergh, Chuck and Heiden, Eric and Lew, Thomas and Cauligi, Abhishek and Heywood, Tristan and Kramer, Andrew and Leopold, Henry A. and Melikyan, Hov and Choi, Hyungho Chris and Daftry, Shreyansh and Toupet, Olivier and Wee, Inhwan and Thakur, Abhishek and Feras, Micah and Beltrame, Giovanni and Nikolakopoulos, George and Shim, David and Carlone, Luca and Burdick, Joel},
	month = jul,
	year = {2022},
}

@inproceedings{miki_learning_2024,
	title = {Learning to walk in confined spaces using {3D} representation},
	urldate = {2026-04-13},
	booktitle = {2024 {IEEE} {International} {Conference} on {Robotics} and {Automation} ({ICRA})},
	author = {Miki, Takahiro and Lee, Joonho and Wellhausen, Lorenz and Hutter, Marco},
	month = may,
	year = {2024},
}

@article{hoeller_neural_2022,
	title = {Neural {Scene} {Representation} for {Locomotion} on {Structured} {Terrain}},
	issn = {2377-3766},
	urldate = {2026-04-13},
	journal = {IEEE Robotics and Automation Letters},
	author = {Hoeller, David and Rudin, Nikita and Choy, Christopher and Anandkumar, Animashree and Hutter, Marco},
	month = oct,
	year = {2022},
}

@article{jacoff_taking_nodate,
	title = {Taking the {First} {Step} {Toward} {Autonomous} {Quadruped} {Robots}: {The} {Quadruped} {Robot} {Challenge} at {ICRA} 2023 in {London} [{Competitions}]},
	issn = {1558-223X},
	shorttitle = {Taking the {First} {Step} {Toward} {Autonomous} {Quadruped} {Robots}},
	doi = {10.1109/MRA.2023.3293296},
	urldate = {2026-04-02},
	journal = {IEEE Robotics \& Automation Magazine},
	author = {Jacoff, Adam and Jeon, Jeongmin and Huke, Oliver and Kanoulas, Dimitrios and Ha, Seehoon and Kim, Donghyun and Moon, Hyungpil},
}

@inproceedings{robotics_fauna_2026,
	title = {Fauna {Sprout}: {A} lightweight, approachable, developer-ready humanoid robot},
	shorttitle = {Fauna {Sprout}},
	urldate = {2026-04-13},
	publisher = {arXiv},
	author = {Robotics, Fauna and Aldarondo, Diego and Pervan, Ana and Corbalan, Daniel and Petrillo, Dave and Dai, Bolun and Iyer, Aadhithya and Mortensen, Nina and Pearson, Erik and Arunachalam, Sridhar Pandian and Reznick, Emma and Weis, David and Davison, Jacob and Patterson, Samuel and Carella, Tess and Suguitan, Michael and Ye, David and Ferro, Oswaldo and Suriyarachchi, Nilesh and Ling, Spencer and Su, Erik and Giebisch, Daniel and Traver, Peter and Fonseca, Sam and Mor, Mack and Singh, Rohan and Guven, Sertac and Liu, Kangni and Orru, Yaswanth Kumar and Batcha, Ashiq Rahman Anwar and Ravindranath, Shruthi and Arora, Silky and Ponte, Hugo and Hernandez, Dez and Chaudhary, Utsav and Walker, Zack and Kelberman, Michael and Veloz, Ivan and Lucia, Christina Santa and Casale, Kat and Han, Helen and Gromis, Michael and Mignatti, Michael and Reisman, Jason and Guerin, Kelleher and Narvaez, Dario and Anderson, Christopher and Moschella, Anthony and Cochran, Robert and Merel, Josh},
	month = jan,
	year = {2026},
}

@misc{yuxin_wu_detectron2_2019,
	title = {Detectron2},
	url = {https://github.com/facebookresearch/detectron2},
	author = {{Yuxin Wu} and {Alexander Kirillov} and {Francisco Massa} and {Wan-Yen Lo} and {Ross Girshick}},
	year = {2019},
}

@misc{grunnet-jepsen_depth_nodate,
	title = {Depth {Post}-{Processing} for {Intel}® {RealSense}™ {D400} {Depth} {Cameras}},
	language = {en},
	author = {Grunnet-Jepsen, Anders and Tong, Dave},
}

@inproceedings{nair_dynabarn_2022,
	title = {{DynaBARN}: {Benchmarking} {Metric} {Ground} {Navigation} in {Dynamic} {Environments}},
	booktitle = {2022 {IEEE} {International} {Symposium} on {Safety}, {Security}, and {Rescue} {Robotics} ({SSRR})},
	author = {Nair, Anirudh and Jiang, Fulin and Hou, Kang and Xu, Zifan and Li, Shuozhe and Xiao, Xuesu and Stone, Peter},
	month = nov,
	year = {2022},
}

@article{chung_into_2023,
	title = {Into the {Robotic} {Depths}: {Analysis} and {Insights} from the {DARPA} {Subterranean} {Challenge}},
	journal = {Annual Review of Control, Robotics, and Autonomous Systems},
	author = {Chung, Timothy H. and Orekhov, Viktor and Maio, Angela},
	month = may,
	year = {2023},
}

@misc{darpa_subterranean_nodate,
	title = {Subterranean {Challenge} {Final} {Event}},
	url = {https://www.darpa.mil/research/challenges/subterranean},
	urldate = {2025-05-26},
	author = {{DARPA}},
}

@misc{boston_dynamics_spot_nodate,
	title = {Spot},
	url = {https://bostondynamics.com/products/spot/},
	language = {en-US},
	urldate = {2026-04-02},
	author = {{Boston Dynamics}},
}

@misc{boston_dynamics_atlas_nodate,
	title = {Atlas {Humanoid} {Robot}},
	url = {https://bostondynamics.com/products/atlas/},
	language = {en-US},
	urldate = {2026-04-02},
	author = {{Boston Dynamics}},
}

@article{hutter_anymal_2017,
	title = {{ANYmal} - toward legged robots for harsh environments},
	journal = {Advanced Robotics},
	author = {Hutter, M. and Gehring, C. and Lauber, A. and Gunther, F. and Bellicoso, C. D. and Tsounis, V. and Fankhauser, P. and Diethelm, R. and Bachmann, S. and Bloesch, M. and Kolvenbach, H. and Bjelonic, M. and Isler, L. and Meyer, K.},
	month = sep,
	year = {2017},
}

@inproceedings{chignoli_mit_2021,
	title = {The {MIT} {Humanoid} {Robot}: {Design}, {Motion} {Planning}, and {Control} {For} {Acrobatic} {Behaviors}},
	booktitle = {2020 {IEEE}-{RAS} 20th {International} {Conference} on {Humanoid} {Robots} ({Humanoids})},
	author = {Chignoli, Matthew and Kim, Donghyun and Stanger-Jones, Elijah and Kim, Sangbae},
	month = jul,
	year = {2021},
}

@inproceedings{bledt_mit_2018,
	title = {{MIT} {Cheetah} 3: {Design} and {Control} of a {Robust}, {Dynamic} {Quadruped} {Robot}},
	booktitle = {2018 {IEEE}/{RSJ} {International} {Conference} on {Intelligent} {Robots} and {Systems} ({IROS})},
	author = {Bledt, Gerardo and Powell, Matthew J. and Katz, Benjamin and Di Carlo, Jared and Wensing, Patrick M. and Kim, Sangbae},
	month = oct,
	year = {2018},
}

@misc{unitree_robotics_robot_nodate,
	title = {Robot {Dog} {Go2}},
	url = {https://www.unitree.com/go2},
	urldate = {2026-04-02},
	author = {{Unitree Robotics}},
}

@misc{unitree_robotics_humanoid_nodate,
	title = {Humanoid robot {G1}},
	url = {https://www.unitree.com/g1},
	urldate = {2026-04-02},
	author = {{Unitree Robotics}},
}

@misc{noauthor_hello_nodate,
	title = {Hello, {Electric} {Atlas} - {IEEE} {Spectrum}},
	url = {https://spectrum.ieee.org/atlas-humanoid-robot},
	language = {en},
	urldate = {2026-04-02},
}

@article{zhu_artemis_nodate,
	title = {{ARTEMIS}: {An} {Open}-{Source}, {Full}-{Sized} {Humanoid} {Robot} for {Dynamic} {Locomotion}},
	language = {en},
	author = {Zhu, Taoyuanmin and Ahn, Min Sung and Hong, Dennis},
}

@article{agha-mohammadi_confidence-rich_2019,
	title = {Confidence-rich grid mapping},
	volume = {38},
	issn = {0278-3649, 1741-3176},
	number = {12-13},
	urldate = {2024-08-07},
	journal = {The International Journal of Robotics Research},
	author = {Agha-mohammadi, Ali-akbar and Heiden, Eric and Hausman, Karol and Sukhatme, Gaurav S.},
	month = oct,
	year = {2019},
	pages = {1352--1374},
}

@article{miki_learning_2022,
	title = {Learning robust perceptive locomotion for quadrupedal robots in the wild},
	volume = {7},
	number = {62},
	urldate = {2024-08-08},
	journal = {Science Robotics},
	author = {Miki, Takahiro and Lee, Joonho and Hwangbo, Jemin and Wellhausen, Lorenz and Koltun, Vladlen and Hutter, Marco},
	month = jan,
	year = {2022},
}

@article{jenelten_dtc_2024,
	title = {{DTC}: {Deep} {Tracking} {Control}},
	volume = {9},
	shorttitle = {{DTC}},
	number = {86},
	urldate = {2024-08-07},
	journal = {Science Robotics},
	author = {Jenelten, Fabian and He, Junzhe and Farshidian, Farbod and Hutter, Marco},
	month = jan,
	year = {2024},
}

@inproceedings{weikersdorfer_depth-adaptive_2012,
	title = {Depth-adaptive superpixels},
	urldate = {2026-01-12},
	booktitle = {Proceedings of the 21st {International} {Conference} on {Pattern} {Recognition} ({ICPR2012})},
	author = {Weikersdorfer, David and Gossow, David and Beetz, Michael},
	month = nov,
	year = {2012},
	pages = {2087--2090},
}

@inproceedings{pardo-castellote_omg_2003,
	title = {{OMG} {Data}-{Distribution} {Service}: architectural overview},
	shorttitle = {{OMG} {Data}-{Distribution} {Service}},
	urldate = {2025-11-23},
	booktitle = {23rd {International} {Conference} on {Distributed} {Computing} {Systems} {Workshops}, 2003. {Proceedings}.},
	author = {Pardo-Castellote, G.},
	month = may,
	year = {2003},
	pages = {200--206},
}

@article{macenski_robot_2022,
	title = {Robot {Operating} {System} 2: {Design}, architecture, and uses in the wild},
	volume = {7},
	shorttitle = {Robot {Operating} {System} 2},
	number = {66},
	urldate = {2025-11-23},
	journal = {Science Robotics},
	publisher = {American Association for the Advancement of Science},
	author = {Macenski, Steven and Foote, Tully and Gerkey, Brian and Lalancette, Chris and Woodall, William},
	month = may,
	year = {2022},
}

@article{hart_formal_1968,
	title = {A {Formal} {Basis} for the {Heuristic} {Determination} of {Minimum} {Cost} {Paths}},
	volume = {4},
	number = {2},
	journal = {IEEE Transactions on Systems Science and Cybernetics},
	publisher = {Institute of Electrical and Electronics Engineers (IEEE)},
	author = {Hart, Peter and Nilsson, Nils and Raphael, Bertram},
	year = {1968},
	pages = {100--107},
}

@article{graham_efficient_1972,
	title = {An efficient algorithm for determining the convex hull of a finite planar set},
	volume = {1},
	language = {en},
	number = {4},
	urldate = {2025-11-19},
	journal = {Information Processing Letters},
	author = {Graham, R.L.},
	month = jun,
	year = {1972},
	pages = {132--133},
}

@inproceedings{weikersdorfer_depth-adaptive_2013,
	title = {Depth-adaptive supervoxels for {RGB}-{D} video segmentation},
	urldate = {2026-01-12},
	booktitle = {2013 {IEEE} {International} {Conference} on {Image} {Processing}},
	author = {Weikersdorfer, David and Schick, Alexander and Cremers, Daniel},
	month = sep,
	year = {2013},
	pages = {2708--2712},
}

@inproceedings{neubert_compact_2014,
	title = {Compact {Watershed} and {Preemptive} {SLIC}: {On} {Improving} {Trade}-offs of {Superpixel} {Segmentation} {Algorithms}},
	shorttitle = {Compact {Watershed} and {Preemptive} {SLIC}},
	urldate = {2025-01-13},
	booktitle = {2014 22nd {International} {Conference} on {Pattern} {Recognition}},
	author = {Neubert, Peer and Protzel, Peter},
	month = aug,
	year = {2014},
	pages = {996--1001},
}

@inproceedings{nakagawa_estimating_2015,
	title = {Estimating {Surface} {Normals} with {Depth} {Image} {Gradients} for {Fast} and {Accurate} {Registration}},
	urldate = {2025-02-13},
	booktitle = {2015 {International} {Conference} on {3D} {Vision}},
	author = {Nakagawa, Yosuke and Uchiyama, Hideaki and Nagahara, Hajime and Taniguchi, Rin-Ichiro},
	month = oct,
	year = {2015},
	pages = {640--647},
}

@inproceedings{matthies_stereo_2014,
	title = {Stereo vision-based obstacle avoidance for micro air vehicles using disparity space},
	urldate = {2025-02-18},
	booktitle = {2014 {IEEE} {International} {Conference} on {Robotics} and {Automation} ({ICRA})},
	author = {Matthies, Larry and Brockers, Roland and Kuwata, Yoshiaki and Weiss, Stephan},
	month = may,
	year = {2014},
	pages = {3242--3249},
}

@inproceedings{frey_locomotion_2022,
	title = {Locomotion {Policy} {Guided} {Traversability} {Learning} using {Volumetric} {Representations} of {Complex} {Environments}},
	urldate = {2024-08-07},
	booktitle = {2022 {IEEE}/{RSJ} {International} {Conference} on {Intelligent} {Robots} and {Systems} ({IROS})},
	author = {Frey, Jonas and Hoeller, David and Khattak, Shehryar and Hutter, Marco},
	month = oct,
	year = {2022},
	pages = {5722--5729},
}

@inproceedings{oleynikova_voxblox_2017,
	title = {Voxblox: {Incremental} {3D} {Euclidean} {Signed} {Distance} {Fields} for {On}-{Board} {MAV} {Planning}},
	shorttitle = {Voxblox},
	language = {en},
	urldate = {2024-05-02},
	booktitle = {2017 {IEEE}/{RSJ} {International} {Conference} on {Intelligent} {Robots} and {Systems} ({IROS})},
	author = {Oleynikova, Helen and Taylor, Zachary and Fehr, Marius and Nieto, Juan and Siegwart, Roland},
	month = sep,
	year = {2017},
	pages = {1366--1373},
}

@article{zucker_optimization_2011,
	title = {Optimization and learning for rough terrain legged locomotion},
	volume = {30},
	language = {en},
	number = {2},
	journal = {The International Journal of Robotics Research},
	author = {Zucker, Matt and Ratliff, Nathan and Stolle, Martin and Chestnutt, Joel and Bagnell, J Andrew and Atkeson, Christopher G and Kuffner, James},
	month = feb,
	year = {2011},
	pages = {175--191},
}

@article{neuhaus_comprehensive_2011,
	title = {Comprehensive summary of the {Institute} for {Human} and {Machine} {Cognition}’s                 experience with {LittleDog}},
	volume = {30},
	language = {en},
	number = {2},
	urldate = {2024-08-08},
	journal = {The International Journal of Robotics Research},
	publisher = {SAGE Publications Ltd STM},
	author = {Neuhaus, Peter D and Pratt, Jerry E and Johnson, Matthew J},
	month = feb,
	year = {2011},
	pages = {216--235},
}

@inproceedings{behnke_team_2015,
	address = {Seoul, South Korea},
	title = {Team {NimbRo} {Rescue} at {DARPA} {Robotics} {Challenge} {Finals}},
	language = {en},
	urldate = {2024-08-12},
	booktitle = {2015 {IEEE}-{RAS} 15th {International} {Conference} on {Humanoid} {Robots} ({Humanoids})},
	publisher = {IEEE},
	author = {Behnke, Sven and Schwarz, Max and Rodehutskors, Tobias and Droeschel, David and Schreiber, Michael and Topelidou-Kyniazopoulou, Angeliki and Schwarz, David and Lenz, Christian and Schuller, Sebastian and Razlaw, Jan and Ivanov, Ivan and Araslanov, Nikita and Beul, Marius},
	month = nov,
	year = {2015},
	pages = {554--554},
}

@inproceedings{fankhauser_robot-centric_2014,
	address = {Poznan, Poland},
	title = {Robot-{Centric} {Elevation} {Mapping} with {Uncertainty} {Estimates}},
	language = {en},
	urldate = {2024-05-02},
	booktitle = {Mobile {Service} {Robotics}},
	publisher = {WORLD SCIENTIFIC},
	author = {Fankhauser, P. and Bloesch, M. and Gehring, C. and Hutter, M. and Siegwart, R.},
	month = aug,
	year = {2014},
	pages = {433--440},
}

@article{thrun_learning_2003,
	title = {Learning {Occupancy} {Grid} {Maps} with {Forward} {Sensor} {Models}},
	volume = {15},
	issn = {1573-7527},
	language = {en},
	number = {2},
	journal = {Autonomous Robots},
	author = {Thrun, Sebastian},
	month = sep,
	year = {2003},
	pages = {111--127},
}

@misc{zhang_h2oflow_2025,
	title = {{H2OFlow}: {Grounding} {Human}-{Object} {Affordances} with {3D} {Generative} {Models} and {Dense} {Diffused} {Flows}},
	shorttitle = {{H2OFlow}},
	publisher = {arXiv},
	author = {Zhang, Harry and Carlone, Luca},
	month = oct,
	year = {2025},
}

@article{hoeller_anymal_2024,
	title = {{ANYmal} parkour: {Learning} agile navigation for quadrupedal robots},
	volume = {9},
	shorttitle = {{ANYmal} parkour},
	number = {88},
	urldate = {2026-02-25},
	journal = {Science Robotics},
	publisher = {American Association for the Advancement of Science},
	author = {Hoeller, David and Rudin, Nikita and Sako, Dhionis and Hutter, Marco},
	month = mar,
	year = {2024},
}

@article{biggie_flexible_2023,
	title = {Flexible {Supervised} {Autonomy} for {Exploration} in {Subterranean} {Environments}},
	volume = {3},
	number = {1},
	urldate = {2024-08-08},
	journal = {Field Robotics},
	author = {Biggie, Harel and Rush, Eugene R. and Riley, Danny G. and Ahmad, Shakeeb and Ohradzansky, Michael T. and Harlow, Kyle and Miles, Michael J. and Torres, Daniel and McGuire, Steve and Frew, Eric W. and Heckman, Christoffer and Humbert, J. Sean},
	month = jan,
	year = {2023},
	pages = {125--189},
}

@article{kottege_heterogeneous_2025,
	title = {Heterogeneous {Robot} {Teams} {With} {Unified} {Perception} and {Autonomy}: {How} {Team} {CSIRO} {Data61} {Tied} for the {Top} {Score} at the {DARPA} {Subterranean} {Challenge}},
	volume = {2},
	urldate = {2025-09-29},
	journal = {IEEE Transactions on Field Robotics},
	author = {Kottege, Navinda and Williams, Jason and Tidd, Brendan and Talbot, Fletcher and Steindl, Ryan and Cox, Mark and Frousheger, Dennis and Hines, Thomas and Pitt, Alex and Tam, Benjamin and Wood, Brett and Hanson, Lauren and Lo Surdo, Katrina and Molnar, Tea and Wildie, Matt and Stepanas, Kazys and Catt, Gavin and Tychsen-Smith, Lachlan and Penfold, Dean and Overs, Les and Ramezani, Milad and Khosoussi, Kasra and Kendoul, Farid and Wagner, Glenn and Palmer, Duncan and Manderson, Jack and Medek, Corey and O'Brien, Matthew and Chen, Shengkang and Arkin, Ronald C.},
	year = {2025},
	pages = {100--130},
}

@article{achanta_slic_2012,
	title = {{SLIC} {Superpixels} {Compared} to {State}-of-the-{Art} {Superpixel} {Methods}},
	volume = {34},
	number = {11},
	urldate = {2025-01-13},
	journal = {IEEE Transactions on Pattern Analysis and Machine Intelligence},
	author = {Achanta, Radhakrishna and Shaji, Appu and Smith, Kevin and Lucchi, Aurelien and Fua, Pascal and Süsstrunk, Sabine},
	month = nov,
	year = {2012},
	pages = {2274--2282},
}

@misc{qiayuan_liao_and_others_legged_control_nodate,
	title = {\{legged\_control\}:  {NMPC}, {WBC}, state estimation, and sim2real framework for legged robots based on {OCS2} and ros-controls},
	author = {{Qiayuan Liao and others}},
	note = {[Online]. Available: {\textbackslash}url\{https://github.com/qiayuanl/legged\_control\}},
}

@inproceedings{feng_gpf-bg_2023,
	title = {{GPF}-{BG}: {A} hierarchical vision-based planning framework for safe quadrupedal navigation},
	booktitle = {2023 {IEEE} {International} {Conference} on {Robotics} and {Automation} ({ICRA})},
	author = {Feng, Shiyu and Zhou, Ziyi and Smith, Justin and Asselmeier, Max and Zhao, Ye and Vela, Patricio A.},
	year = {2023},
}

@inproceedings{griffin_footstep_2019,
	title = {Footstep {Planning} for {Autonomous} {Walking} {Over} {Rough} {Terrain}},
	booktitle = {2019 {IEEE}-{RAS} 19th {International} {Conference} on {Humanoid} {Robots} ({Humanoids})},
	author = {Griffin, Robert J. and Wiedebach, Georg and McCrory, Stephen and Bertrand, Sylvain and Lee, Inho and Pratt, Jerry},
	month = oct,
	year = {2019},
	pages = {9--16},
}

@misc{noauthor_anyboticselevation_mapping_2024,
	title = {{ANYbotics}/elevation\_mapping},
	copyright = {BSD-3-Clause},
	url = {https://github.com/ANYbotics/elevation_mapping},
	urldate = {2024-08-13},
	publisher = {ANYbotics},
	month = aug,
	year = {2024},
	note = {original-date: 2014-05-08T13:26:47Z},
}

@article{stepanas_ohm_2022,
	title = {{OHM}: {GPU} {Based} {Occupancy} {Map} {Generation}},
	volume = {7},
	number = {4},
	journal = {IEEE Robotics and Automation Letters},
	author = {Stepanas, Kazys and Williams, Jason and Hernández, Emili and Ruetz, Fabio and Hines, Thomas},
	month = oct,
	year = {2022},
	pages = {11078--11085},
}

@inproceedings{wellhausen_rough_2021,
	title = {Rough {Terrain} {Navigation} for {Legged} {Robots} using {Reachability} {Planning} and {Template} {Learning}},
	booktitle = {2021 {IEEE}/{RSJ} {International} {Conference} on {Intelligent} {Robots} and {Systems} ({IROS})},
	author = {Wellhausen, Lorenz and Hutter, Marco},
	month = sep,
	year = {2021},
}

@article{tranzatto_cerberus_2022,
	title = {{CERBERUS} in the {DARPA} {Subterranean} {Challenge}},
	volume = {7},
	number = {66},
	journal = {Science Robotics},
	author = {Tranzatto, Marco and Miki, Takahiro and Dharmadhikari, Mihir and Bernreiter, Lukas and Kulkarni, Mihir and Mascarich, Frank and Andersson, Olov and Khattak, Shehryar and Hutter, Marco and Siegwart, Roland and Alexis, Kostas},
	month = may,
	year = {2022},
}

@article{xu_potential_2021,
	title = {Potential {Gap}: {A} {Gap}-{Informed} {Reactive} {Policy} for {Safe} {Hierarchical} {Navigation}},
	volume = {6},
	number = {4},
	journal = {IEEE Robotics and Automation Letters},
	author = {Xu, Ruoyang and Feng, Shiyu and Vela, Patricio A.},
	year = {2021},
	pages = {8325--8332},
}

@book{gibson_ecological_2014,
	address = {New York},
	title = {The {Ecological} {Approach} to {Visual} {Perception}},
	isbn = {978-1-315-74021-8},
	doi = {10.4324/9781315740218},
	publisher = {Psychology Press},
	author = {Gibson, James J.},
	month = dec,
	year = {2014},
}

@article{wellhausen_where_2019,
	title = {Where {Should} {I} {Walk}? {Predicting} {Terrain} {Properties} {From} {Images} {Via} {Self}-{Supervised} {Learning}},
	volume = {4},
	number = {2},
	journal = {IEEE Robotics and Automation Letters},
	author = {Wellhausen, Lorenz and Dosovitskiy, Alexey and Ranftl, René and Walas, Krzysztof and Cadena, Cesar and Hutter, Marco},
	month = apr,
	year = {2019},
	pages = {1509--1516},
}

@article{kingston_scaling_2023,
	title = {Scaling {Multimodal} {Planning}: {Using} {Experience} and {Informing} {Discrete} {Search}},
	volume = {39},
	number = {1},
	journal = {IEEE Transactions on Robotics},
	author = {Kingston, Zachary and Kavraki, Lydia E.},
	month = feb,
	year = {2023},
	pages = {128--146},
}

@inproceedings{dudzik_robust_2020,
	title = {Robust {Autonomous} {Navigation} of a {Small}-{Scale} {Quadruped} {Robot} in {Real}-{World} {Environments}},
	booktitle = {2020 {IEEE}/{RSJ} {International} {Conference} on {Intelligent} {Robots} and {Systems} ({IROS})},
	author = {Dudzik, Thomas and Chignoli, Matthew and Bledt, Gerardo and Lim, Bryan and Miller, Adam and Kim, Donghyun and Kim, Sangbae},
	month = oct,
	year = {2020},
}

@inproceedings{melon_receding-horizon_2021,
	title = {Receding-{Horizon} {Perceptive} {Trajectory} {Optimization} for {Dynamic} {Legged} {Locomotion} with {Learned} {Initialization}},
	booktitle = {2021 {IEEE} {International} {Conference} on {Robotics} and {Automation} ({ICRA})},
	author = {Melon, Oliwier and Orsolino, Romeo and Surovik, David and Geisert, Mathieu and Havoutis, Ioannis and Fallon, Maurice},
	month = may,
	year = {2021},
	pages = {9805--9811},
}

@article{fankhauser_probabilistic_2018,
	title = {Probabilistic {Terrain} {Mapping} for {Mobile} {Robots} {With} {Uncertain} {Localization}},
	volume = {3},
	number = {4},
	journal = {IEEE Robotics and Automation Letters},
	author = {Fankhauser, Péter and Bloesch, Michael and Hutter, Marco},
	month = oct,
	year = {2018},
	pages = {3019--3026},
}

@inproceedings{grandia_perceptive_2022,
	title = {Perceptive {Locomotion} through {Nonlinear} {Model} {Predictive} {Control}},
	booktitle = {{arXiv}},
	author = {Grandia, Ruben and Jenelten, Fabian and Yang, Shaohui and Farshidian, Farbod and Hutter, Marco},
	month = aug,
	year = {2022},
}

@inproceedings{grandia_multi-layered_2021,
	title = {Multi-{Layered} {Safety} for {Legged} {Robots} via {Control} {Barrier} {Functions} and {Model} {Predictive} {Control}},
	booktitle = {2021 {IEEE} {International} {Conference} on {Robotics} and {Automation} ({ICRA})},
	author = {Grandia, Ruben and Taylor, Andrew J. and Ames, Aaron D. and Hutter, Marco},
	month = jun,
	year = {2021},
}

@inproceedings{lin_humanoid_2018,
	title = {Humanoid {Navigation} {Planning} in {Large} {Unstructured} {Environments} {Using} {Traversability} - {Based} {Segmentation}},
	issn = {2153-0866},
	doi = {10.1109/IROS.2018.8593694},
	booktitle = {2018 {IEEE}/{RSJ} {International} {Conference} on {Intelligent} {Robots} and {Systems} ({IROS})},
	author = {Lin, Yu-Chi and Berenson, Dmitry},
	month = oct,
	year = {2018},
	pages = {7375--7382},
}

@inproceedings{kim_highly_2019,
	title = {Highly {Dynamic} {Quadruped} {Locomotion} via {Whole}-{Body} {Impulse} {Control} and {Model} {Predictive} {Control}},
	booktitle = {{arXiv}},
	author = {Kim, Donghyun and Di Carlo, Jared and Katz, Benjamin and Bledt, Gerardo and Kim, Sangbae},
	month = sep,
	year = {2019},
}

@article{asselmeier_hierarchical_2024,
	title = {Hierarchical {Experience}-informed {Navigation} for {Multi}-modal {Quadrupedal} {Rebar} {Grid} {Traversal}},
	journal = {2024 IEEE International Conference on Robotics and Automation (ICRA)},
	author = {Asselmeier, Max and Ivanova, Jane and Zhou, Ziyi and Vela, Patricio A. and Zhao, Ye},
	year = {2024},
}

@inproceedings{lin_using_2020,
	title = {Using synthetic data and deep networks to recognize primitive shapes for object grasping},
	booktitle = {2020 {IEEE} {International} {Conference} on {Robotics} and {Automation} ({ICRA})},
	publisher = {IEEE},
	author = {Lin, Yunzhi and Tang, Chao and Chu, Fu-Jen and Vela, Patricio A},
	year = {2020},
	pages = {10494--10501},
}

@inproceedings{driess_deep_2020,
	title = {Deep {Visual} {Reasoning}: {Learning} to {Predict} {Action} {Sequences} for {Task} and {Motion} {Planning} from an {Initial} {Scene} {Image}},
	booktitle = {Robotics: {Science} and {Systems} {XVI}},
	publisher = {Robotics: Science and Systems Foundation},
	author = {Driess, Danny and Ha, Jung-Su and Toussaint, Marc},
	month = jul,
	year = {2020},
}

@misc{farbod_farshidian_and_others_ocs2_nodate,
	title = {{OCS2}:  {An}  open  source  library  for  {Optimal}  {Control}  of  {Switched} {Systems}},
	url = {https://github.com/leggedrobotics/ocs2}}

@inproceedings{smith_pips_2017,
	title = {{PiPS}: {Planning} in perception space},
	booktitle = {2017 {IEEE} {International} {Conference} on {Robotics} and {Automation} ({ICRA})},
	author = {Smith, Justin S. and Vela, Patricio},
	year = {2017},
	pages = {6204--6209},
}

@inproceedings{smith_egoteb_2020,
	title = {{egoTEB}: {Egocentric}, {Perception} {Space} {Navigation} {Using} {Timed}-{Elastic}-{Bands}},
	booktitle = {2020 {IEEE} {International} {Conference} on {Robotics} and {Automation} ({ICRA})},
	author = {Smith, Justin S. and Xu, Ruoyang and Vela, Patricio},
	year = {2020},
	pages = {2703--2709},
}
